\documentclass[11pt, letterpaper, logo, copyright]{googledeepmind}

\usepackage[utf8]{inputenc} % allow utf-8 input
\usepackage[T1]{fontenc}    % use 8-bit T1 fonts
\usepackage{xcolor}

\usepackage{hyperref}       % hyperlinks
\usepackage{url}            % simple URL typesetting
\usepackage{booktabs}       % professional-quality tables
\usepackage{amsfonts}       % blackboard math symbols
\usepackage{nicefrac}       % compact symbols for 1/2, etc.
\usepackage{microtype}      % microtypography
\usepackage{xcolor}         % colors
\usepackage{booktabs}
\usepackage{pgfplotstable}
\usepackage{amsmath} % For \DeclareMathOperator if needed, or just use text
\usepackage{etoolbox} % For \csdef, robust macro definitions
\usepackage{listings}
\usepackage{xcolor}

\usepackage{siunitx}       % For aligning numbers by decimal point and units
\usepackage{multirow}      % For multi-row cells (e.g., for N spanning header rows)
\usepackage[authoryear, sort&compress, round]{natbib}
\usepackage{booktabs}
\usepackage[font=small]{caption}
\usepackage{makecell}      % For line breaks within table cells (useful for long headers)

\lstdefinelanguage{json}{
    basicstyle=\ttfamily\small,
    numbers=left,
    numberstyle=\tiny\color{gray},
    stepnumber=1,
    numbersep=5pt,
    showstringspaces=false,
    breaklines=true,
    frame=single,
    backgroundcolor=\color{gray!5},
    literate=
     *{0}{{\textcolor{blue}{0}}}{1}
      {1}{{\textcolor{blue}{1}}}{1}
      {2}{{\textcolor{blue}{2}}}{1}
      {3}{{\textcolor{blue}{3}}}{1}
      {4}{{\textcolor{blue}{4}}}{1}
      {5}{{\textcolor{blue}{5}}}{1}
      {6}{{\textcolor{blue}{6}}}{1}
      {7}{{\textcolor{blue}{7}}}{1}
      {8}{{\textcolor{blue}{8}}}{1}
      {9}{{\textcolor{blue}{9}}}{1}
      {:}{{\textcolor{red}{:}}}{1}
      {,}{{\textcolor{red}{,}}}{1}
      {"}{{\textcolor{black}{"}}}{1},
}

\usepackage{float}
\usepackage{graphicx}
\usepackage{subcaption}
\usepackage{tikz}
\usepackage{float}
\usepackage{fontawesome5}
\definecolor{darkred}{RGB}{139,0,0} % You can tweak these values
\definecolor{azure}{RGB}{0,127,255}      % define "azure" (adjust as you like)
\usepackage{tcolorbox}
\tcbset{
  azurebox/.style={
    colback=white,                       % box background
    colframe=azure,                      % box frame color
    sharp corners,                       % square corners
    boxrule=1pt,                         % thickness of frame
    left=4pt, right=4pt, top=4pt, bottom=4pt
  }
}
\usepackage{wrapfig}
\usepackage{amsmath}   % For mathematical symbols
\usepackage{amsfonts}  % For mathematical fonts
\usepackage{algorithmic} % For the algorithmic environment
\usepackage{algorithm}   % For the algorithm environment
\usepackage{paralist}     % For compactitem environment
\newcounter{customalgo}
\usepackage{amssymb,amsmath,amsfonts,bm}
\def\eqref#1{equation~\ref{#1}}
\def\1{\bm{1}}

\DeclareMathAlphabet{\mathsfit}{\encodingdefault}{\sfdefault}{m}{sl}
\SetMathAlphabet{\mathsfit}{bold}{\encodingdefault}{\sfdefault}{bx}{n}

\definecolor{DarkBlue}{rgb}{0.1,0.1,0.5}
\definecolor{DarkGreen}{rgb}{0.1,0.5,0.1}
\definecolor{deepyellow}{RGB}{218, 174, 42}
\definecolor{darkcerulean}{rgb}{0.03, 0.27, 0.49}
\definecolor{denim}{rgb}{0.08, 0.38, 0.74}
\definecolor{AlgHighlight}{HTML}{228B22}
\definecolor{myyellow}{HTML}{FBBC05}

\newtheoremstyle{thmstyle}
{0.5em} % Space above
{0.15em} % Space below
{} % Body font
{} % Indent amount
{\bfseries} % Theorem head font
{.} % Punctuation after theorem head
{.5em} % Space after theorem head
{} % Theorem head spec (can be left empty, meaning `normal')

\theoremstyle{thmstyle}

\theoremstyle{definition}

\theoremstyle{remark}

\definecolor{PaperGreen}{RGB}{0,150,0}   % strong green, prints well
\definecolor{PaperRed}{RGB}{100,0,0}     % deep red, good contrast on white

\renewcommand{\1}{ \mathds{1}}

\renewcommand{\emptyset}{\varnothing}

\newcommand{\CommentLines}[1]{}

\newcolumntype{x}[1]{>{\centering\let\newline\\\arraybackslash\hspace{0pt}}m{#1}}
\definecolor{green}{HTML}{C6EFCE}
\definecolor{red}{HTML}{FFC7CE}
\definecolor{yellow}{HTML}{FFEB9C}
\definecolor{LightGray}{gray}{0.95}
\newcolumntype{H}{>{\columncolor{LightGray}}c}

\definecolor{darkgreen}{rgb}{0, 0.6, 0}

\definecolor{remarkblue}{HTML}{0B5CA3}

\newcommand{\benchmark}{\textsc{MINERVA-Cultural}}

\newcommand{\qwentwovl}{Qwen-2.5-VL}
\newcommand{\geminiflash}{Gemini-2.5-Flash}
\newcommand{\geminipro}{Gemini-2.5-Pro}
\newcommand{\gpt}{GPT-5}
\newcommand{\gptmini}{GPT-5-mini}
\newcommand{\claude}{Claude-Sonnet-4}
\newcommand{\qwenthreevl}{Qwen-3-VL}

\newcommand{\suppmat}{Supplementary Section}

\definecolor{veloblue}{RGB}{83, 149, 218}
\definecolor{veloorange}{RGB}{236, 133, 45}
\definecolor{lightgreen}{HTML}{E6F4EA}

\newcommand{\bluetext}[1]{{\color{veloblue}#1}}
\newcommand{\orangetext}[1]{{\color{veloorange}#1}}

\newcommand{\curators}{\textsc{\orangetext{Curators}}}
\newcommand{\auditors}{\textsc{{\bluetext{Auditors}}}}

\def\eg{\emph{e.g.}}

\newmdenv[backgroundcolor=cyan!10, linecolor=black, linewidth=0.5pt, roundcorner=4pt, innerleftmargin=2pt, innerrightmargin=2pt, innertopmargin=5pt, innerbottommargin=5pt]{exampleprompt}

\newcommand{\cmark}{{\textcolor{forestgreen}{\ding{51}}}}%
\newcommand{\xmark}{{\textcolor{darkred}{\ding{55}}}}%

\definecolor{forestgreen}{rgb}{0.13, 0.55, 0.13}
\definecolor{fireenginered}{rgb}{0.81, 0.09, 0.13}
\definecolor{darkred}{rgb}{0.6, 0.1, 0.1}

\newcommand{\todods}[1]{}
\newcommand{\todohs}[1]{}
\newcommand{\DS}[1]{}
\newcommand{\arsha}[1]{}
\newcommand{\todos}[1]{}

\usepackage{tcolorbox}
\tcbuselibrary{listings, skins, breakable, xparse}

\definecolor{pastelBlue}{RGB}{173, 216, 230}
\definecolor{pastelPink}{RGB}{255, 209, 220}
\definecolor{pastelLavender}{RGB}{230, 230, 250}
\definecolor{pastelMint}{RGB}{204, 255, 204}
\definecolor{softPurple}{RGB}{209, 195, 240}
\definecolor{softGray}{RGB}{240, 240, 245}
\definecolor{deepPurple}{RGB}{95, 75, 139}
\definecolor{darkGray}{RGB}{80, 80, 80}

\newtcolorbox{promptbox}[2][]{%
  enhanced,
  breakable,
  colback=softGray,
  colframe=softPurple,
  fonttitle=\bfseries\rmfamily,
  coltitle=white,
  colbacktitle=deepPurple,
  attach boxed title to top left={xshift=0.5cm, yshift=-\tcboxedtitleheight/5},
  boxed title style={size=small, sharp corners},
  top=2mm,
  bottom=2mm,
  left=2mm,
  right=2mm,
  arc=2mm,
  boxrule=0.8pt,
  titlerule=0mm,
  toptitle=1mm,
  bottomtitle=1mm,
  title={#2},
  overlay={
    \begin{tcbclipinterior}
      \fill[pastelLavender!30] (interior.south west) -- (interior.north west) -- (interior.north east) -- cycle;
    \end{tcbclipinterior}
  },
  listing only,
  listing options={
    basicstyle=\small\ttfamily,
    breaklines=true,
    columns=flexible,
    backgroundcolor=\color{softGray},
    xleftmargin=2pt,
    framexleftmargin=2pt,
    numbers=left,
    numberstyle=\tiny\color{darkGray},
    numbersep=5pt,
    tabsize=2,
    commentstyle=\color{deepPurple},
    keywordstyle=\color{blue!70!black},
    stringstyle=\color{red!70!black},
  },
  #1
}

\newtcolorbox{tablebox}[2][]{%
  enhanced, breakable, colback=pastelLavender!15, colframe=softPurple,
  fonttitle=\bfseries\rmfamily, coltitle=white, colbacktitle=deepPurple,
  attach boxed title to top left={xshift=0.5cm,yshift=-\tcboxedtitleheight/5},
  boxed title style={size=small, sharp corners},
  top=2mm,bottom=2mm,left=2mm,right=2mm,arc=2mm,boxrule=0.8pt,
  title={#2}, #1}

\newtcolorbox{answerbox}[2][]{%
  enhanced, breakable, colback=softGray, colframe=pastelMint,
  fonttitle=\bfseries\rmfamily, coltitle=darkGray, colbacktitle=pastelMint!40!white,
  attach boxed title to top left={xshift=0.5cm,yshift=-\tcboxedtitleheight/5},
  boxed title style={size=small, sharp corners},
  top=2mm,bottom=2mm,left=2mm,right=2mm,arc=2mm,boxrule=0.8pt,
  title={#2}, #1}

\newtcolorbox{metricsbox}[1][]{%
  enhanced, breakable, colback=pastelBlue!15, colframe=pastelBlue!70!black,
  fonttitle=\bfseries\rmfamily, coltitle=white, colbacktitle=pastelBlue!70!black,
  attach boxed title to top left={xshift=0.5cm,yshift=-\tcboxedtitleheight/5},
  boxed title style={size=small, sharp corners},
  top=1.5mm,bottom=1.5mm,left=1.5mm,right=1.5mm,arc=2mm,boxrule=0.6pt,
  title={Attributes and their Relative Importance Scores}, #1}

\newtcolorbox{jsonbox}[2][]{%
  enhanced, breakable,
  colback=softGray, colframe=softPurple, arc=1.5mm, boxrule=0.6pt,
  title={#2}, fonttitle=\bfseries\rmfamily, coltitle=white,
  colbacktitle=deepPurple!90!black,
  attach boxed title to top left={xshift=0.5cm,yshift=-\tcboxedtitleheight/5},
  boxed title style={size=small, sharp corners},
  listing only,
  listing options={basicstyle=\scriptsize\ttfamily,
                   breaklines=true, numbers=left,
                   numberstyle=\tiny\color{darkGray},
                   xleftmargin=2pt, framexleftmargin=2pt,
                   showstringspaces=false},
  #1}

\definecolor{questionBlue}{RGB}{208, 235, 250}       % Lighter blue
\definecolor{questionBorder}{RGB}{121, 182, 242}     % Lighter blue border
\definecolor{questionTitle}{RGB}{70, 130, 180}       % Soft blue title

\definecolor{acceptGreen}{RGB}{225, 250, 225}        % Lighter green
\definecolor{acceptBorder}{RGB}{150, 200, 150}       % Lighter green border
\definecolor{acceptTitle}{RGB}{76, 156, 80}          % Soft green title

\definecolor{rejectPink}{RGB}{255, 233, 233}         % Lighter pink/red
\definecolor{rejectBorder}{RGB}{244, 162, 160}       % Lighter red border 
\definecolor{rejectTitle}{RGB}{175, 70, 70}          % Soft red title

\definecolor{remarkViolet}{RGB}{240, 230, 252}       % Light violet
\definecolor{remarkBorder}{RGB}{190, 158, 230}       % Light violet border
\definecolor{remarkTitle}{RGB}{130, 94, 180}         % Soft purple title

\definecolor{darkGray}{RGB}{100, 100, 100}           % Text color

\newtcolorbox{questionbox}[2][]{%
  enhanced,
  breakable,
  colback=questionBlue!70,
  colframe=questionBorder,
  fonttitle=\bfseries\rmfamily,
  coltitle=white,
  colbacktitle=questionTitle,
  attach boxed title to top left={xshift=0.5cm, yshift=-\tcboxedtitleheight/5},
  boxed title style={size=small, sharp corners},
  top=2mm,
  bottom=2mm,
  left=2mm,
  right=2mm,
  arc=3mm,
  boxrule=0.8pt,
  titlerule=0mm,
  toptitle=1mm,
  bottomtitle=1mm,
  title={#2},
  overlay={
    \begin{tcbclipinterior}
      \fill[questionBlue!40] (interior.south west) -- (interior.north west) -- (interior.north east) -- cycle;
    \end{tcbclipinterior}
  },
  #1
}

\newtcolorbox{acceptedbox}[2][]{%
  enhanced,
  breakable,
  colback=acceptGreen!80,
  colframe=acceptBorder,
  fonttitle=\bfseries\rmfamily,
  coltitle=white,
  colbacktitle=acceptTitle,
  attach boxed title to top right={xshift=-0.5cm, yshift=-\tcboxedtitleheight/5},
  boxed title style={size=small, sharp corners},
  top=2mm,
  bottom=2mm,
  left=2mm,
  right=2mm,
  arc=3mm,
  boxrule=0.8pt,
  titlerule=0mm,
  toptitle=1mm,
  bottomtitle=1mm,
  title={#2},
  overlay={
    \begin{tcbclipinterior}
      \fill[acceptGreen!40] (interior.north east) -- (interior.south east) -- (interior.south west) -- cycle;
    \end{tcbclipinterior}
  },
  #1
}

\newtcolorbox{rejectedbox}[2][]{%
  enhanced,
  breakable,
  colback=rejectPink!80,
  colframe=rejectBorder,
  fonttitle=\bfseries\rmfamily,
  coltitle=white,
  colbacktitle=rejectTitle,
  attach boxed title to top right={xshift=-0.5cm, yshift=-\tcboxedtitleheight/5},
  boxed title style={size=small, sharp corners},
  top=2mm,
  bottom=2mm,
  left=2mm,
  right=2mm,
  arc=3mm,
  boxrule=0.8pt,
  titlerule=0mm,
  toptitle=1mm,
  bottomtitle=1mm,
  title={#2},
  overlay={
    \begin{tcbclipinterior}
      \fill[rejectPink!40] (interior.north east) -- (interior.south east) -- (interior.south west) -- cycle;
    \end{tcbclipinterior}
  },
  #1
}

\newtcolorbox{remarksbox}[2][]{%
  enhanced,
  breakable,
  colback=remarkViolet!70,
  colframe=remarkBorder,
  fonttitle=\bfseries\rmfamily,
  coltitle=white,
  colbacktitle=remarkTitle,
  attach boxed title to top center={yshift=-\tcboxedtitleheight/5},
  boxed title style={size=small, sharp corners},
  top=2mm,
  bottom=2mm,
  left=2mm,
  right=2mm,
  arc=3mm,
  boxrule=0.8pt,
  titlerule=0mm,
  toptitle=1mm,
  bottomtitle=1mm,
  title={#2},
  overlay={
    \begin{tcbclipinterior}
      \fill[remarkViolet!40] (interior.north east) -- (interior.south east) -- (interior.south west) -- cycle;
    \end{tcbclipinterior}
  },
  #1
}

\newtcolorbox{questionboxverbatim}[2][]{%
  enhanced, breakable, colback=questionBlue!70, colframe=questionBorder,
  fonttitle=\bfseries\rmfamily, coltitle=white, colbacktitle=questionTitle,
  attach boxed title to top left={xshift=0.5cm, yshift=-\tcboxedtitleheight/5},
  boxed title style={size=small, sharp corners},
  top=2mm, bottom=2mm, left=2mm, right=2mm, arc=3mm, boxrule=0.8pt,
  titlerule=0mm, toptitle=1mm, bottomtitle=1mm,
  overlay={\begin{tcbclipinterior}\fill[questionBlue!40] (interior.south west) -- (interior.north west) -- (interior.north east) -- cycle;\end{tcbclipinterior}},
  title={#2}, % Use the standard 'title' key
  listing only,
  listing options={
    basicstyle=\small\ttfamily, breaklines=true, columns=flexible,
    xleftmargin=5pt, framexleftmargin=2pt, numbers=left,
    numberstyle=\tiny\color{darkGray}, numbersep=5pt, tabsize=2,
  },
  #1 % Optional arguments
}

\newtcolorbox{acceptedboxverbatim}[2][]{%
  enhanced, breakable, colback=acceptGreen!80, colframe=acceptBorder,
  fonttitle=\bfseries\rmfamily, coltitle=white, colbacktitle=acceptTitle,
  attach boxed title to top right={xshift=-0.5cm, yshift=-\tcboxedtitleheight/5},
  boxed title style={size=small, sharp corners},
  top=2mm, bottom=2mm, left=2mm, right=2mm, arc=3mm, boxrule=0.8pt,
  titlerule=0mm, toptitle=1mm, bottomtitle=1mm,
  overlay={\begin{tcbclipinterior}\fill[acceptGreen!40] (interior.north east) -- (interior.south east) -- (interior.south west) -- cycle;\end{tcbclipinterior}},
  title={#2}, % Use the standard 'title' key
  listing only,
  listing options={
    basicstyle=\small\ttfamily, breaklines=true, columns=flexible,
    xleftmargin=5pt, framexleftmargin=2pt, numbers=left,
    numberstyle=\tiny\color{darkGray}, numbersep=5pt, tabsize=2,
  },
  #1 % Optional arguments
}

\newtcolorbox{rejectedboxverbatim}[2][]{%
  enhanced, breakable, colback=rejectPink!80, colframe=rejectBorder,
  fonttitle=\bfseries\rmfamily, coltitle=white, colbacktitle=rejectTitle,
  attach boxed title to top right={xshift=-0.5cm, yshift=-\tcboxedtitleheight/5},
  boxed title style={size=small, sharp corners},
  top=2mm, bottom=2mm, left=2mm, right=2mm, arc=3mm, boxrule=0.8pt,
  titlerule=0mm, toptitle=1mm, bottomtitle=1mm,
  overlay={\begin{tcbclipinterior}\fill[rejectPink!40] (interior.north east) -- (interior.south east) -- (interior.south west) -- cycle;\end{tcbclipinterior}},
  title={#2}, % Use the standard 'title' key
  listing only,
  listing options={
    basicstyle=\small\ttfamily, breaklines=true, columns=flexible,
    xleftmargin=5pt, framexleftmargin=2pt, numbers=left,
    numberstyle=\tiny\color{darkGray}, numbersep=5pt, tabsize=2,
  },
  #1 % Optional arguments
}

\newtcolorbox{remarksboxverbatim}[2][]{%
  remarksbox={#2},
  listing only,
  listing options={
    basicstyle=\small\ttfamily,
    breaklines=true,
    columns=flexible,
    backgroundcolor=\color{remarkViolet!70},
    xleftmargin=2pt,
    framexleftmargin=2pt,
    numbers=left,
    numberstyle=\tiny\color{darkGray},
    numbersep=5pt,
    tabsize=2,
  },
  #1
}

\NewDocumentCommand{\Question}{O{} m m}{%
  \begin{questionbox}[#1]{#2}
#3
  \end{questionbox}
}

\NewDocumentCommand{\Accepted}{O{} m m}{%
  \begin{acceptedbox}[#1]{#2}
#3
  \end{acceptedbox}
}

\NewDocumentCommand{\Rejected}{O{} m m}{%
  \begin{rejectedbox}[#1]{#2}
#3
  \end{rejectedbox}
}

\NewDocumentCommand{\Remarks}{O{} m m}{%
  \begin{remarksbox}[#1]{#2}
#3
  \end{remarksbox}
}

\NewDocumentCommand{\QuestionV}{O{} m m}{%
  \begin{questionboxverbatim}[#1]{#2}
#3
  \end{questionboxverbatim}
}

\NewDocumentCommand{\AcceptedV}{O{} m m}{%
  \begin{acceptedboxverbatim}[#1]{#2}
#3
  \end{acceptedboxverbatim}
}

\NewDocumentCommand{\RejectedV}{O{} m m}{%
  \begin{rejectedboxverbatim}[#1]{#2}
#3
  \end{rejectedboxverbatim}
}

\NewDocumentCommand{\RemarksV}{O{} m m}{%
  \begin{remarksboxverbatim}[#1]{#2}
#3
  \end{remarksboxverbatim}
}

\title{\benchmark{}: A Benchmark for Cultural and Multilingual Long Video Reasoning}

\correspondingauthor{{\{darshanss, anagrani, shachi\}@google.com}}

\author[1]{Darshan Singh}
\author[1]{Arsha Nagrani}
\author[1]{Kawshik Manikantan}
\author[2]{Harman Singh}
\author[1]{Dinesh Tewari}
\author[1]{Tobias Weyand}
\author[1]{\\ Cordelia Schmid}
\author[1]{Anelia Angelova}
\author[1]{Shachi Dave}

\affil[1]{Google DeepMind}
\affil[2]{UC Berkeley}

\begin{abstract}
\vspace{-0.55cm}
Recent advancements in video models have shown tremendous progress, particularly in long video understanding. However, current benchmarks predominantly feature western-centric data and English as the dominant language, introducing significant biases in evaluation. To address this, we introduce \benchmark{}, a challenging benchmark for multicultural and multilingual video reasoning. \benchmark{} comprises high-quality, entirely human-generated annotations from diverse, region-specific cultural videos across 18 global locales. Unlike prior work that relies on automatic translations, \benchmark{} provides complex questions, answers, and multi-step reasoning steps, all crafted in native languages. Making progress on \benchmark{} requires a deeply situated understanding of visual cultural context. Furthermore, we leverage \benchmark{}'s reasoning traces to construct evidence-based graphs and propose a novel iterative strategy using these graphs to identify fine-grained errors in reasoning. Our evaluations reveal that SoTA Video-LLMs struggle significantly, performing substantially below human-level accuracy, with errors primarily stemming from the visual perception of cultural elements. 
\benchmark{} will be publicly available \href{https://github.com/google-deepmind/neptune?tab=readme-ov-file\#minerva-cultural}{here}.

\end{abstract}

\begin{document}

% \vspace{-0.2cm}
\maketitle

\vspace{-0.4cm}
\section{Introduction}
\label{sec:intro}

\begin{figure}[ht]
    \centering
    \includegraphics[width=\columnwidth]{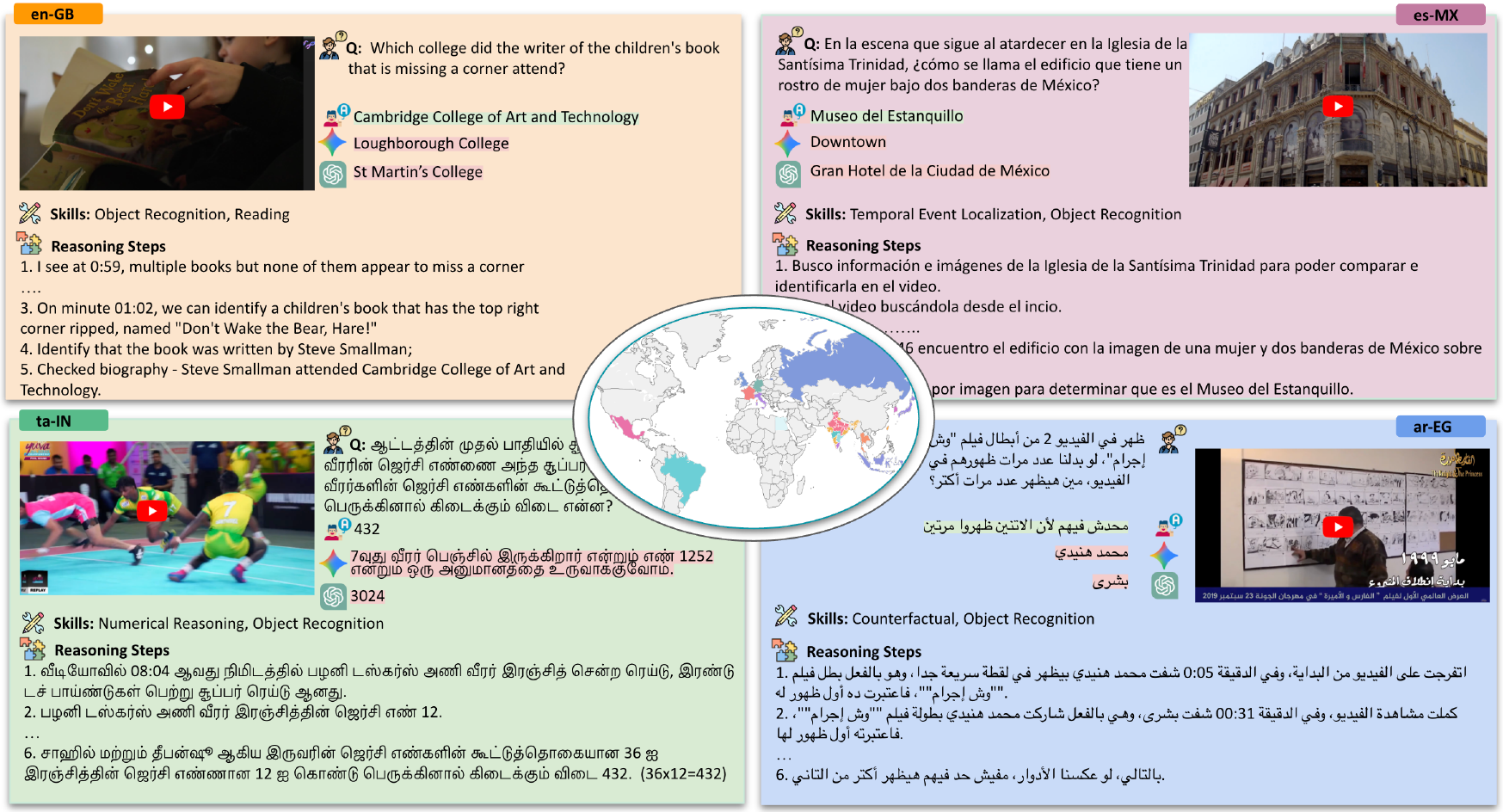}
    \caption{\textbf{\benchmark{} for benchmarking Cultural and Multilingual Video Reasoning}. We show examples from four distinct global locales: \texttt{en-GB} (English, UK), \texttt{es-MX} (Spanish, Mexico), \texttt{ta-IN} (Tamil, India), \texttt{ar-EG} (Arabic, Egypt) within the \textbf{\benchmark{}} benchmark. Each example consists of a \emph{video}, a complex \emph{native-language question}, the ground-truth answer, and two \textit{representative and incorrect SOTA VLM responses}. We also show the \emph{associated reasoning skills} and \emph{annotated human reasoning steps} provided with each question.}
    \label{fig:teaser}
\end{figure}

Video understanding has made significant strides, yet current evaluation benchmarks exhibit a pervasive bias towards western and English-centric content~\citep{Vayani_2025_CVPR, shafique2025culturally, pfeiffer-2022-xgqa}. This bias severely limits the ability of vision-language models (VLMs) to effectively operate in diverse global contexts, where cultural nuances, varied linguistic expressions, and unique visual characteristics are paramount~\citep{romero2024cvqa}. 
While expanding language coverage, existing datasets often rely on direct translation of annotations~\citep{pfeiffer-2022-xgqa, Liu_2021_EMNLP, shafique2025culturally}, but their visual content remains situated in Western concepts.

To address this critical gap, we introduce \benchmark{}, a challenging open-ended video question-answering benchmark. \benchmark{} comprises a rich collection of videos depicting diverse cultural events (for example local sports, diverse festivals, local cuisine) from 18 regions and languages across the world (see Figure~\ref{fig:teaser} for a snapshot). A crucial aspect of our dataset is that each video is uploaded with audio in a native language, and is paired with complex, multi-step reasoning questions crafted in the same native language. Each question is annotated by locally-situated experts with native cultural and language proficiency, and requires models to reason along different axes - (i) locale-specific cultural understanding, (ii) multimodal inference from audio-visual cues, and (iii) intricate temporal relationships (as videos vary in length from 1 minute to upto 1 hour).

\begin{table}[t]
\centering
\footnotesize
\tabcolsep=1.1pt
\resizebox{\columnwidth}{!}{%
\begin{tabular}{l c c c c c c r}
\toprule
{Benchmark} & {{Task}} & \makecell{Human\\Curated$^\dagger$} & \makecell{Multi-cultural\\(\# Regions)} & \makecell{Multi-lingual\\(\# Langs)} & \makecell{Human\\Reasoning\\Traces} & \makecell{Human\\Evals} & \makecell[r]{ Video\\Durn. \\(avg)} \\
\midrule

Percept.Test({\scriptsize \cite{perception_test}}) & M & \cmark & \xmark & \xmark & \xmark & \cmark & 0.40 \\
Cinepile({\scriptsize \cite{cinepile}}) & M & \xmark & \xmark & \xmark & \xmark & \cmark & 2.66 \\
MV-Bench({\scriptsize \cite{mvbench}}) & M & \xmark & \xmark & \xmark & \xmark & \xmark & 16.00 \\
TempCompass({\scriptsize \cite{tempcompass}}) & M & \xmark & \xmark & \xmark & \xmark & \cmark & 0.50 \\
MMBench-Vid.({\scriptsize \cite{mmbenchvideo}}) & O & \cmark & \xmark & \xmark & \xmark & \xmark & 2.75 \\
Video-MME({\scriptsize \cite{videomme}}) & M & \cmark & \xmark & \xmark & \xmark & \xmark & 16.95 \\

LongVideoB.({\scriptsize \cite{wu2024longvideobench}}) & M & \cmark & \xmark & \xmark & \xmark & \xmark & 7.91 \\
MLVU({\scriptsize \cite{zhou2024mlvu}}) & M & \cmark & \xmark & \xmark & \xmark & \xmark & 15.5 \\
EgoSchema({\scriptsize \cite{mangalam2023egoschema}}) & M & \xmark & \xmark & \xmark & \xmark & \cmark & 3.00 \\
Percept.Test'24({\scriptsize \cite{heyward2024perceptiontest2024challenge}}) & M & \cmark & \xmark & \xmark & \xmark & \cmark & 60.00 \\
VRBench({\scriptsize \cite{yu2025vrbenchbenchmarkmultistepreasoning}}) & B & \cmark & \xmark & \cmark~(8) & \cmark & \xmark & 96.60 \\

Neptune({\scriptsize \cite{nagrani2024neptune}}) & M & \xmark & \xmark & \xmark & \xmark & \cmark & 2.50 \\
MINERVA({\scriptsize \cite{minerva}} & M & \cmark & \xmark & \xmark & \cmark & \cmark & 12.00 \\

M3-Med({\scriptsize \cite{liu2025m3medbenchmarkmultilingualmultimodal}}) & M & \cmark & \xmark & \cmark~(2) & \xmark & \xmark & -- \\

ViMUL-Bench({\scriptsize \cite{shafique2025culturally}}) & B & \xmark & \cmark~(14) & \cmark~(14) & \xmark & \xmark & 4.6 \\
\midrule
\rowcolor{lightgreen}
\benchmark{} (Ours) & O & \cmark & \cmark~(18) & \cmark~(18) & \cmark & \cmark & 12.62 \\
\bottomrule
\end{tabular}%
}
\caption{
Representative VideoQA benchmarks and key differences with \benchmark{}. Task formats include Multiple Choice (M), Open-ended QA (O), or Both (B). \benchmark{} introduces multilingual reasoning traces and human-curated multicultural videos spanning diverse regions. $^\dagger$Fully human curated (No synthetic data involved).
}
\label{tab:BenchmarkComparison}
\end{table}

Existing cultural benchmarks, such as ViMUL-Bench~\citep{shafique2025culturally}, often focus solely on the final answer and overlook the underlying reasoning process. To allow for a more detailed analysis of model failures, each question in our dataset also comes with a detailed, multi-step reasoning trace. We use these reasoning traces to dig deeper into analyzing the failures made by models via a evidence-style Directed Acyclic Graph (DAG) analysis. This provides far more insight than a simple measure of final accuracy, allowing us to probe \textit{where} and \textit{how} models make mistakes.  In our graph, nodes represent \emph{atomic evidences} and edges signify prerequisites. A single error can cause cascading failures, creating a dilemma: penalizing all subsequent errors \emph{over-counts} them, while stopping at the first limits diagnostic insight. Hence, we propose \emph{Iterative Error Isolation}, a three-stage loop that traverses the evidence-style DAG, detects and tags errors, and iteratively re-evaluates with corrective hints until the problem is solved correctly. This process allows independent assessment of each intermediate step, revealing that the biggest mistakes are in the visual perception of cultural elements.

Our contributions are: 
(i) We introduce \benchmark{}, a multi-cultural, multi-lingual, long video reasoning benchmark for measuring and fostering the development of more equitable and globally competent multimodal AI systems. \benchmark{} consists of 2400 questions from 540 videos from 18 locales all over the world. Questions in \benchmark{} require various skills along with cultural understanding to answer. The benchmark will be publicly released to accelerate research towards truly multicultural video understanding. (ii) We benchmark SoTA frontier VLMs, and find a significant gap to human performance. We perform extensive analysis showing the effect of modalities, number of frames, and test-time computational scaling. (iii) We perform extensive error tagging using evidence graphs to analyze where the models tend to make mistakes, and show that 75\% of all failures can be attributed to cultural visual perception.  
\section{Related Work}
\label{sec:related}

\begin{figure*}[ht]
\centering
\includegraphics[width=\textwidth]{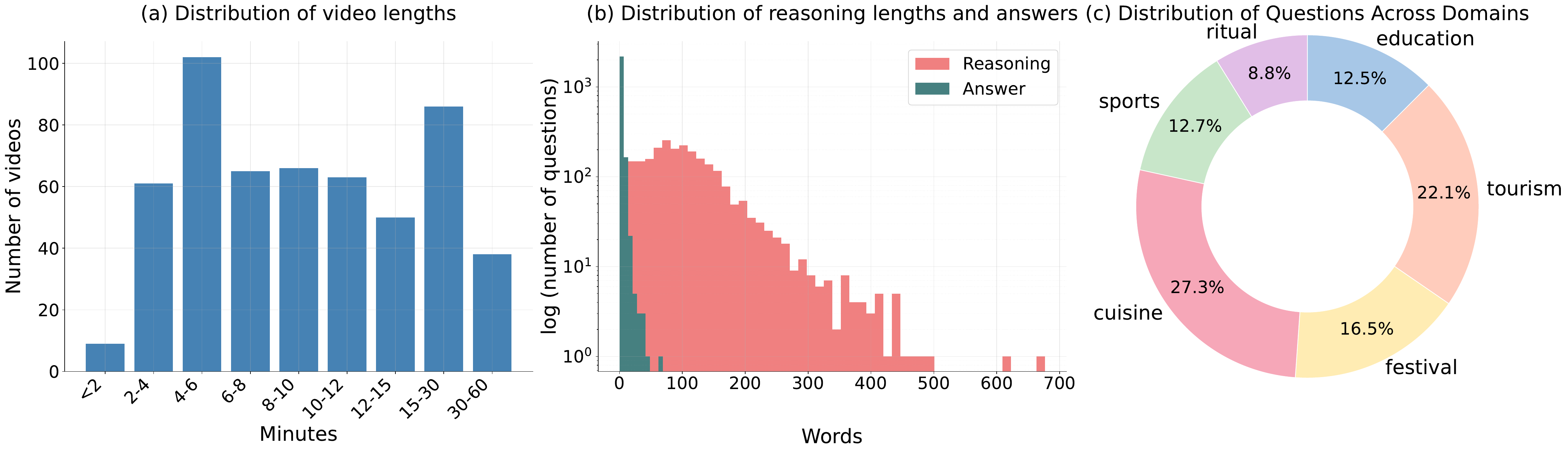}
\caption{\textbf{\benchmark{} Benchmark Statistics.} \textbf{(a)} \benchmark{} contains a wide range of video durations, ranging from one minute to over an hour. \textbf{(b)} Human-authored reasoning traces are long and detailed, often spanning hundreds of words, while the corresponding answers are concise. \textbf{(c)} Questions are spread across six core cultural domains.}
\label{fig:dataset_stats}
\end{figure*}

\textbf{Multicultural-Multimodal Evaluations.} Equitable development and evaluation of foundation models requires cultural and linguistic inclusivity~\citep{pfeiffer-2022-xgqa, romero2024cvqa, Nayak_2024_EMNLP, Liu_2021_EMNLP, Zhang_2023_NeurIPS, Das_2024_ACL, Vayani_2025_CVPR, Khanuja_2024_EMNLP}. Early efforts like xGQA~\citep{pfeiffer-2022-xgqa} and MaRVL~\citep{Liu_2021_EMNLP} expanded language coverage by translating English annotations, but retained Western-centric imagery and potential translation artifacts. Subsequent benchmarks such as CVQA~\citep{romero2024cvqa} and CulturalVQA~\citep{Nayak_2024_EMNLP} incorporated culturally specific imagery, revealing significant performance gaps in modern VLMs while CUBE~\citep{cube} showed a lack of diversity in generations from T2I models. Other works have targeted multilingual image captioning with diverse geography (Crossmodal-3600~\citep{Thapliyal_2022_EMNLP}), academic reasoning (M3Exam~\citep{Zhang_2023_NeurIPS}, EXAMS-V~\citep{Das_2024_ACL}), and general cultural understanding (ALM-Bench~\citep{Vayani_2025_CVPR}), all underscoring persistent model brittleness. Large-scale benchmarks like MMMLU~\citep{Mmmlu} and PangeaBench~\citep{yue2024pangea} have increased language coverage through human and machine translation. \benchmark{} focuses on multicultural and multilingual audio-visual question-answering.

\textbf{VideoQA Evaluation}
VideoQA evaluation benchmarks target complex understanding of perceptual and temporal information~\citep{nagrani2024neptune, Fu_2025_CVPR, wu2024longvideobench, mangalam2023egoschema, zhou2024mlvu, perception_test, cinepile, videocon, seedbench2, tempcompass, vitatecs, minerva, velociti, cvrres, vinoground, tvbench}. A major focus is on long-form video understanding and fine-grained temporal reasoning, as seen in EgoSchema~\citep{mangalam2023egoschema}, LongVideoBench~\citep{wu2024longvideobench}, Perception Test'24~\citep{heyward2024perceptiontest2024challenge}, and MLVU~\citep{zhou2024mlvu}. MINERVA~\citep{minerva} and Neptune~\citep{nagrani2024neptune} probe complex reasoning, showing that models still lag human reasoning and degrade with longer inputs. Other efforts have broadened the scope to diverse content (VideoVista~\citep{li2024videovista}), temporal tasks (MVBench~\citep{mvbench}), or situated reasoning (SOK-Bench~\citep{SOK_Bench}). Most VideoQA benchmarks are predominantly\textit{ English-centric}. Nascent multilingual efforts include captioning (VATEX~\citep{Wang_2019_ICCV}), story understanding (M-SyMoN~\citep{sun2024multilingual}), and domain-specific QA (M\textsuperscript{3}-Med~\citep{liu2025m3medbenchmarkmultilingualmultimodal}). ViMUL-Bench~\citep{shafique2025culturally} offers broad language coverage but mixes cultural videos with generic sources relies partly on translation. We compare \benchmark{} with other video understanding benchmarks in Table \ref{tab:BenchmarkComparison}. \benchmark{} introduces the first large-scale, multicultural video reasoning benchmark with fully human-curated questions, answers and reasoning traces in 18 native languages. It features long, culturally-situated videos with audio, requiring multi-step reasoning supported by ground-truth traces for detailed analysis of SOTA models~\citep{gemini2.5, reid2024gemini, bai2025qwen2, gpt5systemcard, jaech2024openai}.

\textbf{Evaluating Reasoning.}
Recent advances in model reasoning across domains like mathematics~\citep{shao2024deepseekmathpushinglimitsmathematical, Trinh2024SolvingOG}, logic~\citep{zelikman2024quietstarlanguagemodelsteach, zhou2024selfdiscoverlargelanguagemodels}, and coding~\citep{yang2024sweagentagentcomputerinterfacesenable, deepseekai2024deepseekcoderv2breakingbarrierclosedsource} now extend to vision and video~\citep{gpt5systemcard, gemini2.5}, increasing the need for robust evaluation of reasoning processes~\citep{lee2025evaluatingstepbystepreasoningtraces, xiong-etal-2025-mapping, wang2024boostinglanguagemodelsreasoning, atanasova-etal-2023-faithfulness}. Evaluations grounded in human-verified reasoning traces~\citep{minerva, deutsch2022limitations} are often more reliable than automated self-correction methods~\citep{huang2024largelanguagemodelsselfcorrect, kamoi-etal-2024-llms, tyen2024llmsreasoningerrorscorrect}.
\benchmark{}'s human-annotated traces enable such reliable evaluation. While MINERVA~\citep{minerva} also uses human traces to categorize errors, we advance this methodology by introducing a graph-based framework for iterative error localization. This approach effectively captures complex video relations like causality and spatio-temporal dependencies~\citep{arnab2021unifiedgraphstructuredmodels, chu2025understandinglongvideosllmpowered} and reframes evaluation as an evidence classification task, which is well-suited for modern LLMs~\citep{hsieh2024rulerwhatsrealcontext, dougrez-lewis-etal-2025-assessing}.
\vspace{-0.1cm}
\section{\benchmark{} Benchmark}
\label{sec:datacuration}

\benchmark{} is a large-scale benchmark for multicultural video reasoning. It comprises \textbf{2400 questions} across \textbf{540 videos}, spanning \textbf{18 diverse global locales} and their native languages. We strategically selected \benchmark{}'s initial 18 locales for broad diversity across continents and the Global South-North divide. While this set is not exhaustive, it provides a robust starting point, and our framework is designed to be extensible. Each locale has around 130 questions on an average. Throughout this paper, each locale is denoted by a standardized \texttt{language-region} code, where the language is a two- or three-letter lowercase code from \href{https://en.wikipedia.org/wiki/List_of_ISO_639_language_codes}{ISO 639} and the region is a two-letter uppercase code from \href{https://en.wikipedia.org/wiki/ISO_3166-1_alpha-2}{ISO 3166-1 alpha-2}. For example, Tamil (India) is \texttt{ta-IN}, Spanish (Mexico) is \texttt{es-MX}, and English (Britain) is \texttt{en-GB}.

\begin{figure}[ht]
\centering
\includegraphics[width=0.9\textwidth]{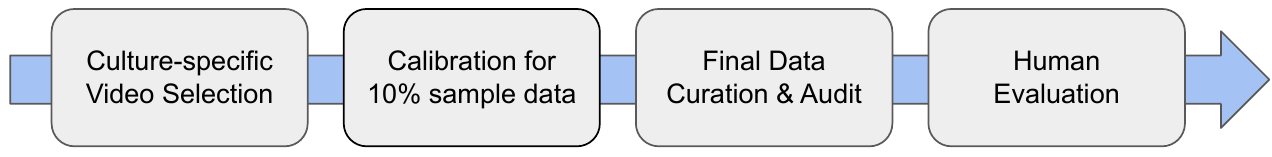}
\caption{\textbf{The Human Annotation Pipeline.} Four stages of our data curation process:  Culture-specific video selection, 10\% sample calibration, final data curation and audit (Section~\ref{sec:human_curation_pipeline}), and human evaluation (Section~\ref{sec:human_eval}). Each stage involves rigorous human curation and verification to ensure data quality and difficulty.}
\label{fig:human_annotation}
\end{figure}

\vspace{-0.3cm}
\subsection{Dataset Overview and Properties}

A key feature of \benchmark{} is its focus on long-form, culturally-rich content that demands complex, temporally-grounded understanding. As shown in Figure~\ref{fig:dataset_stats}(a), video durations range from one minute to over an hour. The dataset is structured around six key cultural domains: Sports, Cuisine, Festivals, Tourism, Rituals, and Education, with a balanced question distribution across them (Figure~\ref{fig:dataset_stats}(c)).

Each question is paired with a detailed, human-authored reasoning trace in the \textbf{native language}, providing a ground-truth for reasoning evaluation. The complexity of the required reasoning is evident in the substantial length of these traces, as detailed in Figure~\ref{fig:dataset_stats}(b). Every question is intentionally crafted: (i) to be answerable only through a deep, situated understanding of visual and cultural context; (ii) require multimodal inference from both audio and visual cues; and (iii) requires at least two of the following skills: Temporal Ordering, Goal Reasoning, Event Occurrence, Reading, Listening, Spatial Perception, Temporal Event Localization, Counting, Cause and Effect, Numerical Reasoning, Object Recognition, Counterfactual Reasoning along with a mandatory \emph{Visual Cultural Understanding} skill. These properties make \benchmark{} a challenging and comprehensive benchmark for assessing the multi-cultural reasoning capabilities of modern VLMs. More details on the skills are provided in the \suppmat{}~\ref{supp_sec:guidelines}.

\subsection{Human-Centric Curation Pipeline}
\label{sec:human_curation_pipeline}
Construction of \benchmark{} followed a meticulous, multi-stage pipeline (Figure~\ref{fig:human_annotation}) to ensure high standards of quality, difficulty, and cultural authenticity. Two groups of locally-situated experts drove this process: \curators{} and \auditors{}. Broadly, \curators{} authored  question-answer-reasoning (QAR) data, and \auditors{} verified the data while providing independent feedback for \curators{} to improve. Each locale's group averaged five experts, with the paper's authors providing guidance throughout.

\textbf{Culture-Specific Video Selection.} The process began by defining a cultural taxonomy. For each locale, \auditors{} expanded the six broad domains into fine-grained, locally-recognized sub-categories (e.g., specific regional festivals in Japan). Guided by this taxonomy, we mined a large pool of YouTube videos, which \auditors{} then manually screened against a strict checklist. Candidate videos had to be primarily in the native language, contain meaningful audio-visual content, depict authentic cultural scenes, exceed one minute in length, and be sufficiently complex to support at least 4--5 distinct multi-step reasoning questions. These guidelines ensured a consistent level of video quality across all locales (see \suppmat{}~\ref{supp_sec:guidelines} for details).

\textbf{Calibration Phase.} Before large-scale annotation, all experts were aligned in a calibration phase using a 10\% data sample to standardize difficulty and correctness. This stage involved two parallel review processes. First, \textbf{Hardness Calibration} trained human experts to create questions that were challenging for LLMs. Both \curators{} and \auditors{} curated QAR sets and received feedback from the authors, teaching both groups about what hard questions should look like. This process identified unsatisfactory examples on which authors provided feedback for experts to improve their curated examples. Through this, authors explained to experts that harder questions should not be solvable from a single frame, by audio alone, or with general knowledge, and must be grounded in the video's cultural elements. For this, the authors performed a manual inspection of the calibration data, used translation tools and VLMs to understand multilingual text and identify potentially easy examples before providing feedback to human experts. Second, \textbf{Correctness Calibration} ensured that answers were \textit{objective and unambiguous}. Independent \auditors{} answered questions without access to the ground truth; disagreements with \curators{}'s
answers triggered a dialogue-driven revision process until consensus was reached. Questions lacking consensus were discarded.

\textbf{Final Curation and Audit.} Once calibrated, \curators{} generated the final QAR set following a stratified procedure for balanced cultural coverage. To maximize dataset quality while managing the significant human curation cost, we implemented a \emph{continuous audit strategy}. \auditors{} performed rigorous reviews and provided feedback in stages, with denser feedback during the initial collection phases to rectify systemic curation errors early. This iterative process, which reviewed 50\% of the total data, also helped \curators{} improve at generating examples as collection progressed. \auditors{} performed two more quality assurance tasks: (i) independently answering 50\% of questions to resolve any ambiguity and (ii) verifying that each question was culturally grounded and demanded multi-frame temporal understanding. See \suppmat{}~\ref{supp_sec:guidelines} for annotator guidelines;  \suppmat{}~\ref{supp_sec:annotator_recruitment} for recruitment and compensation.
\definecolor{ultralightgray}{RGB}{245,245,245}  % extremely light

\begin{table*}[t]
\centering
\caption{\textbf{A Comparative Evaluation of Model Performance Across 18 Locales.} The table benchmarks several prominent models, including variants of QWEN, Claude, GPT, and Gemini. While Gemini-2.5-Pro achieves the highest scores in most locales, all models remain significantly below the human performance standard. \texttt{Aggregate} refers to the weighted average across all the locales.}
\label{tab:model-performance}
\resizebox{\textwidth}{!}{%
\label{tab:main-table}
\begin{tabular}{lccccccc>{\columncolor{ultralightgray}}c}
\toprule
\textbf{Locale} & \textbf{QWEN-2.5-VL} & \textbf{QWEN-3-VL} & \textbf{Claude-Sonnet-4} & \textbf{GPT-5-mini} & \textbf{Gemini-2.5-Flash} & \textbf{GPT-5} & \textbf{Gemini-2.5-Pro} & \textbf{Human Performance} \\
\midrule
\texttt{ar-EG} & 15.12 & 23.90 & 25.61 & 36.59 & 35.12 & 43.17 & \textbf{47.80} & 98.05 \\
\texttt{de-DE} & 10.76 & 22.15 & 29.75 & 46.20 & 51.90 & 48.10 & \textbf{48.73} & 90.51 \\
\texttt{en-GB} & 25.70 & 34.58 & 29.91 & 49.07 & 43.46 & 50.00 & \textbf{54.21} & 94.86 \\
\texttt{en-IN} & 10.55 & 24.77 & 23.85 & 44.04 & 37.61 & 45.41 & \textbf{47.71} & 94.95 \\
\texttt{es-MX} & 16.07 & 26.07 & 26.07 & 40.36 & 40.00 & 44.29 & \textbf{50.00} & 91.07 \\
\texttt{fr-FR} & 11.11 & 22.65 & 26.07 & 40.60 & 37.18 & 52.14 & \textbf{52.56} & 94.87 \\
\texttt{hi-IN} & 11.82 & 23.31 & 24.32 & 31.08 & 30.74 & 39.19 & \textbf{41.89} & 97.97 \\
\texttt{id-ID} & 17.91 & 30.97 & 30.97 & 43.28 & 48.88 & \textbf{56.34} & 55.97 & 97.39 \\
\texttt{it-IT} & 16.18 & 24.71 & 28.53 & 46.76 & 47.65 & 51.18 & \textbf{51.47} & 98.24 \\
\texttt{ja-JP} & 10.16 & 19.92 & 24.39 & 39.02 & 33.74 & 45.94 & \textbf{46.75} & 91.06 \\
\texttt{ko-KR} & 17.14 & 22.86 & 23.33 & 51.90 & 48.10 & 55.71 & \textbf{64.29} & 92.38 \\
\texttt{mr-IN} & 4.89 & 18.80 & 16.54 & 28.57 & 30.08 & 36.47 & \textbf{38.72} & 96.99 \\
\texttt{pt-BR} & 13.02 & 26.04 & 22.40 & 39.58 & 27.08 & 36.98 & \textbf{43.75} & 92.71 \\
\texttt{ru-RU} & 10.96 & 14.61 & 17.70 & 29.21 & 28.37 & 33.71 & \textbf{36.24} & 96.63 \\
\texttt{ta-IN} & 3.60 & 14.00 & 15.60 & 16.40 & 20.00 & 26.40 & \textbf{31.60} & 95.20 \\
\texttt{te-IN} & 9.20 & 12.40 & 14.40 & 24.00 & 27.20 & 23.60 & \textbf{28.00} & 93.20 \\
\texttt{th-TH} & 7.10 & 17.74 & 19.03 & 30.32 & 32.58 & 38.39 & \textbf{39.03} & 95.16 \\
\texttt{zh-TW} & 18.52 & 28.89 & 24.81 & 34.07 & 32.96 & 38.52 & \textbf{40.00} & 95.93 \\
\midrule
\texttt{Aggregate} & 12.75 & 21.50 & 23.36 & 36.64 & 35.84 & 42.20 & \textbf{45.07} & 95.22 \\
\bottomrule
\end{tabular}
}
\vspace{-0.35cm}
\end{table*}
\section{Evaluation and Analysis}
\label{sec:expts}
We evaluate open and proprietary Video-LLMs on \benchmark{}, alongside a human baseline. Given the open-ended nature of the questions in \benchmark{}, standard string matching is an inadequate evaluation method. Therefore, we employ an \emph{LLM Judge}, using \geminiflash{} to score each response on a three-point scale {0, 1, 2} based on its semantic alignment with the ground truth. This approach is suitable because we try to ensure during annotation that the answers are objective and unambiguous (\eg numerical values, object recognition), making automated assessment reliable. To maintain consistency, this same metric was used for both model and human evaluations. We present results across all 18 locales, and the full evaluation prompt for the \emph{LLM Judge} is in the \suppmat{}~\ref{supp_sec:prompts}.

\textbf{Models.} We evaluate two open Video-LLMs: \qwentwovl{}~\citep{bai2025qwen25vltechnicalreport} and \qwenthreevl{}~\citep{qwen3techreport}; and five closed models \claude{}~\citep{claude4}, \gptmini{}~\citep{gpt5systemcard}, \gpt{}~\citep{gpt5systemcard}, \geminiflash{}~\citep{gemini2.5} and \geminipro{}~\citep{gemini2.5}. Frame sampling and hyperparameter details are in the \suppmat{}~\ref{supp_sec:analysis}.

\subsection{Human Evaluation Protocol}
\label{sec:human_eval}
To establish a robust human performance baseline, we recruited a new pool of locally-situated human evaluators who did not have access to the ground-truth answers and reasoning traces used in the dataset's creation. Their task was to watch the associated video for each question and provide a short, objective answer. The evaluation protocol permitted the use of open-web resources (e.g., web search) for grounding unfamiliar cultural entities but strictly prohibited the use of any large language models (LLMs). This strict protocol ensures an unbiased, purely human-generated performance metric for fair model comparison.

\subsection{Model Performance Analysis}
\label{sec:model_performance}
Our primary results, presented in Table~\ref{tab:main-table}, reveal a substantial performance gap between the human baseline (95.22\%) and the top-performing model, \geminipro{} (45.07\%). This gap underscores the core challenge of our benchmark and highlights the limitations of current models in complex multicultural video reasoning. Among the models, a distinct performance hierarchy emerges, consistent with architectural scale and with \geminipro{} and \gpt{} leading. 

A critical finding from our analysis is the pronounced \textbf{cultural disparity in video understanding} exhibited by current models. We observe significant performance variance across locales, indicating a strong dependency on cultural and linguistic factors. While models perform best in locales such as Korean (\texttt{ko-KR}) and British English (\texttt{en-GB}), they struggle profoundly in others. This disparity is particularly stark for South Indian languages, where the top model's accuracy drops to just 28.00\% for Telugu (\texttt{te-IN}) and 31.60\% for Tamil (\texttt{ta-IN}), providing clear evidence of the cultural biases our benchmark is designed to expose. This disparity suggests that model capabilities are unevenly distributed and may correlate with the prevalence of certain languages and cultural contexts within pre-training corpora. The models' struggles in these underrepresented settings provide empirical evidence of the Western and English-centric bias that our benchmark is designed to measure.
\begin{figure}[t]
    \centering
\includegraphics[width=0.8\columnwidth]{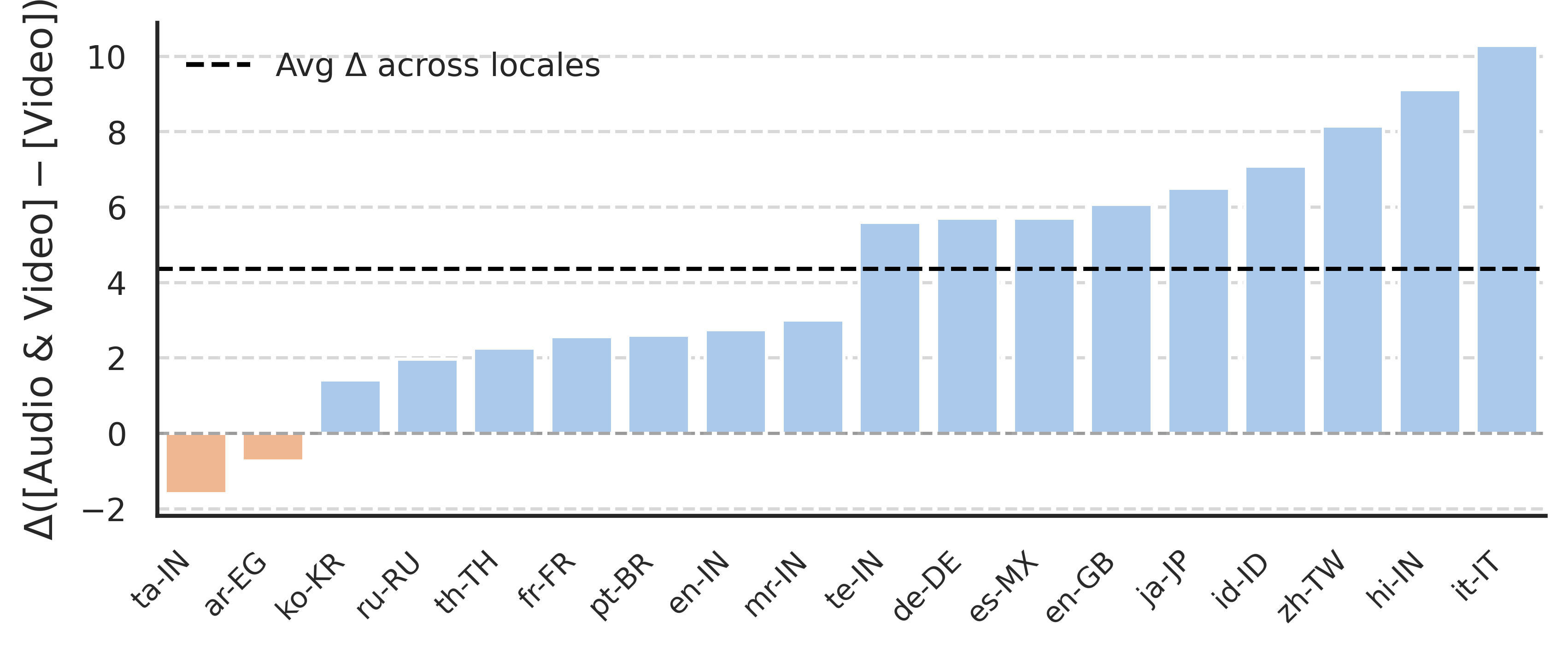}
    \caption{Performance improvement when using both Audio and Video vs. Video only, on Gemini-2.5-pro. We see that adding audio consistently improves performance across most locales.}
    \label{fig:audiovideo-performance}
\end{figure}
We perform several analyses to understand performance differences, as shown below. We evaluate \geminipro{} on a diverse subset of six locales consisting of both relatively high-resource and low-resource languages~\citep{bapna2022building}: Spanish (\texttt{es-MX}), Japanese (\texttt{ja-JP}), British English (\texttt{en-GB}), Russian (\texttt{ru-RU}), Tamil (\texttt{ta-IN}), and Arabic (\texttt{ar-EG}).

\textbf{Importance of Audio.} To quantify the importance of audio, we compare performance on full audio-visual inputs against a video-only baseline (Figure~\ref{fig:audiovideo-performance}). Including audio yields a substantial average performance increase of 4.32\%, with gains being particularly notable for locales like Chinese (\texttt{zh-TW}, +8.15\%) and Indonesian (\texttt{id-ID}, +7.09\%). This finding demonstrates that the audio track contains critical, non-redundant information like native-language dialogue and cultural-specific audio artifacts, confirming that success on \benchmark{} requires holistic multimodal understanding.

\begin{figure}[ht]

\centering
\includegraphics[width=0.8\textwidth]{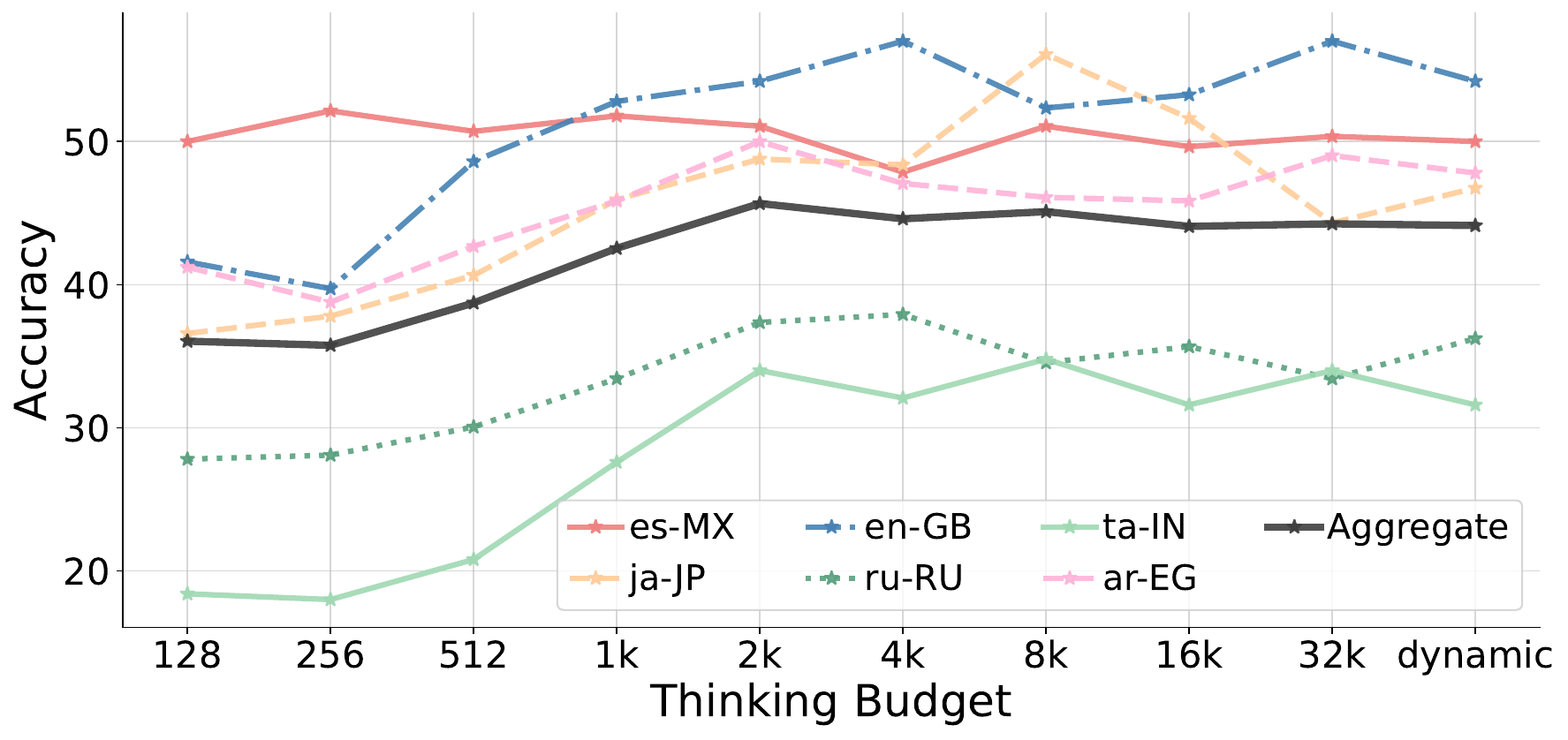}
\caption{Effect of increasing output token budget on Gemini-2.5-Pro's accuracy. Performance scales positively with compute, peaking at a 2k token budget followed by diminishing returns.}
\label{fig:thinking_ablation}
\end{figure}

\textbf{Effect of Scaling the Thinking Budget.} We analyze the effect of test-time compute on performance across locales by varying the model's \emph{thinking budget}, defined as the number of tokens for intermediate reasoning, before generating the final answer. As shown in Figure~\ref{fig:thinking_ablation}, average performance across 6 locales increases from 35.9\% at 128 tokens to a peak of 45.9\% at 2k tokens, beyond which the accuracy plateaus. This shows that while increasing the thinking budget can lead to modest performance improvement on \benchmark{}, the gains saturate quickly and performance still lags human performance by a large margin as seen in Table~\ref{tab:main-table}. In \texttt{es-MX}, the questions appear to require less detailed thinking than the others.
See Section~\ref{sec:error_tagging} for an analysis of error types that remain a challenge even after scaling inference compute.

\textbf{Temporal Understanding Complexity.} We evaluated performance with a varying number of input frames (1 to 512) to assess \benchmark{}'s temporal complexity (see graph in \suppmat{}~\ref{supp_sec:analysis}). The accuracy monotonically increases with more frames, confirming that our tasks require temporal reasoning and cannot be solved from static images alone. However, performance gains diminish at higher frame counts and a large gap to the human baseline persists. This suggests that while sufficient visual sampling is necessary, the primary performance bottleneck is not a lack of visual information but rather the higher-level, culturally-contextualized reasoning that \benchmark{} demands.

\begin{figure}[t]
\centering
\includegraphics[width=0.8\textwidth]{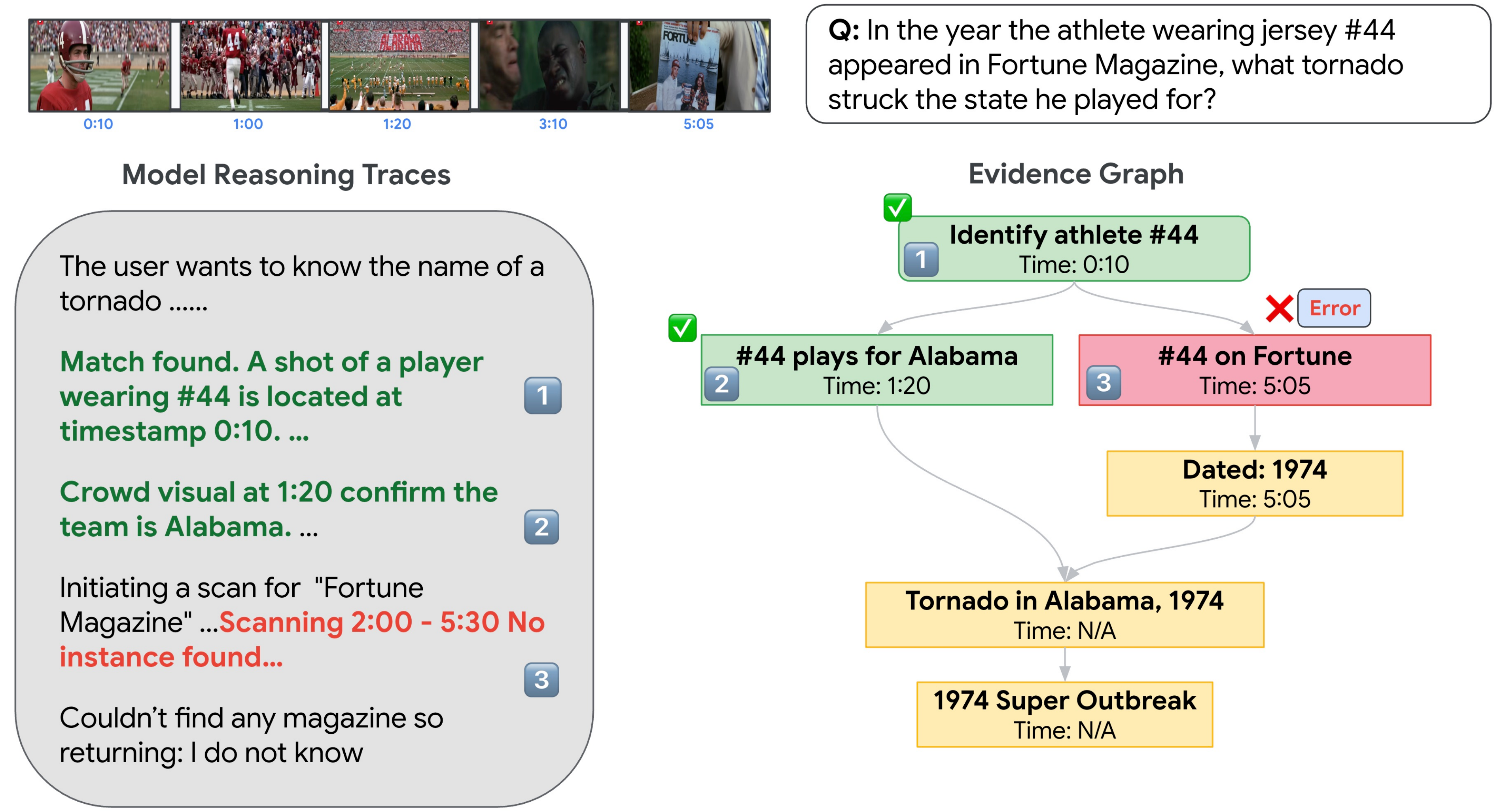}
\caption{First iteration of error isolation and tagging. The \emph{Evidence Graph} (right) compares data against model reasoning. Green nodes indicate matched evidence; red nodes are missing and are tagged; yellow nodes represent unevaluated nodes. For the subsequent iteration, the graph is pruned to the yellow nodes, and the model uses the green and corrected red nodes as \emph{hints}.}
\label{fig:error_tagging}
\vspace{-3mm}
\end{figure}

\begin{figure*}[ht]
\centering
\includegraphics[width=0.95\textwidth]{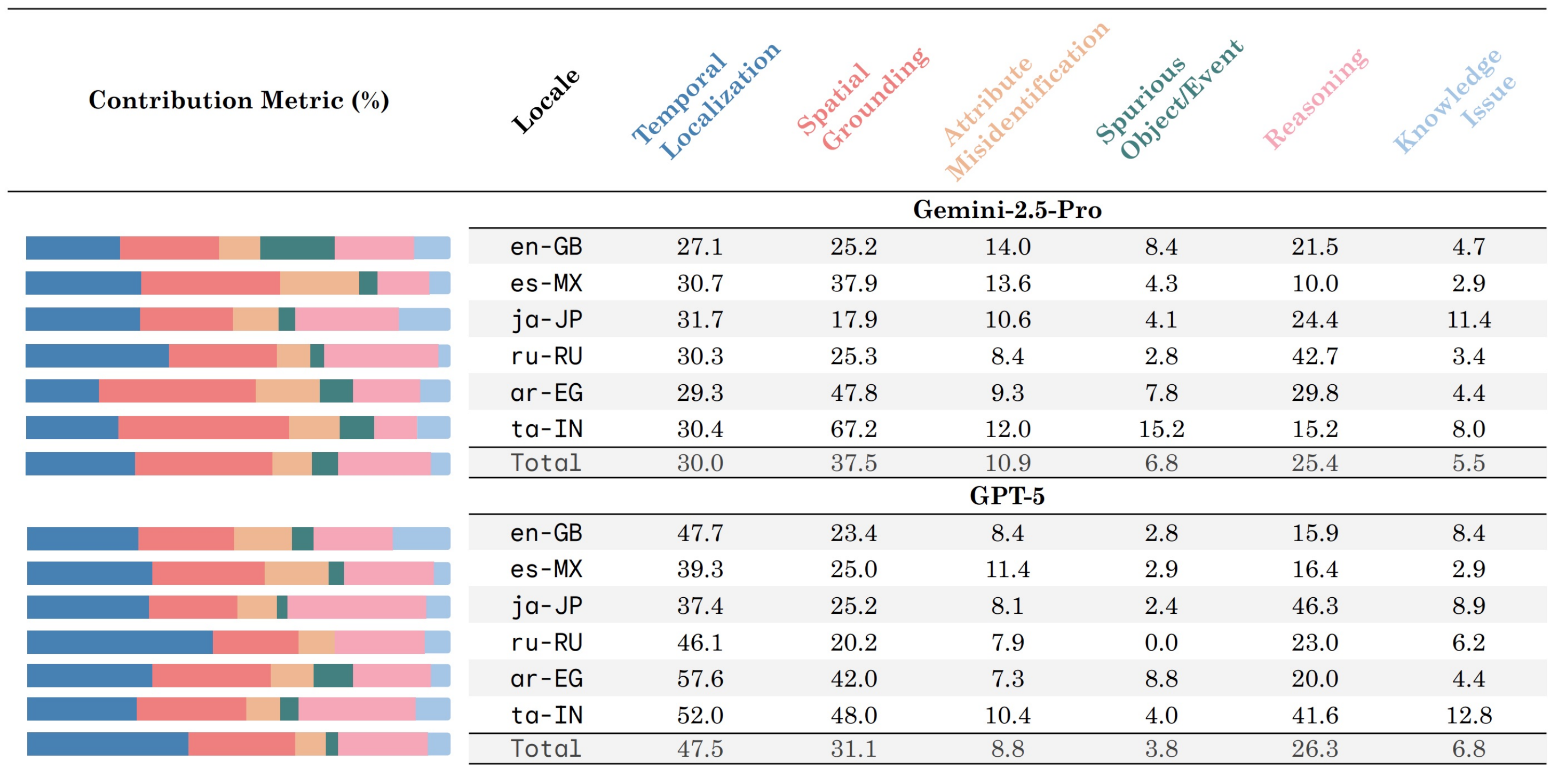}
\caption{\textbf{(Right)} We report the frequency of Error Types per 100 questions across all locales and models. \textbf{(Left)} We visualize the normalized impact of each error type per question as a bar plot. Lower$\downarrow$ is better. Best viewed in color.}
\label{fig:error_isolation}
\vspace{-0.3cm}
\end{figure*}

\section{Error Tagging using Evidence Graphs}
\label{sec:error_tagging}
Questions in \benchmark{} demand complex, multi-step reasoning, requiring models to plan its steps, gather evidence, and integrate their findings.
This complexity introduces diverse failure modes, ranging from errors in the visual perception of a cultural artifact to flaws in the logical reasoning.
Consequently, the simple accuracy metric in Table~\ref{tab:main-table} is too coarse-grained, as it cannot pinpoint precisely \textit{where and why} a model fails. Hence, we introduce a diagnostic methodology built on two core components.
The first is the \emph{Evidence Graph}, a Directed Acyclic Graph that formalizes human reasoning.
The nodes represent \emph{atomic evidences}: single, critical pieces of information required for the final answer.
The edges represent prerequisite dependencies, where obtaining one evidence is required to find its children.

The second component is \emph{Iterative Error Isolation}, a process designed to exhaustively analyze the model performance on all critical steps required to answer the question.
By traversing the \emph{Evidence Graph}, we first identify the model's initial point(s) of failure, or root error(s) (see Figure~\ref{fig:error_tagging}).
Once identified, these point(s) of failures are tagged using a carefully curated taxonomy.
Additionally, to enable a comprehensive diagnosis beyond this first mistake, we then adopt a counterfactual approach of providing the model with a corrective hint for the isolated error and re-initiate the evaluation.
This loop of traversal, error isolation, and evaluation with corrective hinting continues until the entire reasoning chain is successfully completed, allowing us to uncover all potential failures.
We detail the stages of this pipeline in the following sections and Algorithm~\ref{alg:error_isolation}.

\begin{algorithm}[t]
\caption{Iterative Error Isolation}
\label{alg:error_isolation}
\begin{algorithmic}[1]
\REQUIRE Model $M$,Video $V$,Question $ques$, HumanTrace $T_h$
\ENSURE ErrorLog $E$
\STATE $G \leftarrow \textsc{Build-Graph}(T_h)$
\STATE $hints \leftarrow \emptyset$, $E \leftarrow \emptyset$
\WHILE{$G$ is not empty}
    \STATE $T_m \leftarrow \textsc{Eval-Model}(M, V, ques, hints)$
    \STATE $N_{\text{visited}} \leftarrow \emptyset$
    % \STATE $Q \leftarrow \textsc{Get-Graph-Roots}(G)$
    \STATE $Q \leftarrow \{ n \in G.\text{Nodes} \mid n.\text{parents} = \emptyset \}$
    \WHILE{$Q$ is not empty}
        \STATE $n \leftarrow \textsc{DEQUEUE}(Q)$
        \STATE $N_{\text{visited}} \leftarrow N_{\text{visited}} \cup \{n\}$
        \IF{\textsc{Evidence-Found}($n, T_m$)}
            \STATE \textsc{ENQUEUE}($Q, n.\text{children}$)
        \ELSE
            \STATE $\mathit{err} \leftarrow \textsc{TAG-ERROR}(n, T_m)$
            \STATE $E \leftarrow E \cup \{(n, \mathit{err})\}$
        \ENDIF
    \ENDWHILE
    \STATE $hints \leftarrow hints \cup \textsc{Generate-Hints}(N_{\text{visited}})$
    \STATE $G \leftarrow \textsc{Prune-Graph}(G, N_{\text{visited}})$
\ENDWHILE
\STATE \RETURN $E$
\end{algorithmic}
\end{algorithm}
\subsection{Formalizing Reasoning into Evidence Graphs}
We employ a prompted LLM to convert the raw human reasoning traces from unstructured text into a formal \emph{Evidence Graph} (Algorithm~\ref{alg:error_isolation}, L1).
The prompt instructs the LLM to decompose the trace into nodes representing \emph{atomic evidence} (Refer Figure~\ref{fig:prompt_graph}).
We define \emph{atomic evidence} as a single piece of information derived from one of the three sources: (1) a visual observation from a specific video timestamp, (2) a fact retrieved from external knowledge, (3) a logical inference derived from previous evidence.
The LLM then establishes directed edges (prerequisite conditions) between evidences by determining whether an error in the source node would prevent the derivation of the targets.
This automated process transforms the raw human text into a ground-truth \emph{Evidence Graph} that serves as the basis for our error analysis.

\subsection{Iterative Error Isolation}
\label{iterative_error_isolation}
Once the \emph{Evidence Graph} is built, we evaluate it iteratively until the problem is correctly solved.
For each iteration, the model receives the video, the question and any corrective hints from prior steps and the reasoning trace is derived. 
To isolate errors, we follow a three-stage loop detailed below:

\textbf{Traversal:} We prompt an LLM to traverse the \textit{Evidence Graph}, comparing the \textit{atomic evidence} at each node to the model's reasoning. This traversal halts along any path when an expected evidence is missing (Algorithm~\ref{alg:error_isolation}, L3-11).

\textbf{Error Isolation and Tagging:} Once the LLM encounters a failure to map an evidence, it determines if it is a \textit{Divergence} or an \textit{Error}. A \textit{Divergence} occurs when the model initiates a valid, alternative reasoning path not present in the human trace. While it represents a locally correct alternative solution, it cannot be evaluated against our human-derived graph, so the analysis of that path concludes. \textit{Example: Identifying a city via a signboard instead of the landmark used in the human trace.} 
However, we find that true divergences are rare (2\% of questions), as our complex video-based questions often depend on a specific set of visual evidence that allows for fewer alternative paths. Conversely, it's an \textit{Error} if the model does not diverge but fails to produce the required evidence. The LLM tags each error using a detailed taxonomy that distinguishes between failures in perception (Temporal Localization, Spatial Grounding, Attribute Misidentification, Spurious Objects/Events), Knowledge, and Reasoning~(Algorithm~\ref{alg:error_isolation}, L13). This taxonomy builds upon the classification in~\citep{minerva}, which we refine with more granular perception categories based on manual examination. For the complete taxonomy and examples, see Table~\ref{tab:error_taxonomy} in \suppmat{}~\ref{supp_sec:error_tagging}.

\textbf{Hint Generation and Re-evaluation:} The iteration concludes after tagging all reachable nodes. The LLM generates a \textit{corrective hint} that consists of corrections for the failed nodes and the correct evidences. Finally, the \textit{Evidence Graph} is pruned, leaving only unevaluated nodes for the next iteration. Then, the model is re-queried with the original inputs, the correctly gathered evidence plus this new hint, and compared against the pruned \textit{Evidence Graph} (Algorithm~\ref{alg:error_isolation}, L4, L17--18). The prompt for error isolation is provided in Figures~\ref{fig:tag_graph_1},~\ref{fig:tag_graph_2}.

\subsection{Analysis}
\label{error_tagging_analysis}
\paragraph{Experimental Setup.} We apply our diagnostic pipeline to frontier multimodal models: \geminipro{} and \gpt{}.
We analyze the same locales as in Section~\ref{sec:model_performance}, limiting to the set of questions that was scored 0 by the \emph{LLM Judge}.
This comprises 490 and 524 responses out of 878 total questions for \geminipro{} and \gpt{}, respectively.
The entire pipeline, including error tagging and hint generation, uses \geminipro{} as our prompted LLM, chosen for its strong instruction-following and reasoning capabilities.

\emph{Metrics:} We accumulate the total number of errors of each type and report the frequency as errors per 100 questions to account for varying number of questions.
To quantify the individual contribution, we calculate an aggregate score for each error type by summing its fractional contributions from every question in the dataset. 
This ensures that each question contributes equally to the final distribution, regardless of the number of errors it contains.

\textbf{Structural Analysis.} We first characterize \benchmark{} using its evidence graphs.
Questions in the benchmark require an average of 5.0 atomic evidences to solve, and over 63\% of these evidences grounded in specific video timestamps, highlighting the need for detailed temporal analysis.
Furthermore, the graph depth statistics ($\mu = 2.5$ and $\sigma = 1.3$) indicate a mix of problems requiring independent evidence collection (low depth) and sequential reasoning chains (high depth). More stats are presented in \suppmat{}~\ref{supp_sec:error_tagging}.

\textbf{Results.} Our analysis reveals a clear pattern in the failure modes of current models.
As shown in Figure~\ref{fig:error_isolation}, the combination of Temporal Localization, Spatial Grounding, Spurious Object/Event, and Attribute Misidentification errors accounts for approximately 75\% of all failures.
We collectively term these as \textbf{cultural visual perception} errors and observe that they are significantly more prevalent than \textbf{reasoning} errors.
While comparing models, we find distinct profiles. \geminipro{} commits around 7\% fewer cultural perception errors than \gpt{}, significantly outperforming it.
This is despite a slight tendency to overcount, as evidenced by making more Spurious Object/Event errors.
While overall number of errors in reasoning is nearly the same between the models, there are large differences in individual locales.
Lastly, we observe that relatively low-resource languages (\texttt{ar-EG}, \texttt{ta-IN}) suffer from 1.4 times more cultural visual perception errors compared to high-resource languages (\texttt{en-GB}, \texttt{ja-JP}, \texttt{es-MX}, \texttt{ru-RU}).
This highlights that identifying culturally diverse and underrepresented objects still remains a concern.
Additionally, we note that \textit{Iterative Error Isolation} is crucial for a complete picture.
By running up to five iterations (which solves 99.7\% of questions), we uncovered approximately 22\% of Gemini-2.5-Pro's total errors beyond the first iteration.
This iterative method was most effective at revealing latent reasoning failures, identifying 78 additional reasoning errors that would have otherwise been masked by earlier perception failures (see \suppmat{}~\ref{supp_sec:error_tagging} for additional discussion).

% \clearpage

\paragraph{Are all perception failures cultural?} In \benchmark{}, perception and culture are tightly entangled rather than cleanly separable.
Most of the common failure types in our taxonomy such as temporal localization, spatial grounding, attribute errors, arise around culture-specific objects and events (e.g., garments, rituals, signage).
During annotation, we enforce a mandatory ``visual cultural understanding'' skill for every question, grounding all error categories in cultural contexts.
However, we acknowledge that some errors may stem from general visual perception limitations rather than cultural gaps per se.
Since \benchmark{} deliberately situates every task in a concrete cultural context, we posit that most low-level perceptual errors occur on culturally inflected content, though disentangling the two remains an open challenge. 

\textbf{Limitations of the LLM-based Error Tagging.} Our diagnostic pipeline relies on \geminipro{} for error classification and hint generation. This enables automated analysis at a large scale setting. There is growing interest in using LLMs for automatic evaluation, and recent works like ~\cite{evaltree} show how they can generate weakness profiles for a model on a benchmark. However, we do acknowledge the potential for inherent model biases. To mitigate this risk, we took several steps to ensure robustness: (1) we designed a highly structured task with clear, fine-grained error definitions; (2) we employed strong prompting techniques with examples to ground the model's classifications; and (3) we used a majority voting mechanism (over three queries) for each classification decision to improve reliability and reduce stochasticity (more details in \suppmat{}~\ref{supp_sec:error_tagging}). 
Despite this, we consider the development of more robust, LLM-based diagnostic tools or human-in-the-loop verification of LLM-judged errors as an important direction for future work.

\section{Conclusion and Future Opportunities}
We introduced \textbf{\benchmark{}}, a novel benchmark for globally-aware video reasoning. We recognize that the manual curation of culturally authentic data is resource-intensive; by undertaking this foundational work and releasing it publicly, we provide an asset that allows the community to innovate without the prohibitive initial cost. While our 18 locales are not exhaustive, the \benchmark{} methodology offers a robust framework for future expansion. Moreover, the inclusion of detailed human reasoning traces, a novel feature at this scale, presents a unique opportunity for more interpretable, process-oriented models. Finally, our transparent, multi-expert auditing process serves as a blueprint for navigating cultural subjectivity. It underscores that building fair and globally competent ML systems requires a deep investment in rigorous, human-centered curation.
\section{Acknowledgements}

We thank Lisa Anne Hendricks, Laura Rimell, Sunipa Dev, Utkarsh Lath, Mansi Gupta, Aditay Tripathi, Nithish Kannen, Purvam Jain, Vivek Dani, Sagar Gubbi and Partha Talukdar for their valuable feedback on the manuscript; Aditay Tripathi and Partha Talukdar for discussions on error attribution approach; Antoine Yang for helpful discussion on evaluation infrastructure; and Shankar K, Sudhindra Kopalle, Maura O'Brien and Eric Baum for facilitating the human data collection work. We are grateful to Partha Talukdar, Manish Gupta and Rahul Sukthankar for their continuous support.

\clearpage
\newpage

\bibliography{references}

@STRING{CVPR="Conference on Computer Vision and Pattern Recognition (CVPR)"}

@STRING{ICCV="International Conference on Computer Vision (ICCV)"}

@STRING{ECCV="European Conference on Computer Vision (ECCV)"}

@STRING{NeurIPS="Advances in Neural Information Processing Systems (NeurIPS)"}

@STRING{CVPRW="Conference on Computer Vision and Pattern Recognition Workshops (CVPRW)"}

@STRING{NeurIPSData="Advances in Neural Information Processing Systems (NeurIPS): Track on Datasets and Benchmarks"}

@STRING{ACL="Association of Computational Linguistics (ACL)"}

@STRING{ACLFINDINGS="Findings of the Association for Computational Linguistics (ACL)"}

@STRING{EMNLP="Empirical Methods in Natural Language Processing (EMNLP)"}

@STRING{EMNLPFINDINGS="Findings of Empirical Methods in Natural Language Processing (EMNLP)"}

@STRING{COLM="Conference On Language Modeling (COLM)"}

@STRING{CHI="ACM Conference on Human Factors in Computing Systems (CHI)"}

@InProceedings{Fu_2025_CVPR,
    author    = {Fu, Chaoyou and Dai, Yuhan and Luo, Yongdong and Li, Lei and Ren, Shuhuai and Zhang, Renrui and Wang, Zihan and Zhou, Chenyu and Shen, Yunhang and Zhang, Mengdan and Chen, Peixian and Li, Yanwei and Lin, Shaohui and Zhao, Sirui and Li, Ke and Xu, Tong and Zheng, Xiawu and Chen, Enhong and Shan, Caifeng and He, Ran and Sun, Xing},
    title     = {{Video-MME: The First-Ever Comprehensive Evaluation Benchmark of Multi-modal LLMs in Video Analysis}},
    booktitle = CVPR,
    month     = {June},
    year      = {2025},
    pages     = {24108-24118}
}

@article{deutsch2022limitations,
  title={On the limitations of reference-free evaluations of generated text},
  author={Deutsch, Daniel and Dror, Rotem and Roth, Dan},
  journal={arXiv preprint arXiv:2210.12563},
  year={2022}
}

@article{gemini2.5,
  title={{Gemini 2.5: Pushing the Frontier with Advanced Reasoning, Multimodality, Long Context, and Next Generation Agentic Capabilities}},
  author={Comanici, Gheorghe and Bieber, Eric and Schaekermann, Mike and Pasupat, Ice and Sachdeva, Noveen and Dhillon, Inderjit and Blistein, Marcel and Ram, Ori and Zhang, Dan and Rosen, Evan and others},
  journal={arXiv preprint arXiv:2507.06261},
  year={2025}
}

@InProceedings{mvbench,
    author    = {Li, Kunchang and Wang, Yali and He, Yinan and Li, Yizhuo and Wang, Yi and Liu, Yi and Wang, Zun and Xu, Jilan and Chen, Guo and Luo, Ping and Wang, Limin and Qiao, Yu},
    title     = {{MVBench: A Comprehensive Multi-modal Video Understanding Benchmark}},
    booktitle = CVPR,
    month     = {June},
    year      = {2024},
    pages     = {22195-22206}
}

@article{nagrani2024neptune,
  title={{Neptune: The Long Orbit to Benchmarking Long Video Understanding}},
  author={Nagrani, Arsha and Zhang, Mingda and Mehran, Ramin and Hornung, Rachel and Gundavarapu, Nitesh Bharadwaj and Jha, Nilpa and Myers, Austin and Zhou, Xingyi and Gong, Boqing and Schmid, Cordelia and others},
  journal={arXiv preprint arXiv:2412.09582},
  year={2024}
}

@inproceedings{mangalam2023egoschema,
  title={{EgoSchema: A Diagnostic Benchmark for Very Long-form Video Language Understanding}},
  author={Mangalam, Karttikeya and Akshulakov, Raiymbek and Malik, Jitendra},
  booktitle=NeurIPSData,
  year={2023}
}

@article{li2024videovista,
  title={VideoVista: A Versatile Benchmark for Video Understanding and Reasoning},
  author={Li, Yunxin and Chen, Xinyu and Hu, Baotian and Wang, Longyue and Shi, Haoyuan and Zhang, Min},
  journal={arXiv preprint arXiv:2406.11303},
  year={2024}
}

@inproceedings{wu2024longvideobench,
    title = {{LongVideoBench: A Benchmark for Long-context Interleaved Video-Language Understanding}},
    author = {Wu, Haoning and Li, Dongxu and Chen, Bei and Li, Junnan},
    booktitle = NeurIPS,
    year = {2024},
    pages = {28828--28857}
}

@InProceedings{zhou2024mlvu,
    author    = {Zhou, Junjie and Shu, Yan and Zhao, Bo and Wu, Boya and Liang, Zhengyang and Xiao, Shitao and Qin, Minghao and Yang, Xi and Xiong, Yongping and Zhang, Bo and Huang, Tiejun and Liu, Zheng},
    title     = {{MLVU: Benchmarking Multi-task Long Video Understanding}},
    booktitle = CVPR,
    month     = {June},
    year      = {2025},
    pages     = {13691-13701}
}

@article{reid2024gemini,
  title={Gemini 1.5: Unlocking multimodal understanding across millions of tokens of context},
  author={Reid, Machel and Savinov, Nikolay and Teplyashin, Denis and Lepikhin, Dmitry and Lillicrap, Timothy and Alayrac, Jean-baptiste and Soricut, Radu and Lazaridou, Angeliki and Firat, Orhan and Schrittwieser, Julian and others},
  journal={arXiv preprint arXiv:2403.05530},
  year={2024}
}

@article{bai2025qwen2,
  title={Qwen2.5-VL Technical Report},
  author={Bai, Shuai and Chen, Keqin and Liu, Xuejing and Wang, Jialin and Ge, Wenbin and Song, Sibo and Dang, Kai and Wang, Peng and Wang, Shijie and Tang, Jun and others},
  journal={arXiv preprint arXiv:2502.13923},
  year={2025}
}

@misc{claude4,
  title={{System Card: Claude Opus 4 \& Claude Sonnet 4}},
  author={Anthropic},
  year={2025},
  month={May},
  howpublished={\url{https://www-cdn.anthropic.com/4263b940cabb546aa0e3283f35b686f4f3b2ff47.pdf}},
}

@article{jaech2024openai,
  title={Openai o1 system card},
  author={Jaech, Aaron and Kalai, Adam and Lerer, Adam and Richardson, Adam and El-Kishky, Ahmed and Low, Aiden and Helyar, Alec and Madry, Aleksander and Beutel, Alex and Carney, Alex and others},
  journal={arXiv preprint arXiv:2412.16720},
  year={2024}
}

@article{gpt5systemcard,
  title={{GPT-5 System Card}},
  author={{OpenAI}},
  journal={arXiv preprint arXiv:2601.03267},
  year={2026}
}

@misc{yu2025vrbenchbenchmarkmultistepreasoning,
      title={VRBench: A Benchmark for Multi-Step Reasoning in Long Narrative Videos}, 
      author={Jiashuo Yu and Yue Wu and Meng Chu and Zhifei Ren and Zizheng Huang and Pei Chu and Ruijie Zhang and Yinan He and Qirui Li and Songze Li and Zhenxiang Li and Zhongying Tu and Conghui He and Yu Qiao and Yali Wang and Yi Wang and Limin Wang},
      year={2025},
      eprint={2506.10857},
      archivePrefix={arXiv},
      primaryClass={cs.CV},
      url={https://arxiv.org/abs/2506.10857}, 
}

@misc{heyward2024perceptiontest2024challenge,
      title={Perception Test 2024: Challenge Summary and a Novel Hour-Long VideoQA Benchmark}, 
      author={Joseph Heyward and João Carreira and Dima Damen and Andrew Zisserman and Viorica Pătrăucean},
      year={2024},
      eprint={2411.19941},
      archivePrefix={arXiv},
      primaryClass={cs.CV},
      url={https://arxiv.org/abs/2411.19941}, 
}

@inproceedings{Nayak_2024_EMNLP,
    title     = {{Benchmarking Vision Language Models for Cultural Understanding}},
    author    = {Nayak, Shravan and Jain, Kanishk and Awal, Rabiul and Reddy, Siva and Steenkiste, Sjoerd Van and Hendricks, Lisa Anne and Stanczak, Karolina and Agrawal, Aishwarya},
    booktitle = EMNLP,
    year      = {2024},
    pages     = {5769--5790},
    doi       = {10.18653/v1/2024.emnlp-main.329}
}

@inproceedings{Khanuja_2024_EMNLP,
    title     = {{An Image Speaks a Thousand Words, but Can Everyone Listen? On Image Transcreation for Cultural Relevance}},
    author    = {Khanuja, Simran and Ramamoorthy, Sathyanarayanan and Song, Yueqi and Neubig, Graham},
    booktitle = EMNLP,
    year      = {2024},
    pages     = {10258--10279},
    doi       = {10.18653/v1/2024.emnlp-main.573}
}

@inproceedings{Das_2024_ACL,
    title     = {{EXAMS-V: A Multi-Discipline Multilingual Multimodal Exam Benchmark for Evaluating Vision Language Models}},
    author    = {Das, Rocktim and Hristov, Simeon and Li, Haonan and Dimitrov, Dimitar and Koychev, Ivan and Nakov, Preslav},
    booktitle = ACL,
    year      = {2024},
    pages     = {7768--7791},
    doi       = {10.18653/v1/2024.acl-long.420}
}

@inproceedings{Zhang_2023_NeurIPS,
    title     = {{M3Exam: A Multilingual, Multimodal, Multilevel Benchmark for Examining Large Language Models}},
    author    = {Zhang, Wenxuan and Aljunied, Mahani and Gao, Chang and Chia, Yew Ken and Bing, Lidong},
    booktitle = NeurIPS,
    year      = {2023},
    url       = {https://proceedings.neurips.cc/paper_files/paper/2023/hash/117c5c8622b0d539f74f6d1fb082a2e9-Abstract-Datasets_and_Benchmarks.html}
}

@inproceedings{Liu_2021_EMNLP,
    title     = {{Visually Grounded Reasoning Across Languages and Cultures}},
    author    = {Liu, Fangyu and Bugliarello, Emanuele and Ponti, Edoardo Maria and Reddy, Siva and Collier, Nigel and Elliott, Desmond},
    booktitle = EMNLP,
    year      = {2021},
    pages     = {10467--10485},
    doi       = {10.18653/v1/2021.emnlp-main.818}
}

@inproceedings{romero2024cvqa,
    title     = {{CVQA: Culturally-Diverse Multilingual Visual Question Answering Benchmark}},
    author    = {Romero, David and Lyu, Chenyang and Wibowo, Haryo Akbarianto and Lynn, Teresa and Hamed, Injy and Kishore, Aditya Nanda and Mandal, Aishik and Dragonetti, Alina and Abzaliev, Artem and Tonja, Atnafu Lambebo and Balcha, Bontu Fufa and Whitehouse, Chenxi and Salamea, Christian and Velasco, Dan John and Adelani, David Ifeoluwa and Le Meur, David and Villa-Cueva, Emilio and Koto, Fajri and Farooqui, Fauzan and Belcavello, Frederico and Batnasan, Ganzorig and Vallejo, Gisela and Caulfield, Grainne and Ivetta, Guido and Song, Haiyue and Ademtew, Henok Biadglign and Maina, Hern\'{a}n and Lovenia, Holy and Azime, Israel Abebe and Cruz, Jan Christian Blaise and Gala, Jay and Geng, Jiahui and Ortiz-Barajas, Jesus-German and Baek, Jinheon and Dunstan, Jocelyn and Alemany, Laura Alonso and Nagasinghe, Kumaranage Ravindu Yasas and Benotti, Luciana and D'Haro, Luis Fernando and Viridiano, Marcelo and Estecha-Garitagoitia, Marcos and Cabrera, Maria Camila Buitrago and Rodr\'{\i}guez-Cantelar, Mario and Jouitteau, M\'{e}lanie and Mihaylov, Mihail and Etori, Naome and Imam, Mohamed Fazli Mohamed and Adilazuarda, Muhammad Farid and Gochoo, Munkhjargal and Otgonbold, Munkh-Erdene and Niyomugisha, Olivier and Silva, Paula M\'{o}nica and Chitale, Pranjal and Dabre, Raj and Chevi, Rendi and Zhang, Ruochen and Diandaru, Ryandito and Cahyawijaya, Samuel and G\'{o}ngora, Santiago and Jeong, Soyeong and Purkayastha, Sukannya and Kuribayashi, Tatsuki and Clifford, Teresa and Jayakumar, Thanmay and Torrent, Tiago Timponi and Ehsan, Toqeer and Araujo, Vladimir and Kementchedjhieva, Yova and Burzo, Zara and Lim, Zheng Wei and Yong, Zheng Xin and Ignat, Oana and Nwatu, Joan and Mihalcea, Rada and Solorio, Thamar and Aji, Alham Fikri},
    booktitle = NeurIPS,
    year      = {2024},
    pages     = {11479--11505},
    doi       = {10.52202/079017-0366}
}

@inproceedings{shafique2025culturally,
    title = {{A Culturally-Diverse Multilingual Multimodal Video Benchmark {\&} Model}},
    author = {Shafique, Bhuiyan Sanjid and Vayani, Ashmal and Maaz, Muhammad and Rasheed, Hanoona Abdul and Dissanayake, Dinura and Kurpath, Mohammed Irfan and Hmaiti, Yahya and Inoue, Go and Lahoud, Jean and Rashid, Md. Safirur and others},
    booktitle = EMNLP,
    year = {2025},
    pages = {19998--20022},
    doi = {10.18653/v1/2025.emnlp-main.1012}
}

@misc{bai2025qwen25vltechnicalreport,
      title={Qwen2.5-VL Technical Report}, 
      author={Shuai Bai and Keqin Chen and Xuejing Liu and Jialin Wang and Wenbin Ge and Sibo Song and Kai Dang and Peng Wang and Shijie Wang and Jun Tang and Humen Zhong and Yuanzhi Zhu and Mingkun Yang and Zhaohai Li and Jianqiang Wan and Pengfei Wang and Wei Ding and Zheren Fu and Yiheng Xu and Jiabo Ye and Xi Zhang and Tianbao Xie and Zesen Cheng and Hang Zhang and Zhibo Yang and Haiyang Xu and Junyang Lin},
      year={2025},
      eprint={2502.13923},
      archivePrefix={arXiv},
      primaryClass={cs.CV},
      url={https://arxiv.org/abs/2502.13923}, 
}

@inproceedings{Vayani_2025_CVPR,
    title     = {{All Languages Matter: Evaluating LMMs on Culturally Diverse 100 Languages}},
    author    = {Vayani, Ashmal and Dissanayake, Dinura and Watawana, Hasindri and Ahsan, Noor and Sasikumar, Nevasini and Thawakar, Omkar and Ademtew, Henok Biadglign and Hmaiti, Yahya and Kumar, Amandeep and Kukreja, Kartik and Maslych, Mykola and Al Ghallabi, Wafa and Mihaylov, Mihail Minkov and Qin, Chao and Shaker, Abdelrahman M. and Zhang, Mike and Ihsani, Mahardika Krisna and Esplana, Amiel Gian and Gokani, Monil and Mirkin, Shachar and Singh, Harsh and Srivastava, Ashay and Hamerlik, Endre and Izzati, Fathinah Asma and Maani, Fadillah Adamsyah and Cavada, Sebastian and Chim, Jenny and Gupta, Rohit and Manjunath, Sanjay and Zhumakhanova, Kamila and Rabevohitra, Feno Heriniaina and Amirudin, Azril Hafizi and Ridzuan, Muhammad and Kareem, Daniya Najiha Abdul and More, Ketan Pravin and Li, Kunyang and Shakya, Pramesh and Saad, Muhammad and Ghasemaghaei, Amirpouya and Djanibekov, Amirbek and Azizov, Dilshod and Jankovic, Branislava and Bhatia, Naman and Cabrera, Alvaro and Obando-Ceron, Johan and Otieno, Olympiah and Farestam, Febian and Rabbani, Muztoba and Ballah, Sanoojan and Sanjeev, Santosh and Shtanchaev, Abduragim and Fatima, Maheen and Nguyen, Thao and Kareem, Amrin and Aremu, Toluwani and Xavier, Nathan Augusto Zacarias and Bhatkal, Amit and Toyin, Hawau Olamide and Chadha, Aman and Cholakkal, Hisham and Anwer, Rao Muhammad and Felsberg, Michael and Laaksonen, Jorma and Solorio, Thamar and Choudhury, Monojit and Laptev, Ivan and Shah, Mubarak and Khan, Salman and Khan, Fahad Shahbaz},
    booktitle = CVPR,
    year      = {2025},
    pages     = {19565--19575}
}

@inproceedings{pfeiffer-2022-xgqa,
author = {Pfeiffer, Jonas  and Geigle, Gregor and Kamath, Aishwarya  and Steitz, Jan-Martin O. and Roth, Stefan and Vuli{\'c}, Ivan and Gurevych, Iryna},
title = {{x{GQA}: Cross-Lingual Visual Question Answering}},
booktitle = ACLFINDINGS,
year = 2022,
pages = "2497--2511",
}

@misc{liu2025m3medbenchmarkmultilingualmultimodal,
      title={M$^3$-Med: A Benchmark for Multi-lingual, Multi-modal, and Multi-hop Reasoning in Medical Instructional Video Understanding}, 
      author={Shenxi Liu and Kan Li and Mingyang Zhao and Yuhang Tian and Bin Li and Shoujun Zhou and Hongliang Li and Fuxia Yang},
      year={2025},
      eprint={2507.04289},
      archivePrefix={arXiv},
      primaryClass={cs.CV},
      url={https://arxiv.org/abs/2507.04289}, 
}

@inproceedings{sun2024multilingual,
    title = {{Multilingual Synopses of Movie Narratives: A Dataset for Vision-Language Story Understanding}},
    author = {Sun, Yidan and Yu, Jianfei and Li, Boyang},
    booktitle = EMNLPFINDINGS,
    year = {2024},
    pages = {13488--13504}
}

@inproceedings{SOK_Bench,
author={Wang*, Andong and Wu*, Bo and Chen, Sunli and Chen, Zhenfang and Guan, Haotian and Lee, Wei-Ning and Li, Erran Li and Tenenbaum, Joshua B and Gan, Chuang},
title = {{SOK-Bench: A Situated Video Reasoning Benchmark with Aligned Open-World Knowledge}},
booktitle = CVPR,
year = {2024}
}

@InProceedings{Wang_2019_ICCV,
author = {Wang, Xin and Wu, Jiawei and Chen, Junkun and Li, Lei and Wang, Yuan-Fang and Wang, William Yang},
title = {{VaTeX: A Large-Scale, High-Quality Multilingual Dataset for Video-and-Language Research}},
booktitle = ICCV,
month = {October},
year = {2019}
}

@inproceedings{Thapliyal_2022_EMNLP,
    title     = {{Crossmodal-3600: A Massively Multilingual Multimodal Evaluation Dataset}},
    author    = {Thapliyal, Ashish V. and Pont Tuset, Jordi and Chen, Xi and Soricut, Radu},
    booktitle = EMNLP,
    year      = {2022},
    pages     = {715--729},
    doi       = {10.18653/v1/2022.emnlp-main.45}
}

@inproceedings{yue2024pangea,
  title={{Pangea: A Fully Open Multilingual Multimodal LLM for 39 Languages}},
  author={Yue, Xiang and Song, Yueqi and Asai, Akari and Kim, Seungone and de Dieu Nyandwi, Jean and Khanuja, Simran and Kantharuban, Anjali and Sutawika, Lintang and Ramamoorthy, Sathyanarayanan and Neubig, Graham},
  booktitle={The Thirteenth International Conference on Learning Representations},
  year={2024}
}

@misc{Mmmlu,
  title={Mmmlu dataset},
  author={OpenAI},
  url={https://huggingface.co/datasets/openai/MMMLU},
  year={2024}
}

@inproceedings{perception_test,
  title={{Perception Test: A Diagnostic Benchmark for Multimodal Video Models}},
  author={Patraucean, Viorica and Smaira, Lucas and Gupta, Ankush and Recasens, Adria and Markeeva, Larisa and Banarse, Dylan and Koppula, Skanda and Heyward, Joseph and Malinowski, Mateusz and Yang, Yi and Doersch, Carl and Matejovicova, Tatiana and Sulsky, Yury and Miech, Antoine and Fr\'{e}chette, Alexandre and Klimczak, Hanna and Koster, Raphael and Zhang, Junlin and Winkler, Stephanie and Aytar, Yusuf and Osindero, Simon and Damen, Dima and Zisserman, Andrew and Carreira, Joao},
  booktitle=NeurIPS,
  year={2023},
  pages={42748--42761}
}

@inproceedings{cinepile,
  title={{CinePile: A Long Video Question Answering Dataset and Benchmark}}, 
  author={Rawal, Ruchit and Saifullah, Khalid and Basri, Ronen and Jacobs, David and Somepalli, Gowthami and Goldstein, Tom},
  booktitle=CVPRW,
  year={2024},
}

@inproceedings{videocon,
  title={{VideoCon: Robust Video-Language Alignment via Contrast Captions}}, 
  author={Hritik Bansal and Yonatan Bitton and Idan Szpektor and Kai-Wei Chang and Aditya Grover},
  booktitle=CVPR,
  year={2024},
}

@inproceedings{seedbench2,
  title={{SEED-Bench-2: Benchmarking Multimodal Large Language Models}},
  author={Li, Bohao and Ge, Yuying and Ge, Yixiao and Wang, Guangzhi and Wang, Rui and Zhang, Ruimao and Shan, Ying},
  booktitle=CVPR,
  year={2024},
}

@inproceedings{evaltree,
  title={{EvalTree: Profiling Language Model Weaknesses via Hierarchical Capability Trees}},
  author={Zhiyuan Zeng and Yizhong Wang and Hannaneh Hajishirzi and Pang Wei Koh},
  booktitle=COLM,
  year={2025}
}

@inproceedings{tempcompass,
  title={{TempCompass: Do Video LLMs Really Understand Videos?}},
  author={Yuanxin Liu and Shicheng Li and Yi Liu and Yuxiang Wang and Shuhuai Ren and Lei Li and Sishuo Chen and Xu Sun and Lu Hou},
  year={2024},
  booktitle=ACLFINDINGS
}

@inproceedings{mmbenchvideo,
  title={{MMBench-Video: A Long-Form Multi-Shot Benchmark for Holistic Video Understanding}}, 
  author={Xinyu Fang and Kangrui Mao and Haodong Duan and Xiangyu Zhao and Yining Li and Dahua Lin and Kai Chen},
  year={2024},
  booktitle=NeurIPSData
}

@inproceedings{vitatecs,
  title={{VITATECS: A Diagnostic Dataset for Temporal Concept Understanding of Video-Language Models}}, 
  author={Shicheng Li and Lei Li and Shuhuai Ren and Yuanxin Liu and Yi Liu and Rundong Gao and Xu Sun and Lu Hou},
  year={2024},
  booktitle=ECCV
}

@inproceedings{velociti,
  title={{VELOCITI: Benchmarking Video-Language Compositional Reasoning with Strict Entailment}},
  author={Saravanan, Darshana and Gupta, Varun and Singh, Darshan and Khan, Zeeshan and Gandhi, Vineet and Tapaswi, Makarand},
  booktitle=CVPR,
  year={2025}
}

@article{minerva,
  title={{MINERVA: Evaluating Complex Video Reasoning}},
  author={Nagrani, Arsha and Menon, Sachit and Iscen, Ahmet and Buch, Shyamal and Mehran, Ramin and Jha, Nilpa and Hauth, Anja and Zhu, Yukun and Vondrick, Carl and Sirotenko, Mikhail and Schmid, Cordelia and Weyand, Tobias},
  journal=ICCV,
  year={2025}
}

@article{cvrres,
    title={{How Good is my Video LMM? Complex Video Reasoning and Robustness Evaluation Suite for Video-LMMs}},
    author={khattak, Muhammad Uzair and Naeem, Muhammad Ferjad and Hassan, Jameel and Muzzamal, Naseer and Tombari, Federcio and Khan, Fahad Shahbaz and Khan, Salman},
    journal={arXiv:2405.03690},
    year={2024}
}

@inproceedings{videomme,
  title={{Video-MME: The First-Ever Comprehensive Evaluation Benchmark of Multi-modal LLMs in Video Analysis}},
  author={Fu, Chaoyou and Dai, Yuhan and Luo, Yondong and Li, Lei and Ren, Shuhuai and Zhang, Renrui and Wang, Zihan and Zhou, Chenyu and Shen, Yunhang and Zhang, Mengdan and others},
  booktitle=CVPR,
  year={2025}
}

@article{vinoground,
  title={{Vinoground: Scrutinizing LMMs over Dense Temporal Reasoning with Short Videos}},
  author={Zhang, Jianrui and Mu, Cai and Lee, Yong Jae},
  journal={arXiv},
  year={2024},
  eprint={2410.02763},
  archivePrefix={arXiv},
  primaryClass={cs.CV},
  url={https://arxiv.org/abs/2410.02763}, 
}

@article{tvbench,
  title={{TVBench: Redesigning Video-Language Evaluation}},
  author={Daniel Cores and Michael Dorkenwald and Manuel Mucientes and Cees G. M. Snoek and Yuki M. Asano},
  year = {2024},
  journal = {arXiv:2410.07752},

}

@article{Trinh2024SolvingOG,
  title={Solving olympiad geometry without human demonstrations},
  author={Trieu H. Trinh and Yuhuai Wu and Quoc V. Le and He He and Thang Luong},
  journal={Nature},
  year={2024},
  volume={625},
  pages={476 - 482},
  url={https://api.semanticscholar.org/CorpusID:267032902}
}

@misc{shao2024deepseekmathpushinglimitsmathematical,
      title={DeepSeekMath: Pushing the Limits of Mathematical Reasoning in Open Language Models}, 
      author={Zhihong Shao and Peiyi Wang and Qihao Zhu and Runxin Xu and Junxiao Song and Xiao Bi and Haowei Zhang and Mingchuan Zhang and Y. K. Li and Y. Wu and Daya Guo},
      year={2024},
      eprint={2402.03300},
      archivePrefix={arXiv},
      primaryClass={cs.CL},
      url={https://arxiv.org/abs/2402.03300}, 
}

@misc{zhou2024selfdiscoverlargelanguagemodels,
      title={{Self-Discover: Large Language Models Self-Compose Reasoning Structures}}, 
      author={Pei Zhou and Jay Pujara and Xiang Ren and Xinyun Chen and Heng-Tze Cheng and Quoc V. Le and Ed H. Chi and Denny Zhou and Swaroop Mishra and Huaixiu Steven Zheng},
      year={2024},
      eprint={2402.03620},
      archivePrefix={arXiv},
      primaryClass={cs.AI},
      url={https://arxiv.org/abs/2402.03620}, 
}

@misc{zelikman2024quietstarlanguagemodelsteach,
      title={{Quiet-STaR: Language Models Can Teach Themselves to Think Before Speaking}}, 
      author={Eric Zelikman and Georges Harik and Yijia Shao and Varuna Jayasiri and Nick Haber and Noah D. Goodman},
      year={2024},
      eprint={2403.09629},
      archivePrefix={arXiv},
      primaryClass={cs.CL},
      url={https://arxiv.org/abs/2403.09629}, 
}

@misc{yang2024sweagentagentcomputerinterfacesenable,
      title={SWE-agent: Agent-Computer Interfaces Enable Automated Software Engineering}, 
      author={John Yang and Carlos E. Jimenez and Alexander Wettig and Kilian Lieret and Shunyu Yao and Karthik Narasimhan and Ofir Press},
      year={2024},
      eprint={2405.15793},
      archivePrefix={arXiv},
      primaryClass={cs.SE},
      url={https://arxiv.org/abs/2405.15793}, 
}

@misc{deepseekai2024deepseekcoderv2breakingbarrierclosedsource,
      title={DeepSeek-Coder-V2: Breaking the Barrier of Closed-Source Models in Code Intelligence}, 
      author={DeepSeek-AI and Qihao Zhu and Daya Guo and Zhihong Shao and Dejian Yang and Peiyi Wang and Runxin Xu and Y. Wu and Yukun Li and Huazuo Gao and Shirong Ma and Wangding Zeng and Xiao Bi and Zihui Gu and Hanwei Xu and Damai Dai and Kai Dong and Liyue Zhang and Yishi Piao and Zhibin Gou and Zhenda Xie and Zhewen Hao and Bingxuan Wang and Junxiao Song and Deli Chen and Xin Xie and Kang Guan and Yuxiang You and Aixin Liu and Qiushi Du and Wenjun Gao and Xuan Lu and Qinyu Chen and Yaohui Wang and Chengqi Deng and Jiashi Li and Chenggang Zhao and Chong Ruan and Fuli Luo and Wenfeng Liang},
      year={2024},
      eprint={2406.11931},
      archivePrefix={arXiv},
      primaryClass={cs.SE},
      url={https://arxiv.org/abs/2406.11931}, 
}

@misc{wang2024boostinglanguagemodelsreasoning,
      title={Boosting Language Models Reasoning with Chain-of-Knowledge Prompting}, 
      author={Jianing Wang and Qiushi Sun and Xiang Li and Ming Gao},
      year={2024},
      eprint={2306.06427},
      archivePrefix={arXiv},
      primaryClass={cs.CL},
      url={https://arxiv.org/abs/2306.06427}, 
}

@inproceedings{atanasova-etal-2023-faithfulness,
    title = "Faithfulness Tests for Natural Language Explanations",
    author = "Atanasova, Pepa  and
      Camburu, Oana-Maria  and
      Lioma, Christina  and
      Lukasiewicz, Thomas  and
      Simonsen, Jakob Grue  and
      Augenstein, Isabelle",
    editor = "Rogers, Anna  and
      Boyd-Graber, Jordan  and
      Okazaki, Naoaki",
    booktitle = "Proceedings of the 61st Annual Meeting of the Association for Computational Linguistics (Volume 2: Short Papers)",
    month = jul,
    year = "2023",
    address = "Toronto, Canada",
    publisher = "Association for Computational Linguistics",
    url = "https://aclanthology.org/2023.acl-short.25/",
    doi = "10.18653/v1/2023.acl-short.25",
    pages = "283--294"
}

@misc{huang2024largelanguagemodelsselfcorrect,
      title={Large Language Models Cannot Self-Correct Reasoning Yet}, 
      author={Jie Huang and Xinyun Chen and Swaroop Mishra and Huaixiu Steven Zheng and Adams Wei Yu and Xinying Song and Denny Zhou},
      year={2024},
      eprint={2310.01798},
      archivePrefix={arXiv},
      primaryClass={cs.CL},
      url={https://arxiv.org/abs/2310.01798}, 
}

@article{kamoi-etal-2024-llms,
    title = "When Can {LLM}s Actually Correct Their Own Mistakes? A Critical Survey of Self-Correction of {LLM}s",
    author = "Kamoi, Ryo  and
      Zhang, Yusen  and
      Zhang, Nan  and
      Han, Jiawei  and
      Zhang, Rui",
    journal = "Transactions of the Association for Computational Linguistics",
    volume = "12",
    year = "2024",
    address = "Cambridge, MA",
    publisher = "MIT Press",
    url = "https://aclanthology.org/2024.tacl-1.78/",
    doi = "10.1162/tacl_a_00713",
    pages = "1417--1440"
}

@misc{tyen2024llmsreasoningerrorscorrect,
      title={LLMs cannot find reasoning errors, but can correct them given the error location}, 
      author={Gladys Tyen and Hassan Mansoor and Victor Cărbune and Peter Chen and Tony Mak},
      year={2024},
      eprint={2311.08516},
      archivePrefix={arXiv},
      primaryClass={cs.AI},
      url={https://arxiv.org/abs/2311.08516}, 
}

@misc{lee2025evaluatingstepbystepreasoningtraces,
      title={Evaluating Step-by-step Reasoning Traces: A Survey}, 
      author={Jinu Lee and Julia Hockenmaier},
      year={2025},
      eprint={2502.12289},
      archivePrefix={arXiv},
      primaryClass={cs.CL},
      url={https://arxiv.org/abs/2502.12289}, 
}

@inproceedings{xiong-etal-2025-mapping,
    title = "Mapping the Minds of {LLM}s: A Graph-Based Analysis of Reasoning {LLM}s",
    author = "Xiong, Zhen  and
      Cai, Yujun  and
      Li, Zhecheng  and
      Wang, Yiwei",
    editor = "Christodoulopoulos, Christos  and
      Chakraborty, Tanmoy  and
      Rose, Carolyn  and
      Peng, Violet",
    booktitle = "Proceedings of the 2025 Conference on Empirical Methods in Natural Language Processing",
    month = nov,
    year = "2025",
    address = "Suzhou, China",
    publisher = "Association for Computational Linguistics",
    url = "https://aclanthology.org/2025.emnlp-main.896/",
    doi = "10.18653/v1/2025.emnlp-main.896",
    pages = "17762--17774",
    ISBN = "979-8-89176-332-6"
}

@misc{arnab2021unifiedgraphstructuredmodels,
      title={Unified Graph Structured Models for Video Understanding}, 
      author={Anurag Arnab and Chen Sun and Cordelia Schmid},
      year={2021},
      eprint={2103.15662},
      archivePrefix={arXiv},
      primaryClass={cs.CV},
      url={https://arxiv.org/abs/2103.15662}, 
}

@misc{chu2025understandinglongvideosllmpowered,
      title={Understanding Long Videos via LLM-Powered Entity Relation Graphs}, 
      author={Meng Chu and Yicong Li and Tat-Seng Chua},
      year={2025},
      eprint={2501.15953},
      archivePrefix={arXiv},
      primaryClass={cs.IR},
      url={https://arxiv.org/abs/2501.15953}, 
}

@misc{hsieh2024rulerwhatsrealcontext,
      title={RULER: What's the Real Context Size of Your Long-Context Language Models?}, 
      author={Cheng-Ping Hsieh and Simeng Sun and Samuel Kriman and Shantanu Acharya and Dima Rekesh and Fei Jia and Yang Zhang and Boris Ginsburg},
      year={2024},
      eprint={2404.06654},
      archivePrefix={arXiv},
      primaryClass={cs.CL},
      url={https://arxiv.org/abs/2404.06654}, 
}

@inproceedings{dougrez-lewis-etal-2025-assessing,
    title = "Assessing the Reasoning Capabilities of {LLM}s in the context of Evidence-based Claim Verification",
    author = {Dougrez-Lewis, John  and
      Akhter, Mahmud Elahi  and
      Ruggeri, Federico  and
      L{\"o}bbers, Sebastian  and
      He, Yulan  and
      Liakata, Maria},
    editor = "Che, Wanxiang  and
      Nabende, Joyce  and
      Shutova, Ekaterina  and
      Pilehvar, Mohammad Taher",
    booktitle = "Findings of the Association for Computational Linguistics: ACL 2025",
    month = jul,
    year = "2025",
    address = "Vienna, Austria",
    publisher = "Association for Computational Linguistics",
    url = "https://aclanthology.org/2025.findings-acl.1059/",
    doi = "10.18653/v1/2025.findings-acl.1059",
    pages = "20604--20628",
    ISBN = "979-8-89176-256-5"
}

@article{bapna2022building,
  title={Building machine translation systems for the next thousand languages},
  author={Bapna, Ankur and Caswell, Isaac and Kreutzer, Julia and Firat, Orhan and van Esch, Daan and Siddhant, Aditya and Niu, Mengmeng and Baljekar, Pallavi and Garcia, Xavier and Macherey, Wolfgang and others},
  journal={arXiv preprint arXiv:2205.03983},
  year={2022}
}

@article{qwen3techreport,
  title={{Qwen3 Technical Report}},
  author={An Yang and Anfeng Li and Baosong Yang and Beichen Zhang and Binyuan Hui and Bo Zheng and Bowen Yu and Chang Gao and Chengen Huang and Chenxu Lv and Chujie Zheng and Dayiheng Liu and others},
  journal={arXiv:2505.09388},
  year={2025}
}

@inproceedings{cube,
  title={{Beyond Aesthetics: Cultural Competence in Text-to-Image Models}},
  author={Nithish Kannen and Arif Ahmad and Marco Andreetto and Vinodkumar Prabhakaran and Utsav Prabhu and Adji Bousso Dieng and Pushpak Bhattacharyya and Shachi Dave},
  booktitle=NeurIPSData,
  year={2024},
}

@article{hong2025glm,
  title={Glm-4.5 v and glm-4.1 v-thinking: Towards versatile multimodal reasoning with scalable reinforcement learning},
  author={Hong, Wenyi and Yu, Wenmeng and Gu, Xiaotao and Wang, Guo and Gan, Guobing and Tang, Haomiao and Cheng, Jiale and Qi, Ji and Ji, Junhui and Pan, Lihang and others},
  journal={arXiv preprint arXiv:2507.01006},
  year={2025}
}

@misc{coreteam2025mimovltechnicalreport,
      title={MiMo-VL Technical Report}, 
      author={LLM-Core-Team Xiaomi},
      year={2025},
      eprint={2506.03569},
      archivePrefix={arXiv},
      primaryClass={cs.CL},
      url={https://arxiv.org/abs/2506.03569}, 
}

% APPENDIX
%%%%%%%%%%%%%%%%%%%%%%%%%%%%%%%%%%%%%%%%%%%%%%%%%%%%%%%%%%%%%%%%%%%%%%%%%%%%%%%
%%%%%%%%%%%%%%%%%%%%%%%%%%%%%%%%%%%%%%%%%%%%%%%%%%%%%%%%%%%%%%%%%%%%%%%%%%%%%%%
% \newpage
\appendix
\onecolumn

\vspace{1cm}
\hrule
\par\vspace{0.5cm}
{\Large\bfseries\centering
{Supplementary Material}
\par\vspace{0.5cm}}
\hrule
\vspace{0.5cm}
% \noindent These supplementary materials provide additional details for our paper. The appendix is organized as follows:

\appendix

\noindent The supplementary section is structured as follows: We begin by elaborating on our \emph{Ethical Considerations} (Section~\ref{supp_sec:ethical}) and provide an in-depth description of our \emph{Annotator Guidelines} (Section~\ref{supp_sec:guidelines}), detailing the meticulous protocols for both \curators{} (Section~\ref{curators_guidelines}) in creating culturally-grounded questions and \auditors{} (Section~\ref{auditors_guidelines}) in verifying their objectivity. We then present \emph{Additional Analysis and Experiments} (Section~\ref{supp_sec:analysis}), including studies on the effect of frame sampling (Section~\ref{supp_sec:frame_sampling}), cross-lingual consistency (Section~\ref{fig:cross_lingual_study}), and autorater variance (Section~\ref{supp_sec:autorater_variance}). This is followed by a granular breakdown of our \emph{Error Tagging} methodology (Section~\ref{supp_sec:error_tagging}), which includes the error taxonomy, evidence graph statistics, and details on our multi-iteration analysis. We provide details on \emph{Annotator Recruitment and Compensation} (Section~\ref{supp_sec:annotator_recruitment}) and include the complete \emph{Prompts used} (Section~\ref{supp_sec:prompts}) in our evaluation and diagnostic pipelines. Finally, qualitative examples from the benchmark and their \emph{Error Tagging} outputs are shown in Section~\ref{supp_sec:more_qual_examples}.

\section{Ethical Considerations}
\label{supp_sec:ethical}

The development and deployment of \benchmark{} are guided by a commitment to ethical AI research. Our primary motivation is to counter the Western and English-centric bias prevalent in existing benchmarks. To achieve this, we intentionally curated videos and native-language annotations from 18 diverse global locales through a meticulous, human-in-the-loop process. This approach ensures cultural authenticity and avoids the pitfalls of automated translation. All human annotators were compensated fairly at rates above their local market standards, and our data sourcing included explicit content moderation to exclude violent, explicit, or otherwise harmful material. To further mitigate the risk of perpetuating stereotypes, multiple cultural experts reviewed content for nuanced and respectful representation. The benchmark will be publicly released with clear documentation to guide responsible use, and its fine-grained error analysis is designed to provide a crucial tool for identifying and addressing biases in future multimodal models.

\begin{table}[!ht]
    \centering
    \label{tab:locales_mapping}
    \begin{tabular}{c l l}
        \toprule
        \textbf{Locale Code} & \textbf{Language Name} & \textbf{Region} \\
        \midrule
        \texttt{ar-EG} & Arabic & Egypt \\
        \texttt{de-DE} & German & Germany \\
        \texttt{en-GB} & English & United Kingdom \\
        \texttt{en-IN} & English & India \\
        \texttt{es-MX} & Spanish & Mexico \\
        \texttt{fr-FR} & French & France \\
        \texttt{hi-IN} & Hindi & India \\
        \texttt{id-ID} & Indonesian & Indonesia \\
        \texttt{it-IT} & Italian & Italy \\
        \texttt{ja-JP} & Japanese & Japan \\
        \texttt{ko-KR} & Korean & South Korea \\
        \texttt{mr-IN} & Marathi & India \\
        \texttt{pt-BR} & Portuguese & Brazil \\
        \texttt{ru-RU} & Russian & Russia \\
        \texttt{ta-IN} & Tamil & India \\
        \texttt{te-IN} & Telugu & India \\
        \texttt{th-TH} & Thai & Thailand \\
        \texttt{zh-TW} & Chinese & Taiwan \\
        \bottomrule
    \end{tabular}
    \caption{\textbf{\benchmark{} Locales and their Codes.} This table provides a comprehensive mapping of the 18 distinct cultural and linguistic locales included in our benchmark, showing their standardized \texttt{language-region} codes, corresponding language names (from ISO 639), and region/country names (from ISO 3166-1 alpha-2).}
\end{table}

\section{Annotator Guidelines}
\label{supp_sec:guidelines}
All the textual data in \benchmark{} was completely annotated by human annotators.
All raters are expert native speakers of the language for which they created data.
Here we provide guidelines given to the raters for the dataset creation. 
As described in the Section~\ref{sec:human_curation_pipeline} of the main paper we have two kind of experts for the annotation in our pipeline \curators{} and \auditors{}. We first describe the guidelines for the \curators{} in Section~\ref{curators_guidelines} and then the guidelines for the \auditors{} in Section~\ref{auditors_guidelines}.

\subsection{Guidelines for \curators{}}
\label{curators_guidelines}

\subsubsection{Good Questions}
\begin{compactitem}
    \item The question should \textbf{involve cultural entity/object} of your specific locale that you are working with.
    \item The question should require watching the video to answer the question and 
    \begin{compactitem}
        \item should \textbf{not} be able to solve by \textbf{just listening to the audio}
        \item should \textbf{not} be able to solve by \textbf{just reading some text in the video}
        \item should \textbf{not} be able to solve by \textbf{just using general knowledge}
        \item should \textbf{not} be solvable using \textbf{single frame}
    \end{compactitem}
    \item The question should require multiple reasoning steps to solve.
    \item The question should primarily ask about visual elements in the video (and not just focus on the speech).
    \item The question should have only one right answer and should not be subjective or ambiguous.
    \item The question should be open-ended i.e. it should not be a Multiple Choice Question (MCQ).
    \item The questions, answers and reasoning traces should be strictly in the \textbf{native language} of the video.
    \item If the question involves timestamp in either question/answer/reasoning trace, then use the standard format of \texttt{mm:ss} always. For ex: \texttt{12:56}, \texttt{06:13}
\end{compactitem}

\textbf{Examples of good questions}

\textbf{1. YouTube ID:} \texttt{4EaRHj2Qa3g}  

\textbf{Question:} How many total no. of empty raids happened in the first half from the time the player from Nilgiri Knights gets injured till the time when a team takes the first review?

\textbf{Answer: }Five (5)

\textbf{Skills:} Temporal Event Localization, Counting, Visual Cultural Understanding

\textbf{Why is this a good question?}
This question involves the video from the time the player from Nilgiri Knights gets injured till the time when a team takes the first review in first half (temporal event localization), and counting all the instances of empty raid (counting). The question is about the player in the Kabaddi game (which is a cultural entity). Hence, it satisfies all the criteria of requiring at least three skills (temporal event localization, counting and visual cultural understanding) which makes it a good question.

\textbf{Examples of bad questions}

\textbf{1. Question:} Which event occurs after the old lady serves prasad in this video?

\textbf{Answer:} A red cloth is being placed on young girls head

\textbf{Why is this a bad question?}
This question tests just temporal ordering (which event happens after an event?) and visual cultural understanding (old lady serving prasad). Hence, this just requires 2 skills while we require questions to have a minimum of 3 skills. Hence, this is a bad question.

\textbf{2. Question:} In what order do the following events happen?: People throwing colored powder (gulal) and water on each other, lighting a bonfire (Holika Dahan), sharing sweets \& festive treats with friends and family.

\textbf{Why is this a bad question?}
Because it can be answered with general cultural knowledge of Holi WITHOUT even looking at the video. 

\textbf{3. Question:} How many times does the woman in red saree say the word "namak" in the video?

\textbf{Why is it a bad question?}
Because it is too easy and can be solved by listening to the speech alone. 

\subsubsection{Reasoning steps}
\begin{compactitem}
    \item A reasoning step is an action that you would take to break down the question solving process. You can think of them as the building blocks to the solution.
    \item A good question requires multiple reasoning steps to be performed in sequence to arrive at the answer.
    \item All steps you describe under "Reasoning Steps” must be \textbf{required} to solve the question.
    \item Without one of the steps, a person should not be able to get the answer. 
    \item List all the reasoning steps as a numbered list (one by one)
\end{compactitem}

\paragraph{Good examples of reasoning steps}

\begin{compactitem}
    \item Watch the whole video to understand/obtain a certain piece of information that is important for the answer. 
    Example: I watched the whole video to understand the big plot twist was the nice man was secretly the criminal.
    \item Move to a certain timestamp in a video mm:ss to find an object/person/event. 
    Example: I looked for the red balloon and found it at 12:34
    \item Listen to a word/phrase in the speech. 
    Example: I heard that the man in the black shirt declared "check" at 05:53-05:57
    \item Read a word/phrase on the screen.
    Example: At 04:03 I read that the title of the book is “Amar Chitra Katha”.  
    \item Find all instances of a particular object/person/event. 
    Example: (1) Find all the men in the video wearing a white shirt. (2) Find all the times a goal is scored. 
    \item Counting
    Example: Count how many people wearing white shirts appear in the video.
    \item Temporal Ordering 
    Example: The question mentions a dog, cow and sheep appearing in the video. I looked through the video and noted their time of appearance to bring them in the right order.
    \item Look at something specific in one frame of the video (e.g. you can pause the video and see it) 
    Example:
    \begin{compactitem}
        \item Observe the expression on somebody’s face. 
        \item See the colour of something 
        \item See what the background is 
        \item See what objects are present in the scene 
    \end{compactitem}
    \item Look at an event/action that is happening in a short section of the video (eg. a few seconds)
\end{compactitem}

\textbf{Bad examples of reasoning steps.} Do not add irrelevant information in the steps. 

Example: Question: “How many dogs appear in the video”

Reasoning steps: “1. The video starts by showing the title in white text on a black background. [...]”

\paragraph{What is the difference between a reasoning step and the answer?}
\begin{compactitem}
    \item The final answer can be very short e.g. a single word or a phrase. 
    \item However, the reasoning steps are the entire process to get to the answer.
    \item Hence just because an answer is simple, does not mean multiple steps are not required to get to it.
\end{compactitem}

\subsubsection{Usage of External Tools}
\textbf{Which tools are ok to use and what are not ok to use?}
External tools such as Google Search, reverse Image Search, Wikipedia or any other reliable source are ok to use.

\textbf{Which tools are ok to use and what are not ok to use?}
Use of any \textbf{(Gemini/ChatGPT/Perplexity, Deep Research etc.)} for any kind of task is \textbf{not allowed}. Strictly do not use any kind of LLMs to brainstorm novel questions or to rewrite your questions/answers/reasoning traces. This is a strict requirement. Write the questions/answers/reasoning traces in your own words even if you feel the grammar might not be correct.

\subsection{Guidelines for \auditors{}}
\label{auditors_guidelines}

Here we outline the audit process for the \benchmark{} benchmark. The primary goal of this audit is to ensure the quality, accuracy, objectivity, and consistency of the question-answer (QA) pairs generated by our raters. Your role as \auditors{} is critical in achieving a high-quality benchmark.

\subsubsection{Audit Workflow}
Audit process has two main steps:

\textbf{Step 1: Answer the Question Yourself}

\begin{compactitem}
    \item The \auditors{} will receive a video and a question.
    \item Watch the video.
    \item Read the Question First:
    \begin{compactitem}
        \item Is the question clear and unambiguous? 
        \item If NO (the question is ambiguous, unclear, or could have multiple answers):
        \begin{compactitem}
            \item STOP. Don't try to answer it yet.
            \item Send feedback to the rater explaining why the question is problematic.
            \item Work with the rater to modify the question until it is clear, objective, and aims for a single answer.
            \item Once the question is fixed and clear, proceed to the next point.
        \end{compactitem}
        \item If YES (the question is clear):
        \begin{compactitem}
            \item Now, answer this clear question yourself.
            \item Keep your answer very short (just a few words or phrases).
        \end{compactitem}
    \end{compactitem}
    \item If you find the question ambiguous (ie., not very objective or multiple answers are possible), send this feedback to the rater and after the question is modified then answer the question.
    \item Keep your answer very short (just a few words or phrases) and objective.
    \item You are free to use google search/reverse image search or any other website to answer. 
    \item However, please Do NOT use any kind of LLMs or ChatBots (like ChatGPT, Gemini etc) for anything. 
    \item Very Important: Do NOT look at the rater's answer and reasoning steps (the ``ground truth" answer) yet! 
\end{compactitem}

\textbf{Step 2: Check Your Answer Against the Rater's Answer}
Now, compare the answer you wrote with the ``ground truth" (GT) answer the rater provided. Based on what you find, you'll decide on one of the following scenarios. You can use these scenario codes (like "Scenario A") when you make notes or report issues.

\paragraph{Scenario A: Good Match - QA Approved}
\noindent
    \begin{compactitem}
        \item \textbf{What it means:} Your answer is the same or very similar to the rater's GT answer, and both are correct based on the video.
        \item \textbf{Action:} Mark this QA pair as “Approved”. Move to the next item.
        \begin{compactitem}
            \item \textbf{Example – 1:}
            \begin{compactitem}
                \item Video shows: A video of a woman performing Bharatanatyam.
                \item Question: How many different types of dance moves were performed?
                \item Your Answer: Two or \raisebox{-0.1ex}{\includegraphics{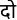}}
                \item Rater's Answer: 2
                \item Decision: This is a good match (Scenario A). Both are correct and essentially the same.
            \end{compactitem}
        \end{compactitem}
    \end{compactitem}

\paragraph{Scenario B: Slight Difference - Needs Minor Fix}
\noindent
    \begin{compactitem}
        \item \textbf{What it means:} Your answer and the rater's GT answer are mostly right and aim for the same thing, but the rater's GT answer could be a bit better (clearer, more exact). The GT isn't wrong, just not perfect. (The question should already be clear from Step 1).
        \item \textbf{Action:}
        \begin{compactitem}
            \item Give feedback to the rater.
            \item Suggest how to improve their answer.
            \item Work with the rater to agree on the best version. The rater will update it.
            \item \textbf{Example:}
            \begin{compactitem}
                \item Video shows: A cat slowly walking across a room.
                \item Question: What is the cat doing?
                \item Your Answer: Walking slowly.
                \item Rater's GT Answer: Moving.
                \item Decision: Scenario B. The rater's GT “Moving” is true, but “Walking slowly” is more precise from the video.
                \item Feedback to Rater: “Could we make the answer more specific, like ‘walking slowly’ or ‘strolling,’ as that's clearly visible in the video?”
            \end{compactitem}
        \end{compactitem}
    \end{compactitem}

    \paragraph{Scenario C: Big Difference - Needs More Review}
    \begin{compactitem}
        \item \textbf{What it means:} Your answer is very different from the rater's GT answer. You need to look closely at why.
        \item \textbf{Action:} Re-watch the video, look at both answers carefully. This will lead to one of three sub-cases:
        \vspace{-3mm}
    \end{compactitem}
        \subparagraph*{Sub-Scenario C1: You're Right, Rater's Wrong.}
        \begin{compactitem}
            \item \textbf{What it means:} Your independent answer is correct, and the rater's GT answer is incorrect based on the video.
            \item \textbf{Action:}
            \begin{compactitem}
                \item Explain to the rater why their GT answer is wrong, using video evidence.
                \item Tell them to update their GT answer to be correct (it might be your answer, or a version you both agree on).
            \end{compactitem}
        \end{compactitem}

        \subparagraph*{Sub-Scenario C2: You're Wrong, Rater's Right.}
        \begin{compactitem}
            \item \textbf{What it means:} After review, you realize your first answer was incorrect, and the rater's GT answer is correct.
            \item \textbf{Action:}
            \begin{compactitem}
                \item No need to tell the rater anything for this specific item if their answer is good. (Good job catching your own mistakes! This helps you too.)
            \end{compactitem}
        \end{compactitem}

        \paragraph{Sub-Scenario C3: Question is Confusing / Multiple Answers Possible.}
        \begin{compactitem}
            \item \textbf{What it means:} The question itself is unclear, ambiguous, or could be interpreted in ways that lead to different “correct” answers. Both your answer and the rater's answer might seem okay (or both wrong) because the question is flawed.
            \item \textbf{Action:} This is a key area for improvement!
            \begin{compactitem}
                \item Talk with the rater about why the question is confusing.
                \item Work together to rewrite the question to be very clear, specific, and have only ONE obvious correct answer from the video.
                \item Then, agree on the new, single correct GT answer for the improved question.
                \item The rater will update both the question and the GT answer.
                \item Our Goal: Fix these so they become clear like Scenario A.
                \item Work with the rater to make the question specific.
            \end{compactitem}
        \end{compactitem}

\section{Additional Analysis, Experiments and Results}
\label{supp_sec:analysis}

\noindent In this section we explore the impact of temporal sampling density on model performance (Section~\ref{supp_sec:frame_sampling}), assess cross-lingual reasoning consistency (Section~\ref{supp_sec:cross_lingual_exps}), and confirm the robustness of our evaluation pipeline through an autorater reliability analysis Section~\ref{supp_sec:autorater_variance}.

\subsection{Effect of Number of Frames}
\begin{figure}[ht]
    \centering
    \begin{minipage}[t]{0.48\textwidth}
        \centering
        \includegraphics[width=\linewidth]{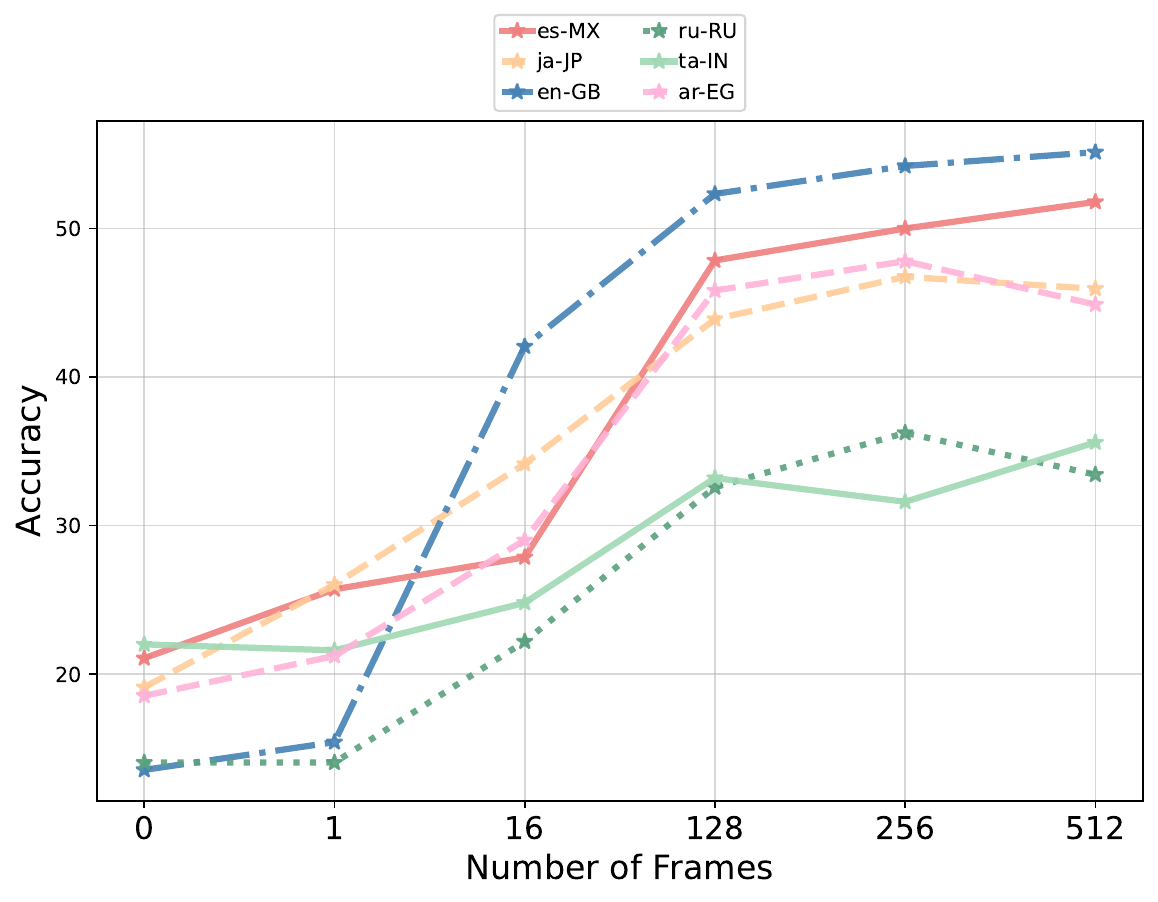}
        \caption{\textbf{Impact of frame sampling on model accuracy.} Gemini-2.5-Pro's performance improves with more frames, validating the temporal nature of our benchmark.}
        \label{fig:frame_ablation_gemini}
    \end{minipage}
    \hfill
    \begin{minipage}[t]{0.48\textwidth}
        \centering
        \includegraphics[width=\linewidth]{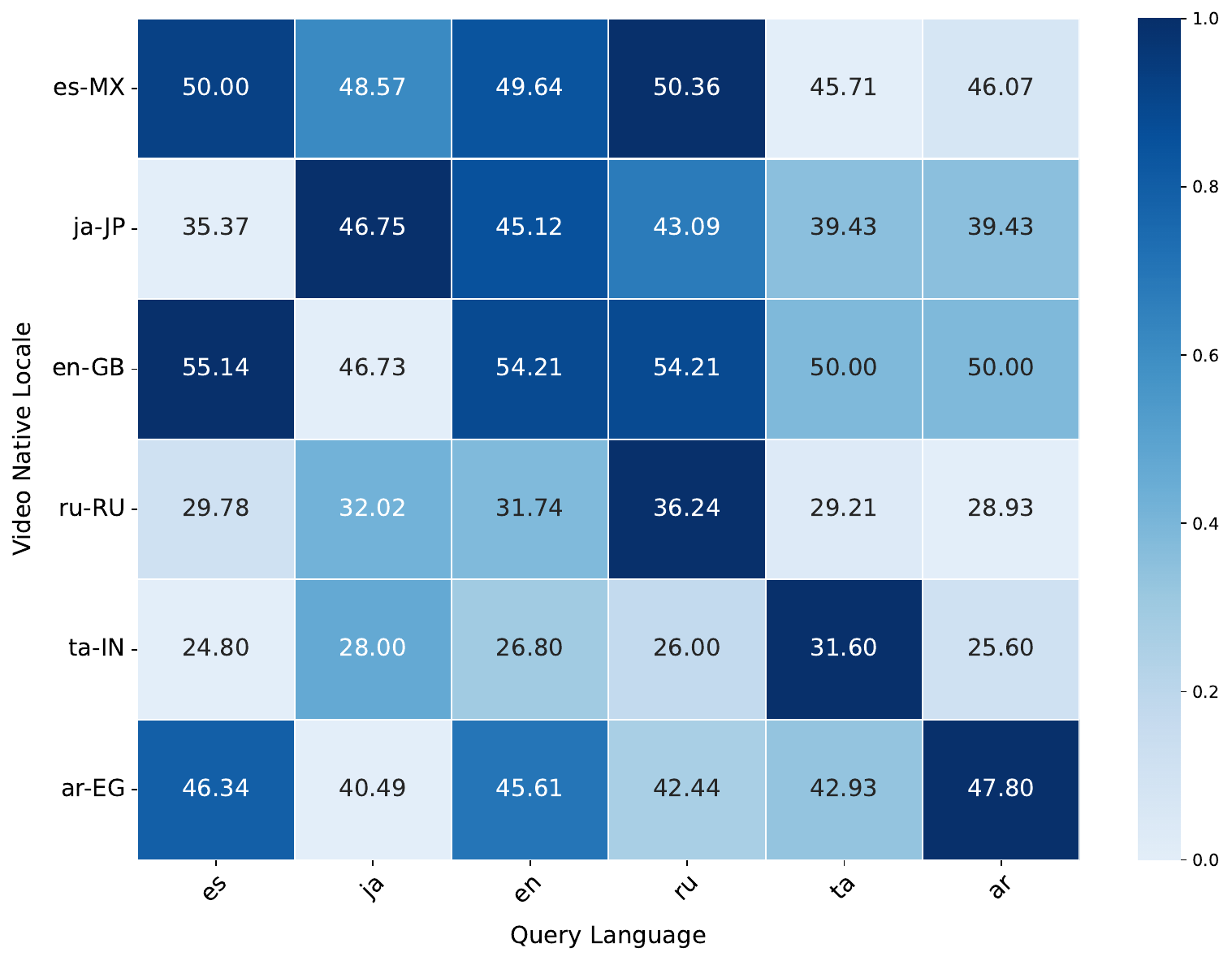}
        \caption{\textbf{Cross-Lingual Consistency Analysis.} Colors are normalized row-wise. The consistently high scores along the diagonal demonstrate that the model performs best when the query is in the video's native language, while off-diagonal variance highlights significant cross-lingual performance gaps.}
        \label{fig:cross_lingual_study}
    \end{minipage}
\end{figure}

As discussed in Section~\ref{sec:model_performance} of the main paper, we evaluated the temporal complexity of \benchmark{} by varying the number of sampled input frames from 1 to 512. The results of this analysis on \geminipro{}, visualized in Figure~\ref{fig:frame_ablation_gemini}, confirm our primary findings. The plot demonstrates a monotonic increase in accuracy with more frames across a diverse subset of locales, validating that tasks in \benchmark{} require temporal reasoning and cannot be resolved from static images. Concurrently, the diminishing performance gains at higher frame counts reinforce our conclusion that the primary performance bottleneck is the higher-level, culturally-contextualized reasoning demanded by the benchmark, rather than a mere lack of visual information.

\label{supp_sec:frame_sampling}

% Model Hyperparams
\begin{table*}[ht]
\centering
\caption{Hyperparameters for all model baselines}
\label{tab:hyperparameters}
\begin{tabular}{llc}
\toprule
\textbf{Method} & \textbf{\# of Frames} & \textbf{Hyperparameters} \\
\midrule
Qwen-2.5-VL & default & togetherAI API \footnotemark[1] default inference\\
\midrule
Qwen-3-VL & default & vLLM inference \footnotemark[2], max-seq-length=128k \\
\midrule
Claude-Sonnet-4 & 100 & thinking\_budget\_tokens=10000 \\
\midrule
GPT-5-mini & 256 & reasoning\_effort=high, verbosity=default, max\_output\_tokens=default \\
\midrule
GPT-5 & 256 & reasoning\_effort=high, verbosity=default, max\_output\_tokens=default \\
\midrule
Gemini-2.5-flash & 256 & temperature=0, thinking=dynamic, seed=default, sampling=default \\
\midrule
Gemini-2.5-pro & 256 & temperature=0, thinking=dynamic, seed=default, sampling=default \\
\bottomrule
\end{tabular}
\end{table*}

\subsection{Cross-Lingual Consistency Experiments}
\label{supp_sec:cross_lingual_exps}
To assess how query language affects reasoning, we performed a cross-lingual evaluation where we translated questions into the five other languages while keeping the visual input fixed. As shown in Figure~\ref{fig:cross_lingual_study}, the results confirm that model performance is not language-agnostic. Accuracy is usually highest when the query is in the video's native language (the diagonal). While English and Russian serve as more robust translation targets, performance varies significantly based on both the source and target languages. This demonstrates that simply translating queries is an insufficient strategy to overcome cultural and linguistic dependencies.

\subsection{Autorater Variance and Judge Reliability}
\label{supp_sec:autorater_variance}
\begin{table}[!ht]
\centering
\begin{tabular}{l S[table-format=2.2] S[table-format=2.2]}
\toprule
& \multicolumn{2}{c}{\textbf{Autorater}} \\
\cmidrule(lr){2-3}
\textbf{Locale} & {\textbf{Gemini-2.5-Flash}} & {\textbf{GPT-5-mini}} \\
\midrule
\texttt{ar-EG} & 47.80 & 48.29 \\
\texttt{de-DE} & 48.73 & 48.73 \\
\texttt{en-GB} & 54.21 & 53.74 \\
\texttt{en-IN} & 47.71 & 47.79 \\
\texttt{es-MX} & 50.00 & 49.64 \\
\texttt{fr-FR} & 52.56 & 53.84 \\
\texttt{hi-IN} & 41.89 & 43.24 \\
\texttt{id-ID} & 55.97 & 54.85 \\
\texttt{it-IT} & 51.47 & 51.47 \\
\texttt{ja-JP} & 46.75 & 46.75 \\
\texttt{ko-KO} & 64.29 & 64.29 \\
\texttt{mr-IN} & 38.72 & 38.35 \\
\texttt{pt-BR} & 43.75 & 44.27 \\
\texttt{ru-RU} & 36.24 & 36.24 \\
\texttt{ta-IN} & 31.60 & 30.80 \\
\texttt{te-IN} & 28.00 & 25.60 \\
\texttt{th-TH} & 39.03 & 38.71 \\
\texttt{zh-TW} & 40.00 & 40.00 \\
\midrule
Macro Avg & 45.48 & 45.37 \\
Weighted Avg & 45.07 & 44.97 \\
\bottomrule
\end{tabular}
\caption{\textbf{Autorater Reliability Analysis}. Comparison of scores for \geminipro{} on the \benchmark{} benchmark as evaluated by two distinct LLM judges. The minimal variance in both per-locale and aggregate scores demonstrates the robustness of our evaluation methodology.}
\label{tab:autorater}
\end{table}

% The primary evaluation pipeline for CURVE relies on \geminiflash{} as an LLM Judge to score model responses, as detailed in Section~\ref{sec:expts} of the main paper. To ensure the robustness and reproducibility of our findings, we conducted a validation experiment to assess the variability of this autorater. The central concern was to verify that our reported results are not biased by the specific LLM Judge chosen, but rather a true reflection of the models' capabilities on our benchmark.
% For this analysis, we re-evaluated the complete set of responses from our top-performing model, \geminipro{} (Table~\ref{tab:model-performance} of the main paper), using an alternative powerful model, \gptmini{}, as a secondary autorater. We used the same scoring prompt for both judges. Table~\ref{tab:autorater} presents a comparison of the scores assigned by \geminiflash{} and \gptmini{} across all 18 locales. The results demonstrate a minimal variance between the autoraters. The aggregate scores are remarkably close. The weighted average from \geminiflash{} is 45.07, while \gptmini{} yields 44.97, a negligible difference of just 0.1\%. The macro averages are similarly aligned at 45.48 and 45.37.

The primary evaluation pipeline for \benchmark{} relies on \geminiflash{} as an LLM Judge to score model responses, as detailed in Section~\ref{sec:expts} of the main paper. To ensure the robustness and reproducibility of our findings, we conducted a validation experiment to assess the variability of this autorater. The central concern was to verify that our reported results are not biased by the specific LLM Judge chosen, but rather a true reflection of the models' capabilities on our benchmark.
For this analysis, we re-evaluated the complete set of responses from our top-performing model, \geminipro{} (Table~\ref{tab:model-performance} of the main paper), using an alternative powerful model, \gptmini{}, as a secondary autorater. We used the same scoring prompt for both judges.
Table~\ref{tab:autorater} presents a comparison of the scores assigned by \geminiflash{} and \gptmini{} across all 18 locales. The results demonstrate a minimal variance between the autoraters. The aggregate scores are remarkably close. The weighted average from \geminiflash{} is 45.07, while \gptmini{} yields 44.97, a negligible difference of just 0.1\%. The macro averages are similarly aligned at 45.48 and 45.37.

\footnotetext[1]{https://www.together.ai/qwen}
\footnotetext[2]{https://docs.vllm.ai/projects/recipes/en/latest/Qwen/Qwen3-VL.html}

\paragraph{Open-Source Judge Validation.} To further address potential concerns about relying solely on closed-source LLM judges, we also evaluated using QWEN-3, an open-source model. As shown in Table~\ref{tab:judge_reliability_supp} (Right), the QWEN-3 judge produces a weighted average of 44.73, with an agreement of 93.4\% with \geminiflash{}. The high consistency across all three judges (\geminiflash{}, \gptmini{}, and QWEN-3) is attributable to the fact that questions in \benchmark{} are designed to be objective (\eg{} counting) and unambiguous, making the judging task a straightforward semantic match.

\paragraph{Human-Judge Alignment.} We also conducted a human-judge agreement study to validate the LLM-based evaluation against human grading. Human evaluators independently scored model responses for two locales: \texttt{en-GB} (a high-resource language) and \texttt{ta-IN} (a low-resource language). As shown in Table~\ref{tab:judge_reliability_supp} (Left), the \geminiflash{} judge achieves an agreement of 96.6\% with human evaluators, confirming the efficacy of the autorater across both high- and low-resource settings.

\begin{table}[t]
\centering
\small
\tabcolsep=3.0mm
\begin{tabular}[t]{@{}l c@{}}
\toprule
Judge & Agreement \\
\midrule
GPT-5-Mini & 95.3 \\
QWEN-3 & 93.4 \\
Human* & 96.6 \\
\bottomrule
\end{tabular}
\qquad \qquad
\begin{tabular}[t]{@{}l c@{}}
\toprule
Judge & Weighted Avg. \\
\midrule
Gemini-2.5-Flash & 45.07 \\
GPT-5-Mini & 44.97 \\
QWEN-3 & 44.73 \\
\bottomrule
\end{tabular}
\vspace{-2mm}
\caption{(Left) Judge agreement with \geminiflash{}. (Right) Weighted average score of \geminipro{} on \benchmark{} when graded by different judges. \footnotesize * Human agreement calculated across \texttt{en-GB} and \texttt{ta-IN} locales.}
\label{tab:judge_reliability_supp}
\end{table}

\subsection{Web Search Augmented Evaluation}
\label{supp_sec:search_eval}
To assess whether access to external knowledge can help bridge the performance gap on \benchmark{}, we evaluated GPT-5 with web search enabled. As shown in Table~\ref{tab:search_supp}, enabling search yields a modest improvement of 2.6\% (from 42.2 to 44.8 weighted average). While search helps, a massive performance gap to the human baseline remains, confirming that the primary bottleneck in \benchmark{} is \emph{visual grounding} of cultural concepts, which text-based retrieval cannot fully address.

\begin{table}[t]
\centering
\small
\begin{tabular}{@{}l c c@{}}
\toprule
Model & GPT-5 & GPT-5 + Search \\
\midrule
Weighted Avg. & 42.2 & 44.8 \\
\bottomrule
\end{tabular}
\vspace{-2mm}
\caption{Effect of enabling web search on GPT-5's performance on \benchmark{}.}
\label{tab:search_supp}
\end{table}

\subsection{Additional Open-Source Models}
\label{supp_sec:opensource_models}
In addition to the QWEN family of models evaluated in the main paper, we further benchmark two recent open-source Video-LLMs on \benchmark{}: GLM-V4.1-9B-Thinking~\citep{hong2025glm} and MiMo-VL-7B-RL~\citep{coreteam2025mimovltechnicalreport}. As shown in Table~\ref{tab:opensource_supp}, their weighted average scores are comparable to Qwen-2.5-VL, and the overall gap between these models and human performance remains massive ($\sim$80\%), underscoring the difficulty of \benchmark{} for current open-source models.

\begin{table}[t]
\centering
\small
\begin{tabular}{@{}l c c c c@{}}
\toprule
Model & Qwen-3 & GLM-V4.1 & Qwen-2.5 & MiMo \\
\midrule
Weighted Avg. & 21.5 & 15.2 & 12.8 & 11.6 \\
\bottomrule
\end{tabular}
\vspace{-2mm}
\caption{Weighted average scores of additional open-source models on \benchmark{}.}
\label{tab:opensource_supp}
\end{table}

\subsection{Text-Only Baseline}
\label{supp_sec:text_only}
To verify that \benchmark{} questions cannot be solved through textual shortcuts alone, we evaluate \geminipro{} without any video context (i.e., 0 frames). Across the six evaluation locales, the average accuracy is just 18\% (Figure~\ref{fig:frame_ablation_gemini}), indicating that the questions require genuine video understanding and cannot be answered through world knowledge or linguistic cues alone. This confirms the multimodal nature of the reasoning required by \benchmark{}.

\section{Error Tagging}
\label{supp_sec:error_tagging}
\subsection{Graph Statistics}
\begin{table}[ht]
\centering
\begin{tabular}{lrrr}
\toprule
\textbf{Property} & \textbf{Mean} & \textbf{$\sigma$} & \textbf{Total} \\
\midrule
Nodes & 5.0 & 2.5 & 4351 \\
Nodes w/ Timestamps & 3.1 & 2.5 & 2743 \\
Edges & 4.4 & 3.1 & 3865 \\
Depths & 2.5 & 1.3 & N/A \\
\bottomrule
\end{tabular}
\caption{Statistical overview of graphs derived from human reasoning in selected locales (N=878 questions).}
\label{tab:full_stats}
\end{table}
The first step of the Reasoning Trace based analysis is to develop the \emph{Evidence Graph}.
Once the \emph{Evidence Graph} is developed, we attempt to understand the \benchmark{} with the help of the graph's structural properties.
The key properties are discussed in Section~\ref{error_tagging_analysis} under the \textcolor{red}{Structural Analysis} heading.
The statistics are described in Table~\ref{tab:full_stats}.

\subsection{Taxonomy}
\begin{table*}[h!]
    \centering
    \begin{tabular}{l p{0.35\textwidth} p{0.35\textwidth}}
        \toprule
        \textbf{Error Type} & \textbf{Definition} & \textbf{Illustrative Example} \\
        \midrule
        
        \textbf{Temporal Localization} & A failure to search the correct time segment in the video. & A required clue is at 01:15, but the model only searches before 00:30. \\
        \addlinespace

        \textbf{Spatial Grounding} & A failure to "see" an object or event present in the correct spatiotemporal location. & Looking at a parade float but failing to identify the banner on its side. \\
        \addlinespace

        \textbf{Spurious Object/Event} & An invention error where the model over-counts or fabricates an object or event that is not present. & Claiming three dancers are wearing hats when only two do. \\
        \addlinespace

        \textbf{Attribute Misidentification} & A failure to correctly identify the properties of a detected object. & Correctly identifying a car but stating it is blue when it is green. \\
        \addlinespace

        \textbf{Knowledge-Dependent Issue} & A failure to recall or apply correct external factual knowledge. & Identifying Japan's flag but failing to recall that Tokyo is its capital. \\
        \addlinespace

        \textbf{Reasoning} & A logical failure in connecting evidence when all prerequisite information is correct. & Seeing Team A celebrating and Team B sad, but concluding that Team B won. \\

        \bottomrule
    \end{tabular}
    \caption{\textbf{Error Taxonomy with Definitions and Examples.} This table details the classification system used for our fine-grained error analysis. Each category represents a distinct failure mode, allowing for a precise diagnosis of model weaknesses in multicultural video reasoning.}
    \label{tab:error_taxonomy}
\end{table*}

We tag each failure to map the evidence to the reasoning trace in the node.
The tagging is supposed to understand the core reason why the evidence was not retrieved.
MINERVA uses an LLM to score for Perceptual Correctness, Temporal Localization, and Logical Reasoning.
We expand across two dimensions: first, we expand the taxonomy by splitting the broad class of Perceptual Correctness into three new categories of Spatial Grounding, Attribute Misidentification, and Spurious Objects/Events.
The exact definition and a brief example is given in Table~\ref{tab:error_taxonomy}.
Second, instead of scoring the entire solution, we divide the solution into different evidences(nodes) and then classify the type of error.
This helps us better identify \textit{where and when} the model goes wrong.

\subsection{Majority Voting and Human Assessment}
To improve the robustness of the prompt-based analysis on the task of error tagging, we employ majority voting.
For each question, we query \geminipro{} three times and take the majority consensus.
We present the consensus between 3 LLMs in the error tagging of Iteration 1 in Table~\ref{tab:majority_consensus}.
It highlights the high consensus between the LLMs with over 97.7\% of examples achieving majority consensus. 
The locale-wise consensus status is described in Table~\ref{tab:majority_consensus}.

\begin{table}[H]
\centering
\begin{tabular}{lrr}
\toprule
\textbf{Locale} & \textbf{Majority Consensus} & \textbf{No Consensus} \\
 & ($\ge$ 2/3 Agree) & (All Differ) \\
\midrule
\texttt{en-GB} & 106 & 11 \\
\texttt{es-MX} & 109 & 0 \\
\texttt{ja-JP} & 95 & 1 \\
\texttt{ar-EG} & 247 & 4 \\
\texttt{ru-RU} & 140 & 3 \\
\texttt{ta-IN} & 157 & 1 \\
\bottomrule
\end{tabular}
\caption{LLM Majority Consensus by Locale}
\label{tab:majority_consensus}
\end{table}

The verification of tagging requires significant human effort, taking into consideration the human reasoning trace, the evidence graph and the model thoughts.
We manually verify the error tagging for a small randomly selected subset of 60 questions distributed amongst the six selected locales.
We find that the model-human agreement is at an high $88.5\%$, verifying the validity of the error tagging.
However, it also shows the scope of improvement that is possible in the LLM-based evaluation.

\subsection{Multi-Iteration Error Isolation}
\begin{figure}[t]
\centering
\includegraphics[width=0.7\textwidth]{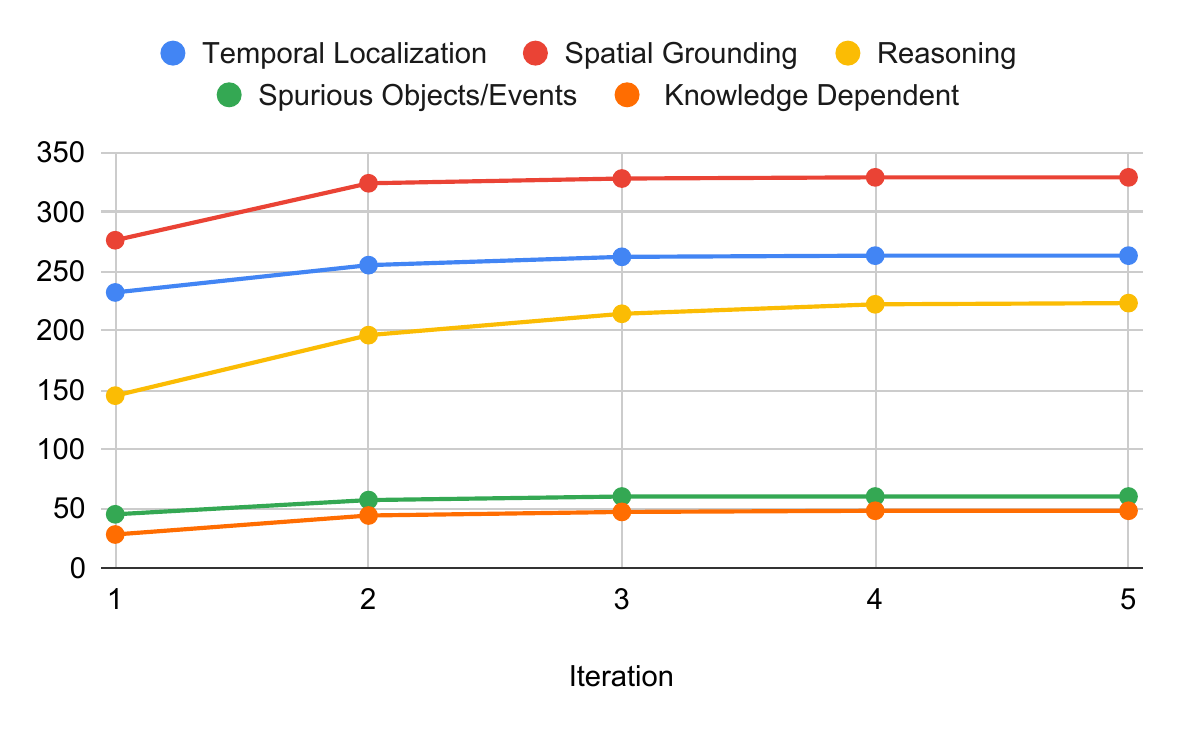}
\caption{Iteration-wise Error Distribution for Gemini 2.5 Pro.}
\label{fig:iteration_chart}
\end{figure}
We find that the Multi-Iteration Error Isolation helps us gather a complete understanding through the counterfactual hint generation as described in the Section~\ref{iterative_error_isolation} of the main paper.
We see a steady increase in error types like Temporal Localization, Spatial Grounding and Reasoning.
Reasoning sees the maximum gain indicating that the later steps of the question hidden by the errors in the initial errors have a higher proportion of reasoning and evidence aggregation tasks.
The complete graph can be studied at Figure~\ref{fig:iteration_chart}.

\section{Annotator Recruitment and Compensation}
\label{supp_sec:annotator_recruitment}
Our data was annotated by professional data labelers. We recruited an average of six \curators{} and five \auditors{} from each of the 18 countries. The selection process for \curators{} and \auditors{} was based on content writing expertise and native language proficiency with deep, situated cultural knowledge, which was assessed through both verbal and written language assessments. In addition, the \auditors{} underwent a vetting process to assess their analytical and observational skills and a proven ability to maintain consistency and accuracy across a large volume of annotations. All annotators were compensated at rates above the prevalent market rates in their respective countries, in full compliance with local minimum wage regulations. 

% \vspace{-300mm}
\section{Prompts used}
\label{supp_sec:prompts}
\begin{figure*}[t]
\centering
\noindent\begin{minipage}{\textwidth}
\mdfsetup{%
middlelinewidth=1pt,
backgroundcolor=cyan!10,
innerleftmargin=0.5cm,
innerrightmargin=0.5cm,
font=\small,
roundcorner=15pt}
\begin{mdframed}
\vspace{0.2em}

\textbf{CULTURAL AUTORATER PROMPT\\}

You are an expert at global culture and you are grading an exam where the answer could be given in any language.

You will be provided with a question, the reference answer and the model's answer. Your job is to judge if the model's answer was equivalent to the reference answer and provide a single integer score between \texttt{0} and \texttt{2} with the following criteria:

\texttt{0}: The model's answer is completely wrong and is not related to the reference.

\texttt{1}: The model's answer is partially correct and misses some details from the reference answer.

\texttt{2}: The model's answer is completely correct.

\vspace{1mm}
\hrule
\vspace{1mm}

\textbf{INSTRUCTION:}

\begin{compactenum}

    \item You should accept answers that are accurate translations or transliterations of the reference answer. The language or script of the answer should not be taken into account. For example:
    \begin{compactenum}[a)]
        \item reference answer = \raisebox{-0.8ex}{\includegraphics[width=0.05\textwidth]{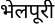}}
    
        \texttt{model's answer} =  \texttt{Bhel Puri}
    
    \texttt{score} = \texttt{2}
    \end{compactenum}

    \item Allow for alternate names of the same cultural concept. Don't penalize for spelling errors or minor variations. Focus solely on the cultural concept. For example:
    \begin{compactenum}[a)]
    
    \item \texttt{reference answer} = \texttt{The London Eye};
    
        \texttt{model's answer} =  \texttt{Millennium Wheel}
    
        \texttt{score}= \texttt{2}

    % \item \texttt{reference answer} = \texttt{Mumbai}
    
    %     \texttt{model's answer} = \texttt{Bombay}
    
    %     \texttt{score} = \texttt{2}

    % \item \texttt{reference answer} = \texttt{Angkor Wat}
    
    %     \texttt{model's answer} = \texttt{Angkor Vat}
    
    %     \texttt{score} = \texttt{2}

    \end{compactenum}

    \item Partial scoring of \texttt{1} can be given in cases where model misses details or the answer is not complete. For example:

    \begin{compactenum}[a)]
    
    % \item \texttt{reference answer} = \texttt{Yangzhou Fried Rice}
    
    %     \texttt{model's answer} = \texttt{Fried Rice}
    
    %     \texttt{score} = \texttt{1}
    
    % \item \texttt{reference answer} = \texttt{Thallesary Biriyani}
    
    %     \texttt{model's answer} = \texttt{Biryiani}
    
    %     \texttt{score} = \texttt{1}
    
    \item \texttt{reference answer} = \texttt{Sun Temple}
    
        \texttt{model's answer} = \texttt{Temple}
    
        \texttt{score} = \texttt{1}

    \end{compactenum}

    \item For questions with a numerical answer, the score is determined by an exact value match. Award a full score of \texttt{2} if the model's answer represents the same numerical value as the reference answer, regardless of whether it is written in digits (e.g., \texttt{10}) or words (e.g., \texttt{ten}). If the model's answer represents any other number and there is no exact value match, it must receive a score of \texttt{0}. No partial credit is given. For example,

    \begin{compactenum}[a)]
        \item \texttt{reference answer} = \texttt{10}
    
        \texttt{model's answer} = \texttt{ten}
    
        \texttt{score} = \texttt{2}
    
        \item \texttt{reference answer} = \texttt{10}
    
        \texttt{model's answer} = \texttt{11}
    
        \texttt{score} = \texttt{0}

    \end{compactenum}

    \item Your answer should only contain a single integer value in \texttt{[0, 1, 2]} and nothing else.

\end{compactenum}

\vspace{1mm}
\hrule
\vspace{1mm}

\texttt{Question:} \texttt{question}

\texttt{Reference answer:} \texttt{gt}

\texttt{Model's answer:} \texttt{pred}

\texttt{Your response:}

\end{mdframed}
\end{minipage}
\caption{\textbf{The prompt used for the LLM Judge (\geminiflash)} in our main evaluation pipeline (Tab. 2 of the main paper). It details the three-point (0-2) scoring criteria for assessing the semantic equivalence between a model's response and the ground-truth answer.}
\label{fig:main_autorater_prompt}
\vspace{-1em}
\end{figure*}

For completeness and to ensure full reproducibility of our methodology, this section provides the exact prompts used for our automated evaluation and diagnostic analysis. We show the prompt of our LLM Judge used for the main model evaluation (Table~\ref{tab:model-performance} of the main paper) in Figure~\ref{fig:main_autorater_prompt}. Following this, we provide the prompt (in Figure~\ref{fig:prompt_graph}) used to formalize the unstructured human reasoning traces into a structured Directed Acyclic Graph (DAG).
This serves as the ground-truth \emph{Evidence Graph} for our error analysis (Figure~\ref{fig:error_tagging} of main paper).
Finally, we detail the comprehensive prompt (in Figures~\ref{fig:tag_graph_1},~\ref{fig:tag_graph_2}) that governs our Iterative Error Isolation pipeline (Figure~\ref{fig:error_isolation} of main).
This prompt instructs the LLM to traverse the Evidence Graph, tag specific failure modes according to our taxonomy, and generate corrective hints for subsequent iterations.

\clearpage

\begin{figure*}[t]
\centering
\noindent\begin{minipage}{\textwidth}
\mdfsetup{%
middlelinewidth=1pt,
backgroundcolor=cyan!10,
innerleftmargin=0.5cm,
innerrightmargin=0.5cm,
font=\small,
roundcorner=15pt}
\begin{mdframed}
\vspace{0.2em}

\textbf{REASONING GRAPH PROMPT\\}

\noindent You are a Reasoning Analyst. Your task is to convert a Video QA reasoning process into a directed graph representing the direct solution path.

\vspace{2mm}
\hrule
\vspace{2mm}

\noindent \textbf{Step-by-Step Instructions}

\vspace{2mm}
\noindent \textbf{1. Identify Atomic Nodes:} Extract the causal chain of evidence, \textbf{excluding} all procedural text, negative findings, and dead ends. Create an ``atomic evidence node'' (Node1, Node2...) for each distinct step. A node must be: \\
\indent $\bullet$ Derived from a single video timestamp, OR \\
\indent $\bullet$ A single piece of external information, OR \\
\indent $\bullet$ A single inference over existing evidence. \\
\indent $\bullet$ \textit{Action:} Strip conversational preambles from \texttt{<Evidence Text>} and record the \texttt{<Timestamp>} (or ``N/A'').

\vspace{2mm}
\noindent \textbf{2. Construct Graph:} Define directed edges. Add a node ID to \texttt{parent\_nodes} \textbf{ONLY} if the current node is strictly dependent on that parent node's information.

\vspace{2mm}
\noindent \textbf{3. Validate Termination:} Ensure the graph ends with the final \texttt{Answer}. If no existing node contains the answer, create a final node containing the answer text and link it to its immediate evidence.

\vspace{2mm}
\noindent \textbf{Output Format:} \\
Return \textbf{ONLY} a single valid JSON object.

\vspace{2mm}
\noindent \texttt{\{ \\
\hspace*{1em} "solution\_graph": \{ \\
\hspace*{2em} "Node1": \{ \\
\hspace*{3em} "evidence": "<Evidence Text>", \\
\hspace*{3em} "timestamp": "<Timestamp or N/A>", \\
\hspace*{3em} "parent\_nodes": ["<List of parent IDs>"] \\
\hspace*{2em} \}, \\
\hspace*{2em} ... \\
\hspace*{1em} \} \\
\}}

\vspace{2mm}
\hrule
\vspace{2mm}

\noindent \textbf{Input} \\
\texttt{[BEGIN INPUT]} \\
\texttt{Question: \{question\}} \\
\texttt{Answer: \{answer\}} \\
\texttt{Human Response: \{human\_response\}}

\end{mdframed}
\end{minipage}
\caption{\textbf{The prompt used for the Reasoning Analyst LLM} to derive the Directed Acyclic Graph (DAG) from human reasoning steps. It instructs the model to strip procedural noise and isolate the atomic causal chain leading to the final answer.}
\label{fig:prompt_graph}
\vspace{-1em}
\end{figure*}
\begin{figure*}[t]
\centering
\noindent\begin{minipage}{\textwidth}
\mdfsetup{%
middlelinewidth=1pt,
backgroundcolor=cyan!10,
innerleftmargin=0.5cm,
innerrightmargin=0.5cm,
font=\small,
roundcorner=15pt}
\begin{mdframed}
\vspace{0.2em}

\textbf{VIDEO QA EVALUATOR PROMPT (PART 1)\\}

You are a rigorous AI evaluator for Video QA. Your goal is to analyze a language model's reasoning against a ground-truth \texttt{Evidence Graph} to generate a cue for its next attempt.

\textbf{Golden Rule:} The \texttt{Evidence Graph} is the absolute source of truth. Flag only \textbf{causally relevant} errors that damage the answering process.

\vspace{1mm}
\hrule
\vspace{1mm}

\textbf{Task 1: Evaluate the Solution Graph}

Evaluate each node in the \texttt{Evidence Graph} sequentially. Stop at the first rule that applies.

\begin{compactenum}
    \item \textbf{Check for Inherited Errors (\texttt{status}: undetermined)} \\
    If any parent node is \texttt{wrong}, \texttt{undetermined}, or has \texttt{divergence: true}, the current node inherits the failure.
    \begin{compactenum}
        \item \textbf{Action:} Set \texttt{status} to \texttt{undetermined}. Provide \texttt{justification}. Stop.
    \end{compactenum}

    \item \textbf{Check for Valid Divergence (\texttt{divergence}: true)} \\
    If all parents are \texttt{right}, check if the model pursues a valid alternative path. Set \texttt{divergence} to \texttt{true} (and stop) \textbf{only if} all conditions are met:
    \begin{compactenum}[1.]
        \item \textbf{Alternative Evidence:} The model uses new evidence distinct from the human reasoning.
        \item \textbf{Productive:} The path is logical and leads toward the answer (not a dead end).
        \item \textbf{Non-Contradictory:} The path does not contradict or compete against the human reasoning, question, or answer.
    \end{compactenum}

    \item \textbf{Evaluation} \\
    Determine \texttt{status} based on answer-affecting information only (ignoring methodology).
    \begin{compactenum}[1.]
        \item \textbf{Content Check:} Did the model find the correct evidence content?
        \begin{compactenum}
            \item \textbf{If No:} Set \texttt{status} to \texttt{wrong}. Proceed to \textbf{Causal Analysis}.
            \item \textbf{If Yes:} Proceed to Timestamp Check.
        \end{compactenum}
        \item \textbf{Timestamp Check:} Is the timestamp critical \textbf{AND} outside the +/- 5s tolerance?
        \begin{compactenum}
            \item \textbf{If Yes:} Set \texttt{status} to \texttt{wrong}. Proceed to \textbf{Causal Analysis}.
            \item \textbf{If No:} (Timestamp not critical OR within tolerance). Set \texttt{status} to \texttt{right}. Evaluation complete.
        \end{compactenum}
    \end{compactenum}
\end{compactenum}

\vspace{1mm}
\hrule
\vspace{1mm}

\textbf{Part B: Causal Analysis for Failures} \\
If \texttt{status} is \texttt{wrong}, identify the primary cause of failure by finding the \textbf{first critical mistake} in the \texttt{model\_output\_thoughts}. Go through the causal categories in order and stop at the first match.

\textbf{Causal Analysis: A Step-by-Step Diagnosis}
\begin{compactenum}
    \item \textbf{Intent/Planning Failure} (\texttt{Reasoning}): Did the model misinterpret the user's goal or fail to form a correct search plan?
    \item \textbf{Knowledge Failure} (\texttt{Knowledge-Dependent Issue}): Was the plan correct, but the model relied on an incorrect internal fact or tool result?
    \item \textbf{Temporal Failure} (\texttt{Temporal Localization}): Did the model fail to search within the correct time segments or miss a critical timestamp?
    \item \textbf{Spatial / Detection Failure} (\texttt{Spatial Grounding}): *(Assumption: Time is correct)*. Did the model miss a visual/audio object, under-count objects, or fail to find an object matching the attribute specifications?
    \item \textbf{Attribute Failure} (\texttt{Attribute Misidentification}): *(Assumption: Object detection is correct)*. Did the model incorrectly identify properties (color, text, type) of the found object?
    \item \textbf{Hallucination} (\texttt{Spurious Object/Event}): Did the model \textbf{over-count} objects or invent events?
    \item \textbf{Logical Failure} (\texttt{Reasoning}): Did the model find the correct evidence but fail to connect it to the right conclusion, or ignore provided info?
\end{compactenum}

\end{mdframed}
\end{minipage}
\caption{\textbf{Prompt to traverse through the graph and tag errors. (Part 1)}}
\label{fig:tag_graph_1}
\end{figure*}

\pagebreak

\begin{figure*}[t]
\centering
\noindent\begin{minipage}{\textwidth}
\mdfsetup{%
middlelinewidth=1pt,
backgroundcolor=cyan!10,
innerleftmargin=0.5cm,
innerrightmargin=0.5cm,
font=\small,
roundcorner=15pt}
\begin{mdframed}
\vspace{0.2em}

\textbf{VIDEO QA EVALUATOR PROMPT (PART 2)\\}

\textbf{4. Justify}
\begin{compactenum}
    \item \textbf{If \texttt{divergence} is \texttt{true} AND \texttt{evidence\_retrieved} is \texttt{no}}: Explain \textit{why} it failed using the findings from Part A and Part B.
    \item \textbf{Otherwise}: Set to \texttt{N/A}.
\end{compactenum}

\vspace{1mm}
\hrule
\vspace{1mm}

\textbf{Task 2: Generate the \texttt{next\_attempt\_cue}} \\
Create a single, cumulative block of text to guide the model's next attempt.
\begin{compactenum}
    \item \textbf{Start:} Include all text from \texttt{Additional Cues}.
    \item \textbf{Summarize Success:} For all \texttt{right} nodes leading up to the error, state the evidence, timestamps, and logic as established facts (e.g., "Here is what we know so far...").
    \item \textbf{Guide the Error:} For the \textbf{first} \texttt{wrong} or \texttt{divergent} node:
    \begin{compactenum}
        \item \textit{Evidence Errors:} State the correct \texttt{evidence} and \texttt{timestamp}.
        \item \textit{Reasoning Errors:} State the correct \texttt{evidence}, \texttt{timestamp}, and the correct line of reasoning.
    \end{compactenum}
    \item \textbf{Constraints:} Do \textbf{not} mention past failures (e.g., "You missed") and do \textbf{not} reveal info about \texttt{undetermined} nodes.
\end{compactenum}

\vspace{1mm}
\hrule
\vspace{1mm}

\textbf{Input} \\
\textbf{Question (English):} \texttt{\{question\_english\}} \\
\textbf{Additional Cues:} \texttt{\{additional\_cues\}} \\
\textbf{Ground Truth Answer (English):} \texttt{\{ground\_truth\_answer\_english\}} \\
\textbf{Model Prediction (English):} \texttt{\{model\_prediction\_english\}} \\
\textbf{Model Thoughts/Outputs:} \texttt{\{model\_output\_thoughts\}} \\
\textbf{Evidence Graph:} \texttt{```json\{evidence\_graph\_json\}```} \\
\textbf{Human Reasoning (Raw):} \texttt{\{human\_reasoning\}}

\vspace{1mm}
\hrule
\vspace{1mm}

\textbf{Output Format} \\
Your output must be a single JSON object. Do not include any other text or explanation outside the JSON.

\begin{quote}
\texttt{\{ \\
\indent "solution\_graph": \{ \\
\indent \indent "node\_id": \{ \\
\indent \indent \indent "evidence": "string", \\
\indent \indent \indent "timestamp": "string or null", \\
\indent \indent \indent "parent\_nodes": ["string"], \\
\indent \indent \indent "outputs": \{ \\
\indent \indent \indent \indent "divergence": "boolean or null", \\
\indent \indent \indent \indent "evidence\_retrieved": "string or null", \\
\indent \indent \indent \indent "status": "string or null", \\
\indent \indent \indent \indent "missing\_evidence\_reason": "string or null", \\
\indent \indent \indent \indent "justification": "string" \\
\indent \indent \indent \} \\
\indent \indent \} \\
\indent \}, \\
\indent "next\_attempt\_cue": "string or null" \\
\}}
\end{quote}
\end{mdframed}
\end{minipage}
\caption{\textbf{Prompt to traverse through the graph and tag errors. (Part 2)}}
\label{fig:tag_graph_2}
\end{figure*}

\pagebreak
\clearpage

\newpage
\section{Additional qualitative examples}
\label{supp_sec:more_qual_examples}
\begin{figure*}[p]
    \centering
    % OPTIMIZATION: Make caption text small
    \captionsetup{font=small, skip=2pt, justification=centering}

    % --- PART 1: TOP HEADER ---
    \begin{minipage}{\textwidth}
        \centering
        \includegraphics[width=\linewidth, height=3.5cm, keepaspectratio]{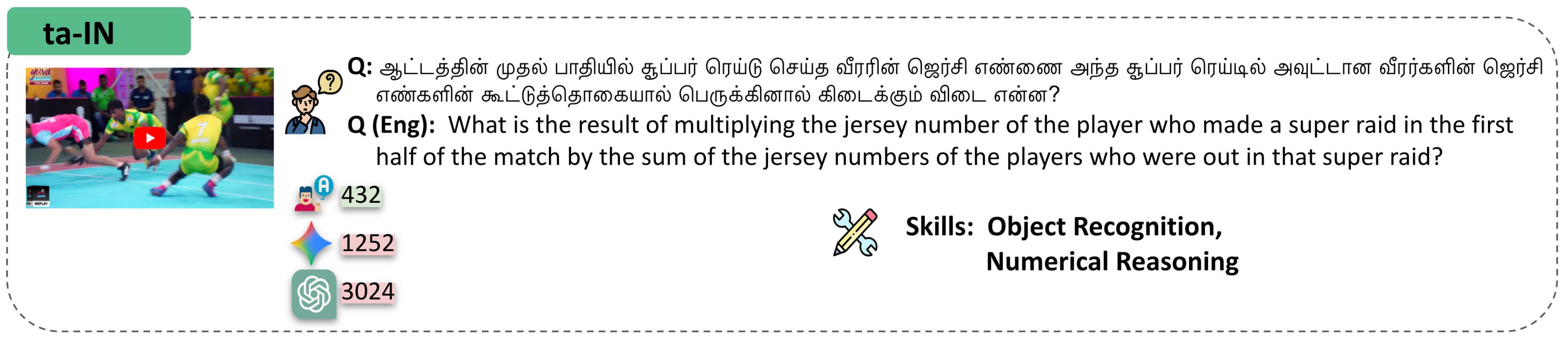} 
        \par
        \small
        \textbf{Context:} A question from the \texttt{ta-IN} locale. Question is translated here for better understanding. The reasoning trace is depicted as \emph{Evidence Graph} in the figures below. We notice that both \gpt{} and \geminipro{} both arrive at the wrong answer but different types of errors occur as seen in \emph{Iterative Error Isolation}.
    \end{minipage}
    
    \par\vspace{1em} 
    \hrule
    \par\vspace{1em}

    % --- LEFT COLUMN (NOW 3 IMAGES) ---
    % [c] aligns this block to the vertical center relative to the right block
    \begin{minipage}[c]{0.48\textwidth} 
        \centering
        \textbf{\geminipro{}: Iterative Error Isolation} \par\medskip
        \setcounter{subfigure}{0} % Reset counter to (a)

        % Step 1
        \begin{subfigure}[b]{\linewidth}
            \centering
            % INCREASED HEIGHT: From 3.2cm to 5cm since we have more room now
            \includegraphics[width=\linewidth, height=3.2cm, keepaspectratio]{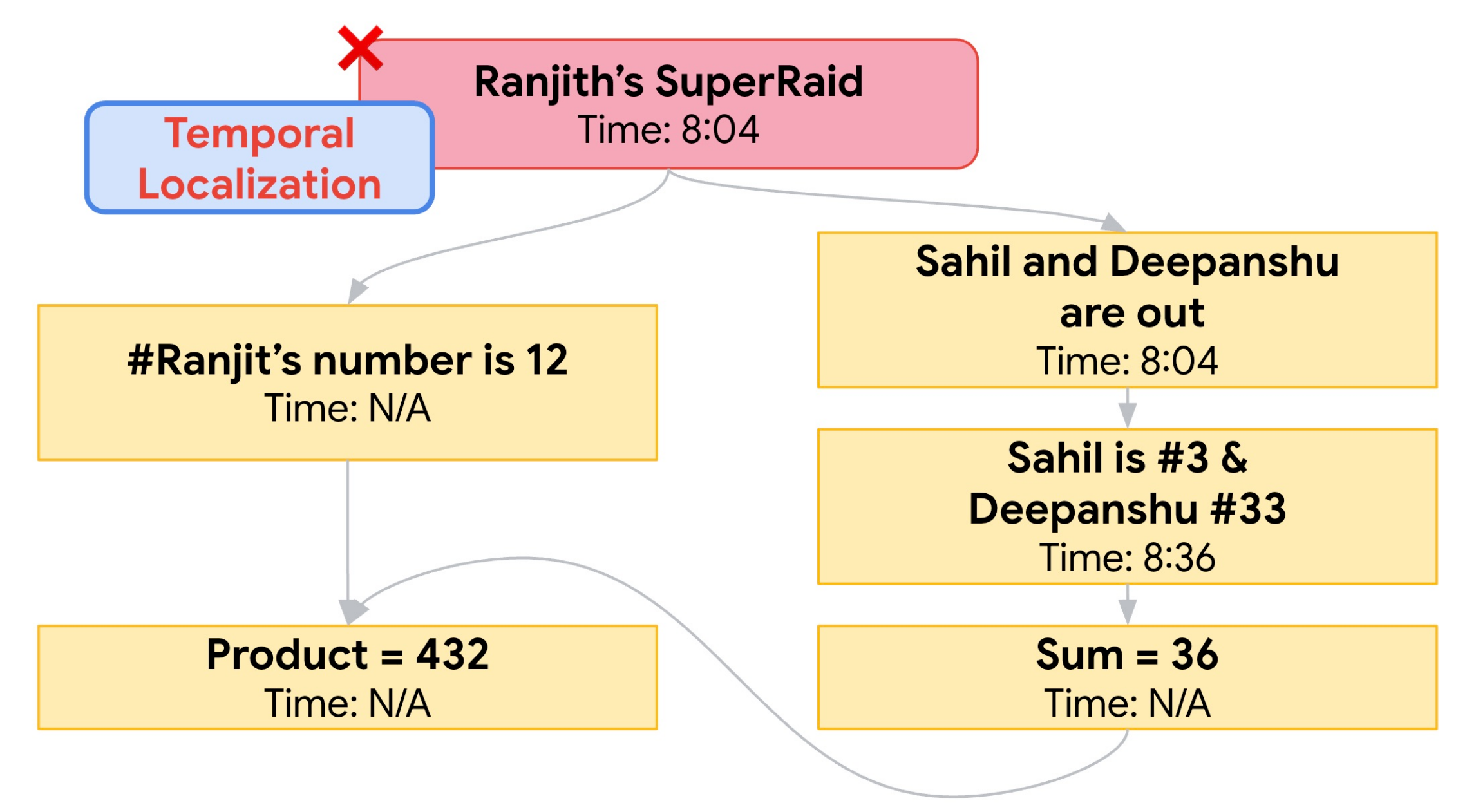}
            \caption{\textbf{Iter 1:} Model was unable to even retrieve the correct timestamp, hence \textcolor{red}{Temporal Localization} and hence other unevaluated nodes are in \textcolor{myyellow}{yellow}.}
        \end{subfigure}
        \vspace{1.5em} % Increased spacing
        
        % Step 2
        \begin{subfigure}[b]{\linewidth}
            \centering
            % INCREASED HEIGHT: From 3.2cm to 5cm
            \includegraphics[width=\linewidth, height=3.2cm, keepaspectratio]{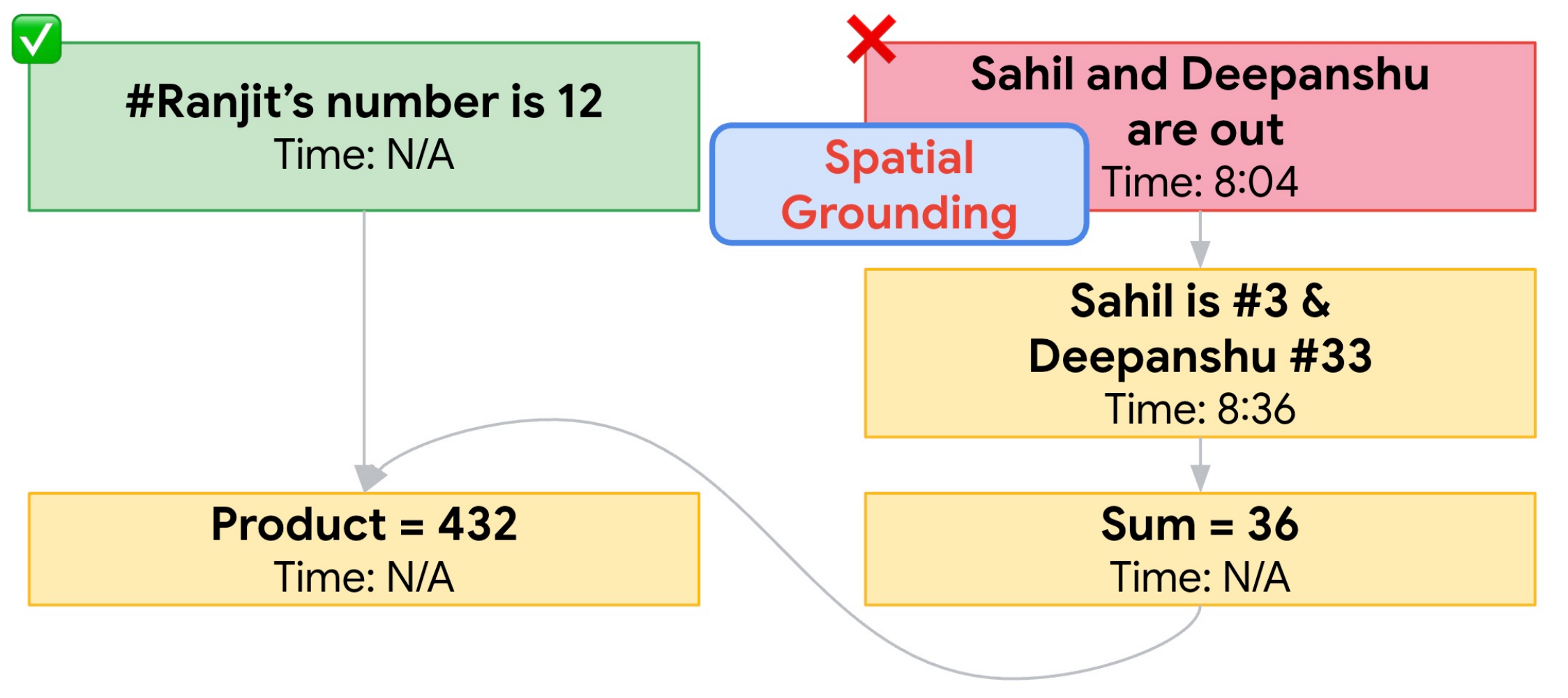}
            \caption{\textbf{Iter 2:} Provided additional context of the timestamp of the raid. Finds Ranjit's number \textcolor{forestgreen}{correctly} but fails to locate others (\textcolor{red}{Spatial Grounding}).}
        \end{subfigure}
        \vspace{1.5em} % Increased spacing

        % --- REMOVED STEP 3 HERE ---

        % Step 4 (Now the 3rd visual element)
        \begin{subfigure}[b]{\linewidth}
            \centering
            % INCREASED HEIGHT: From 2.5cm to 4.5cm
            \includegraphics[width=\linewidth, height=3.2cm, keepaspectratio]{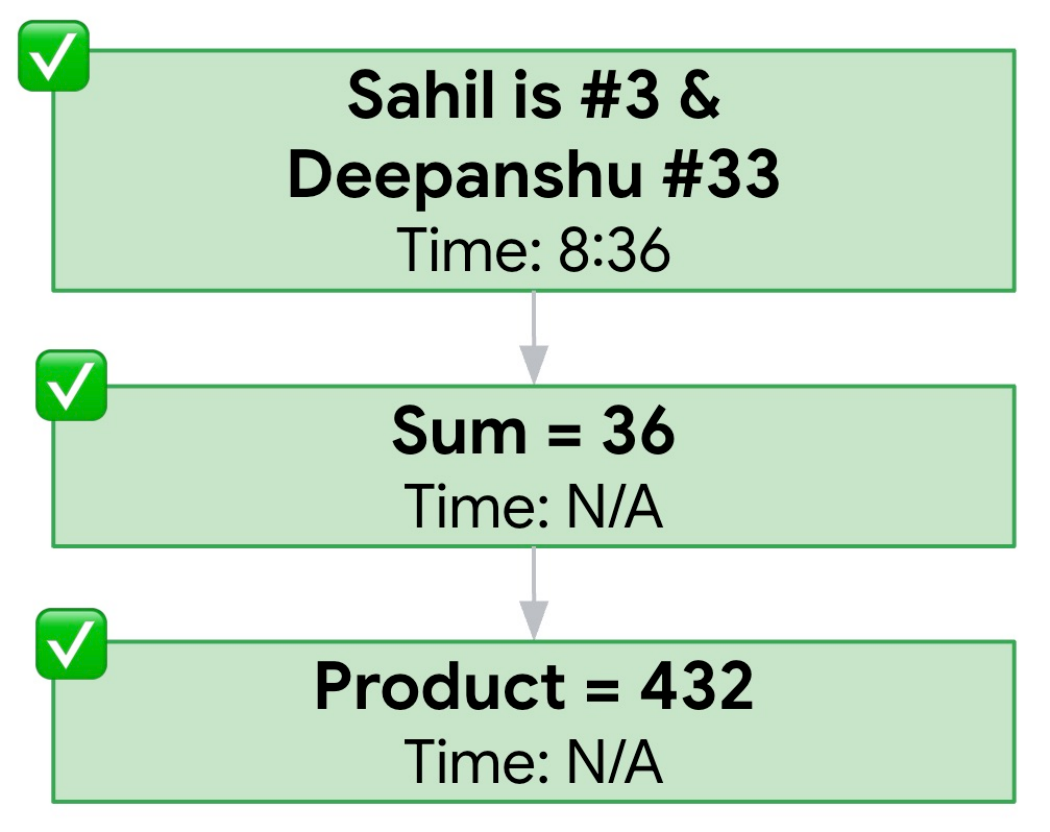}
            \caption{\textbf{Iter 3:} After all previous contexts, model answers \textcolor{forestgreen}{correctly}.}
        \end{subfigure}
    \end{minipage}
    \hfill \vline \hfill % Vertical Separator
    % --- RIGHT COLUMN (2 IMAGES) ---
    \begin{minipage}[c]{0.45\textwidth}
        \centering
        \textbf{\gpt{}: Iterative Error Isolation} \par\medskip
        \setcounter{subfigure}{0} % Reset counter to (a)

        % Comparison A
        \begin{subfigure}[b]{\linewidth}
            \centering
            \includegraphics[width=\linewidth, height=4cm, keepaspectratio]{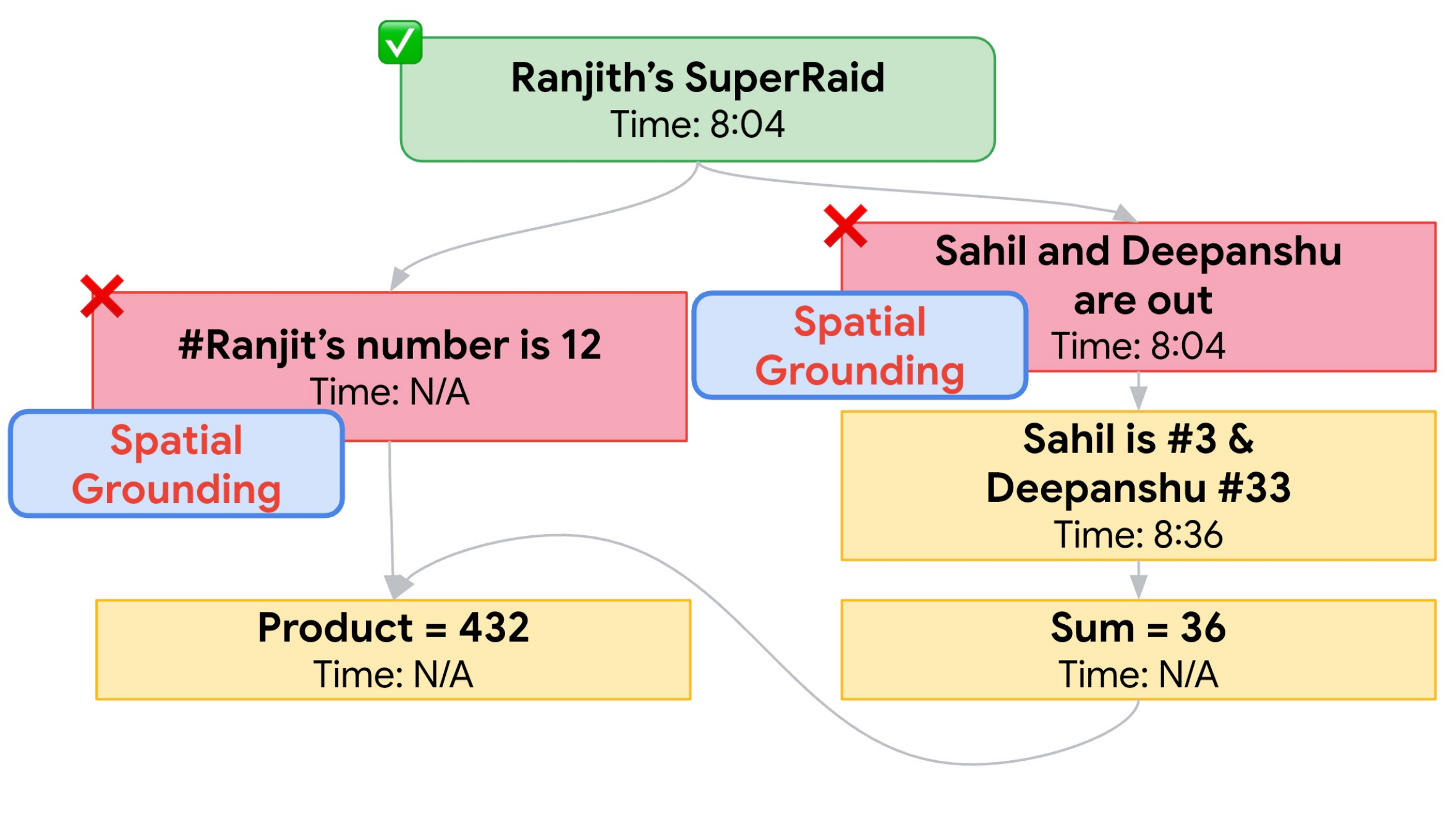}
            \caption{\textbf{Iter 1:} Model was able to retrieve the \textcolor{forestgreen}{correct timestamp}. However, it didn't identify any of the players right \textcolor{red}{Spatial Grounding} and hence other unevaluated nodes are in \textcolor{myyellow}{yellow}.}
        \end{subfigure}
        \vspace{2em} 

        % Comparison B
        \begin{subfigure}[b]{\linewidth}
            \centering
            \includegraphics[width=\linewidth, height=4cm, keepaspectratio]{figures/appendix_figures/graph6_q2.pdf}
            \caption{\textbf{Iter 2:} Provided additional context of the raid and the player information from visited nodes. Model is able to complete all the remaining steps \textcolor{forestgreen}{correctly}.}
        \end{subfigure}
    \end{minipage}

    % Main Caption
    \caption{Question from \texttt{ta-IN} locale of \benchmark{} and \emph{Error Isolation} graphs for \geminipro{} and \gpt{}.}
    \label{fig:ta_IN}
\end{figure*}

\begin{figure*}[t!]
    \centering
    % Setup for centered, readable captions
    \captionsetup[subfigure]{justification=centering}

    % --- PART 1: Full Page Width Header ---
    \begin{minipage}{\textwidth}
        \centering
        % We can use a generous height here since the page is not crowded
        \includegraphics[width=\linewidth, height=3.5cm, keepaspectratio]{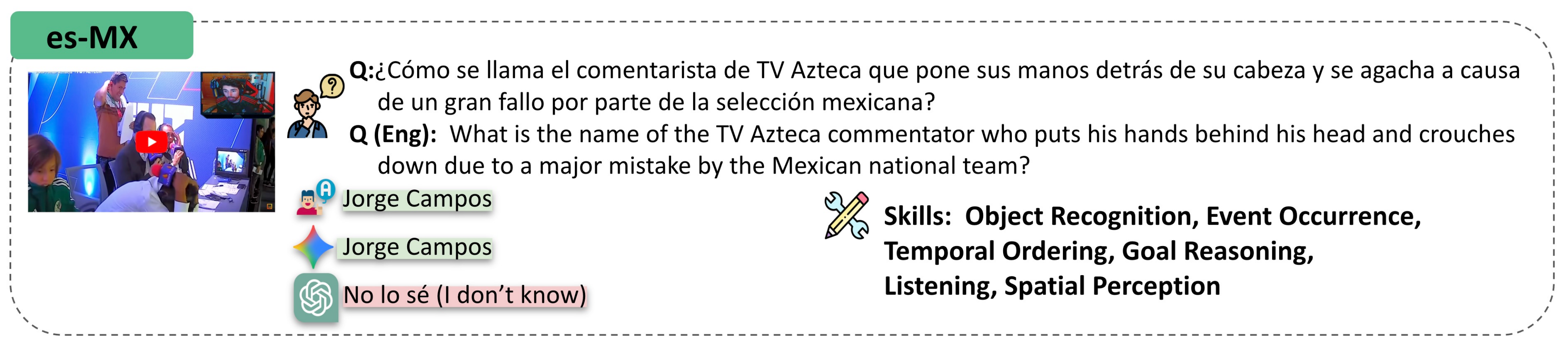} 
        \par\medskip
        \small
        \textbf{Context:} A question from the \texttt{es-MX} locale. Question is translated here for better understanding. The reasoning trace is depicted as \emph{Evidence Graph} in the figures below.
    \end{minipage}
    
    \par\bigskip 
    \hrule % Horizontal separator
    \par\bigskip

    % --- LEFT COLUMN ---
    % [c] aligns vertically to the center. Since this side is shorter, 
    % the single image will sit in the vertical middle relative to the right column.
    \begin{minipage}[c]{0.45\textwidth}
        \centering
        \textbf{\geminipro{}: Iterative Error Isolation} \par\medskip
        
        % Restart counter for safety (starts at a)
        \setcounter{subfigure}{0}

        % Image 1 (Kept)
        \begin{subfigure}[b]{\linewidth}
            \begin{center}
                \includegraphics[width=\linewidth, height=4cm, keepaspectratio]{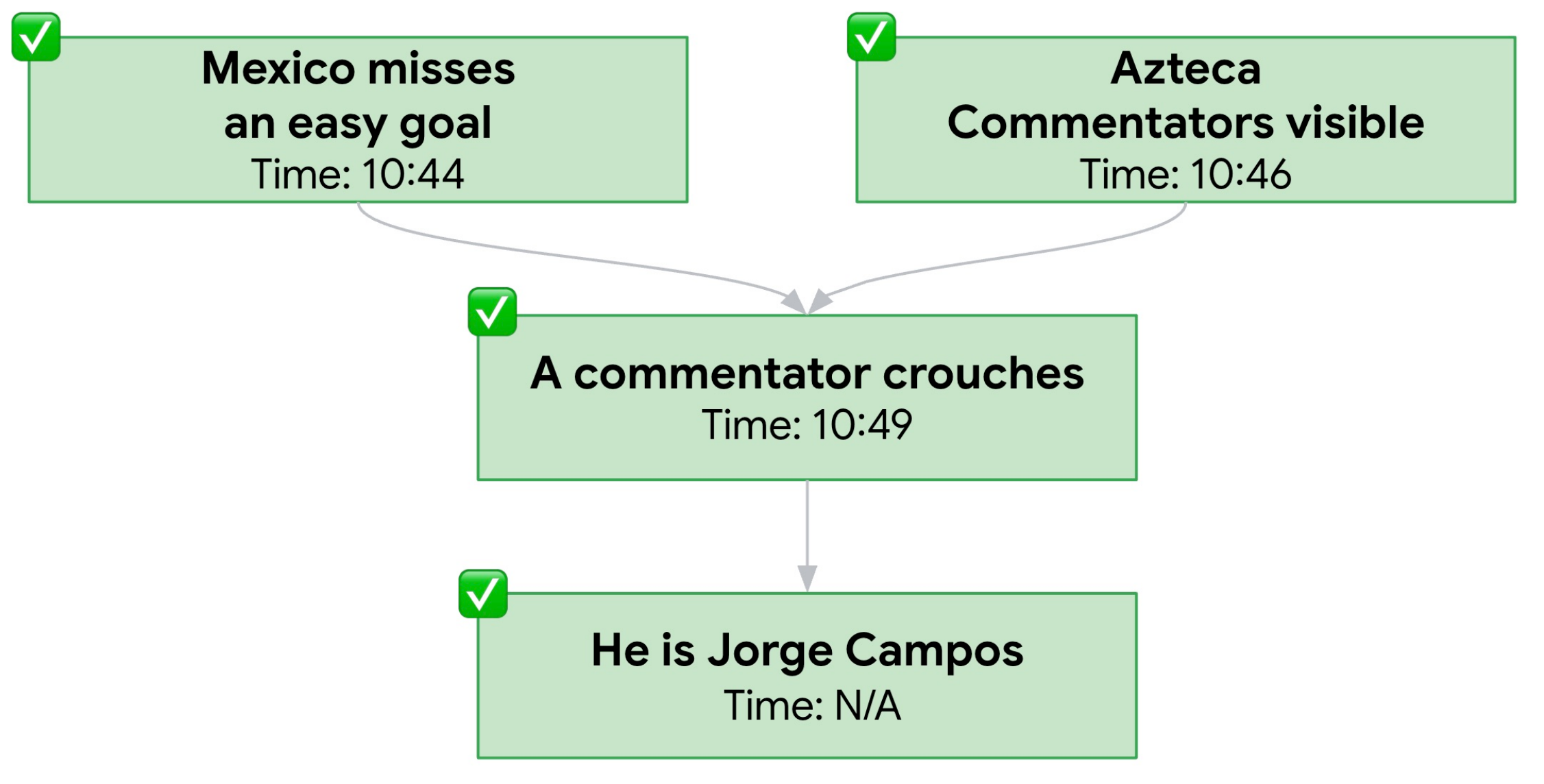}
            \end{center}
            \caption{\textbf{Iter 1:} Model was able to solve the question \textcolor{forestgreen}{correctly} in the first inference. Hence, all evidences are \textcolor{forestgreen}{correctly mapped} in the model reasoning traces.}
            % \label{fig:part2a}
        \end{subfigure}
        
        % Second image removed from here
    \end{minipage}
    \hfill % Push columns apart
    \vline % Vertical Line
    \hfill % Push columns apart
    % --- RIGHT COLUMN ---
    \begin{minipage}[c]{0.45\textwidth}
        \centering
        \textbf{\gpt{}: Iterative Error Isolation} \par\medskip
        
        % !!! RESTART NUMBERING !!!
        \setcounter{subfigure}{0}

        % Image 1 (Kept)
        \begin{subfigure}[b]{\linewidth}
            \centering
            \includegraphics[width=\linewidth, height=4cm, keepaspectratio]{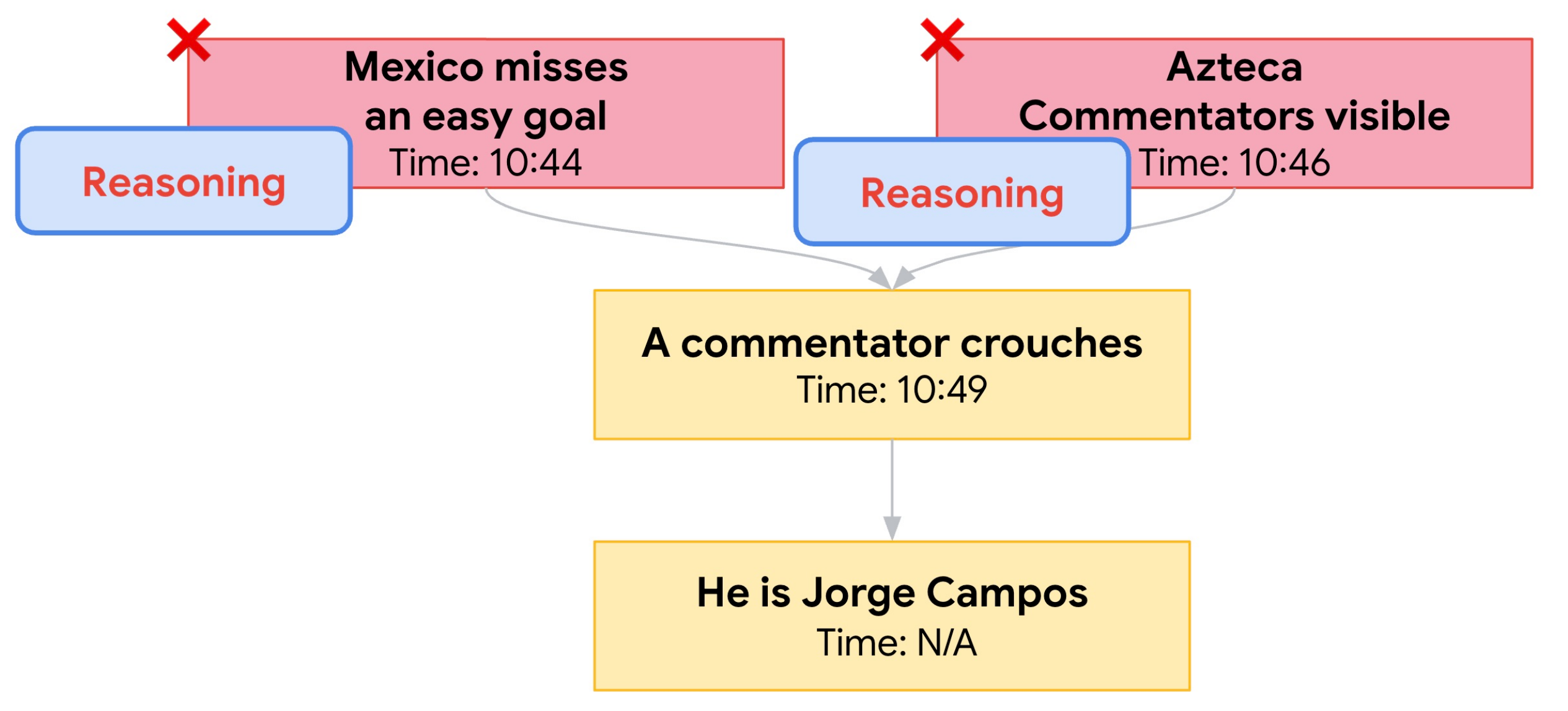}
            \caption{\textbf{Iter 1:} Model did not even plan how to solve the problem (\textcolor{red}{Reasoning}). It directly decides that it does not know the answer. All other unevaluated nodes are in \textcolor{myyellow}{yellow}.}
            % \label{fig:part2b}
        \end{subfigure}
        \vspace{2em} % Good spacing between images
        
        % Image 2 (Kept)
        \begin{subfigure}[b]{\linewidth}
            \centering
            \includegraphics[width=\linewidth, height=3cm, keepaspectratio]{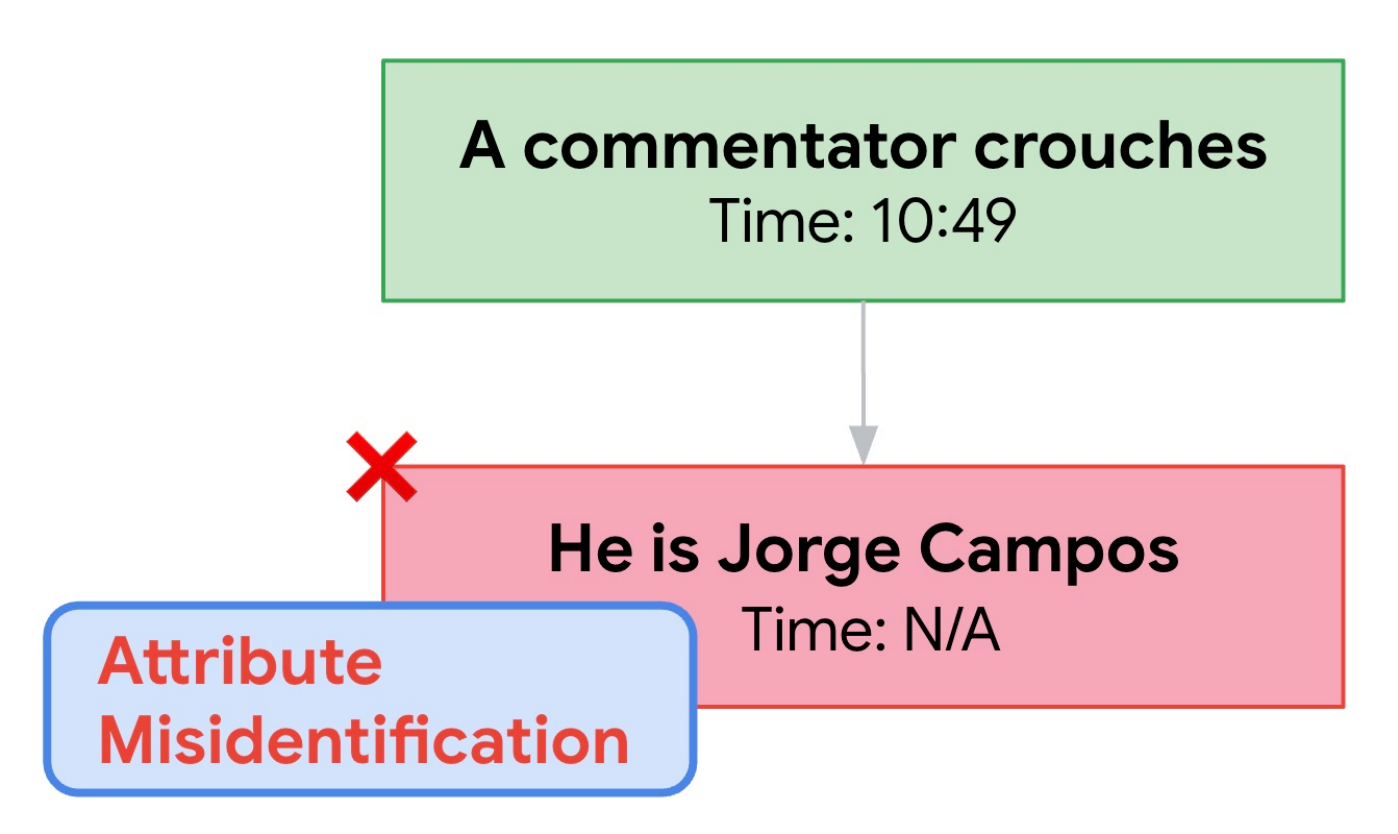}
            \caption{\textbf{Iter 2:} The wrong evidences in \textbf{Iter 1} are additionally provided to the model. The model then \textcolor{forestgreen}{correctly spatially grounds} the crouching commentator but misidentifies him as \textcolor{red}{Luis García}. Since the model successfully grounded the target, the failure in the visual-task of naming is classified as a \textcolor{red}{Attribute MisIdentification} error rather than a lack of External Knowledge.}
            % \label{fig:part3b}
        \end{subfigure}
    \end{minipage}

    % Main Caption
    \caption{Question from \texttt{es-MX} locale of \benchmark{} and \emph{Error Isolation} graphs for \geminipro{} and \gpt{}. \geminipro{} gets this question right, while \gpt{} makes an error.}
    \label{fig:es_MX}

\end{figure*}

\begin{figure*}[t!]
    \centering
    % Setup for centered, readable captions
    \captionsetup[subfigure]{justification=centering}

    % --- PART 1: Full Page Width Header ---
    \begin{minipage}{\textwidth}
        \centering
        % We can use a generous height here since the page is not crowded
        \includegraphics[width=\linewidth, height=3.5cm, keepaspectratio]{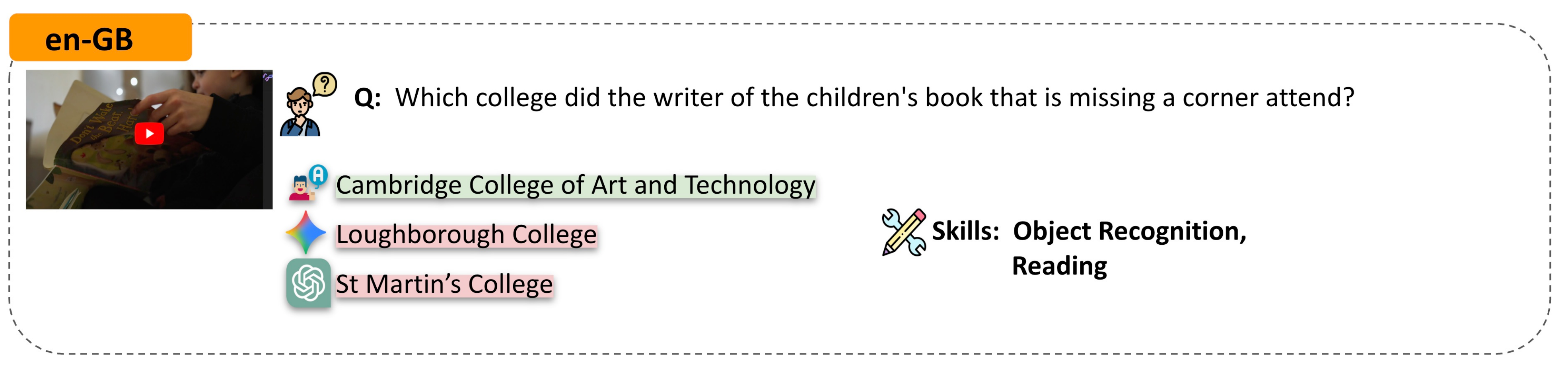} 
        \par\medskip
        \small
        \textbf{Context:} A question from the \texttt{en-GB} locale. The reasoning trace is depicted as \emph{Evidence Graph} in the figures below..
    \end{minipage}
    
    \par\bigskip 
    \hrule % Horizontal separator
    \par\bigskip

    % --- LEFT COLUMN ---
    % [c] aligns vertically to the center
    \begin{minipage}[c]{0.45\textwidth}
        \centering
        \textbf{\geminipro{}: Iterative Error Isolation} \par\medskip
        
        % Restart counter for safety (starts at a)
        \setcounter{subfigure}{0}

        % Image 1 (Previously Row 2 Left)
        \begin{subfigure}[b]{\linewidth}
            \begin{center}
                \includegraphics[width=\linewidth, height=4cm, keepaspectratio]{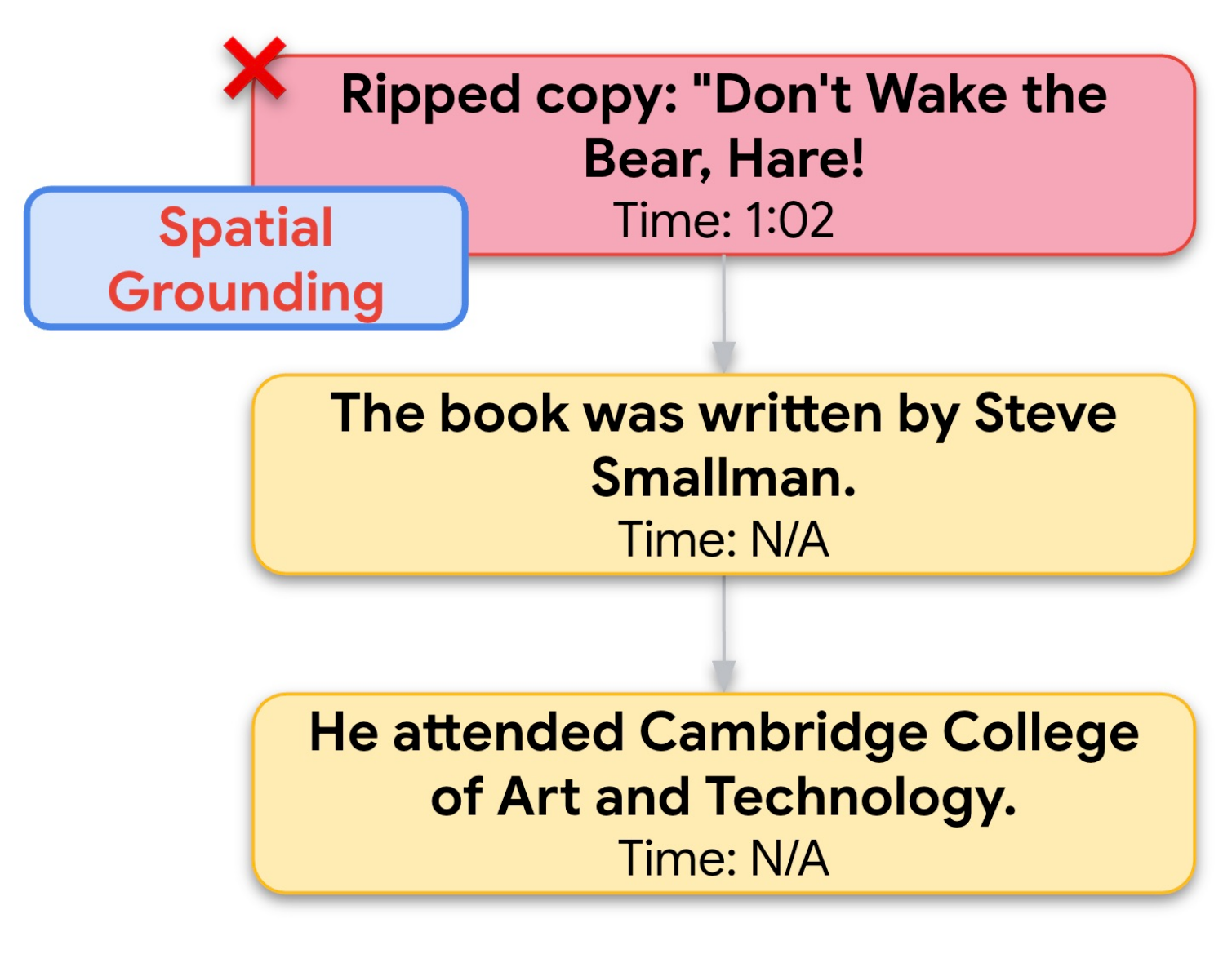}
            \end{center}
            \caption{\textbf{Iter 1:} Model was able to look at the \textcolor{forestgreen}{correct timestamp}. However, it didn't identify the required book with a ripped corner (\textcolor{red}{Spatial Grounding}). It instead focused on another book , and hence other unevaluated nodes are in \textcolor{myyellow}{yellow}.}
            % \label{fig:part2a}
        \end{subfigure}
        \vspace{2em} % Good spacing between images
        
        % Image 2 (Previously Row 3 Left)
        \begin{subfigure}[b]{\linewidth}
            \centering
            \includegraphics[width=\linewidth, height=3cm, keepaspectratio]{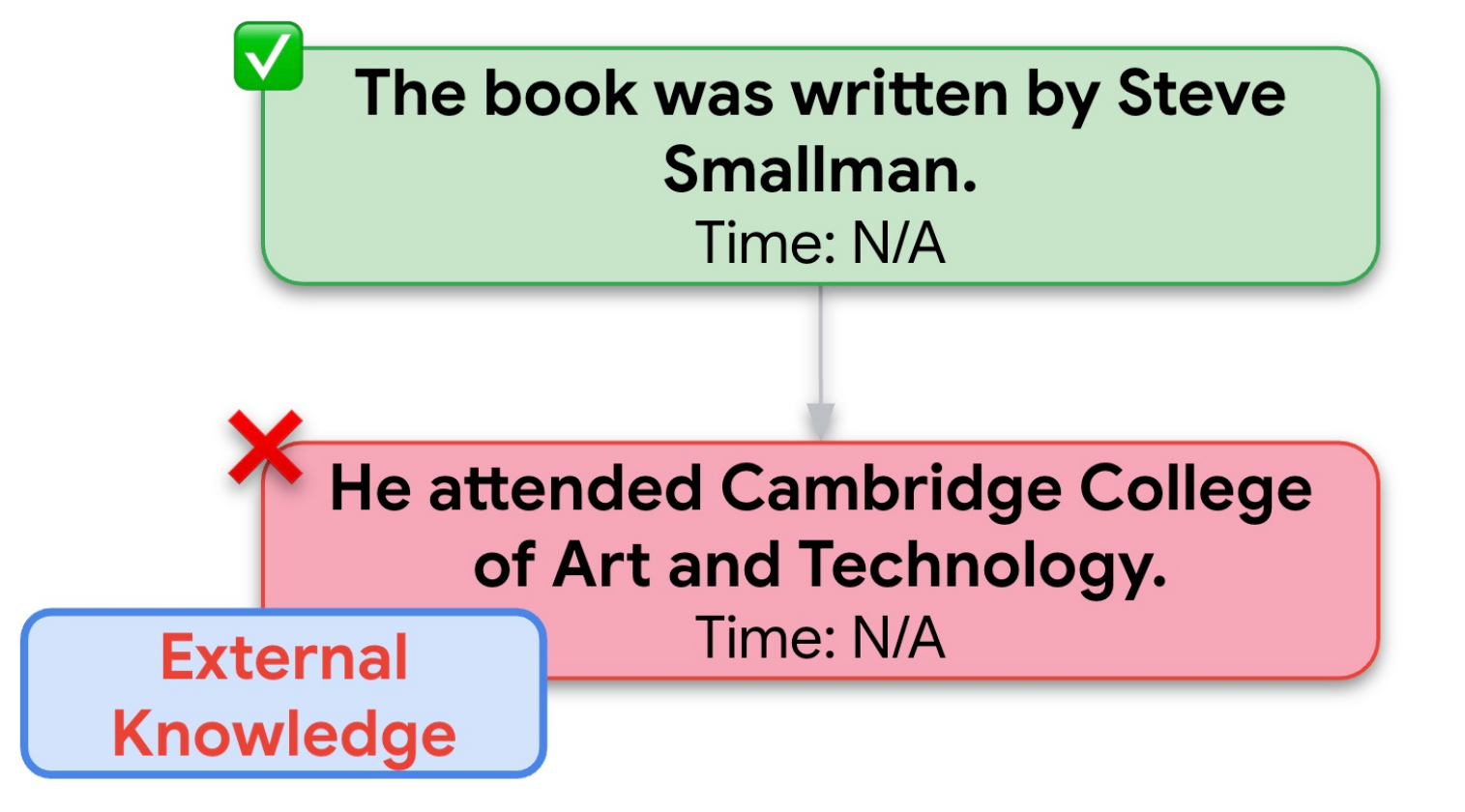}
            \caption{\textbf{Iter 2:} Additional context of the book and the timestamp is provided. The model identifies the author \textcolor{forestgreen}{correctly}. However, it didn't identify the college he attended, an information that was not in the video (\textcolor{red}{External Knowledge}).}
            % \label{fig:part3a}
        \end{subfigure}
    \end{minipage}
    \hfill % Push columns apart
    \vline % Vertical Line
    \hfill % Push columns apart
    % --- RIGHT COLUMN ---
    \begin{minipage}[c]{0.45\textwidth}
        \centering
        \textbf{\gpt{}: Iterative Error Isolation} \par\medskip
        
        % !!! RESTART NUMBERING !!!
        \setcounter{subfigure}{0}

        % Image 1 (Previously Row 2 Right)
        \begin{subfigure}[b]{\linewidth}
            \centering
            \includegraphics[width=\linewidth, height=4cm, keepaspectratio]{figures/appendix_figures/graph1_q1.pdf}
            \caption{\textbf{Iter 1:} Model was able to look at the \textcolor{forestgreen}{correct timestamp}. However, it didn't identify the required book itself (\textcolor{red}{Spatial Grounding}). It instead focused on another book , and hence other unevaluated nodes are in \textcolor{myyellow}{yellow}.}
            % \label{fig:part2b}
        \end{subfigure}
        \vspace{2em} % Good spacing between images
        
        % Image 2 (Previously Row 3 Right)
        \begin{subfigure}[b]{\linewidth}
            \centering
            \includegraphics[width=\linewidth, height=3cm, keepaspectratio]{figures/appendix_figures/graph2_q1.pdf}
            \caption{\textbf{Iter 2:} Additional context of the book and the timestamp is provided. The model identifies the author \textcolor{forestgreen}{correctly}. However, it didn't identify the college he attended, an information that was not in the video (\textcolor{red}{External Knowledge}).}
            % \label{fig:part3b}
        \end{subfigure}
    \end{minipage}

    % Main Caption
    \caption{Question from \texttt{en-GB} locale of \benchmark{} and \emph{Error Isolation} graphs for \geminipro{} and \gpt{}. Both models make similar type of errors in this example.}
    \label{fig:en_GB}

\end{figure*}

We present qualitative examples from \benchmark{} in Figure~\ref{fig:ta_IN} (\texttt{ta-IN}), Figure~\ref{fig:es_MX} (\texttt{es-MX}), and Figure~\ref{fig:en_GB} (\texttt{en-GB}). These figures illustrate the \emph{Iterative Error Isolation} analysis applied to both \geminipro{} and \gpt{}.

\end{document}